\documentclass[letterpaper]{article} %
\usepackage{aaai2027}
\nocopyright %
\usepackage[hyphens]{url}  %
\usepackage{graphicx} %
\usepackage{natbib}  %
\usepackage{caption} %
\usepackage{algorithm}
\usepackage{algorithmic}

\usepackage{newfloat}
\usepackage{listings}
\usepackage{tcolorbox}
\tcbuselibrary{breakable,skins}
\newtcolorbox{promptbox}[1]{enhanced,breakable,colback=gray!3,colframe=gray!40,boxrule=0.4pt,arc=0pt,left=3pt,right=3pt,top=2pt,bottom=2pt,toptitle=1.5pt,bottomtitle=1.5pt,coltitle=black,colbacktitle=gray!15,fonttitle=\sffamily\bfseries\footnotesize,fontupper=\small,before skip=6pt,after skip=6pt,title={#1}}
\DeclareCaptionStyle{ruled}{labelfont=normalfont,labelsep=colon,strut=off} %
\floatstyle{ruled}
\newfloat{listing}{tb}{lst}{}
\floatname{listing}{Listing}

\usepackage{booktabs}

\title{How Should We Rerank Multi-Constraint Video Search?\\Whole-Query Judgments, Frames, and Their Cost}
\author{
    Isaac Chung\textsuperscript{\rm 1},
    Bolin Pan\textsuperscript{\rm 1},
    Chen Liang\textsuperscript{\rm 1},
    Nawal Singh\textsuperscript{\rm 1},
    Anas Khan\textsuperscript{\rm 1},
    Daniel Del Castillo Llano\textsuperscript{\rm 1},
    Eran Nussinovitch\textsuperscript{\rm 1}
}
\affiliations{
    \textsuperscript{\rm 1}Foam (foam.io)
}
\newcommand{\codeurl}{\url{https://github.com/isaac-chung/mcvr}}
\newcommand{\dataurl}{\url{https://hf.co/datasets/isaacchung/MCVR}}

\begin{document}

\maketitle

\begin{abstract}
Video search often ranks videos by a short description that a vision-language model writes before any query arrives, and that description can omit what separates a full match from a near miss.
This paper investigates whether splitting the query helps, or whether the system needs richer evidence.
We build 722 LLM-written, human-checked queries over 8,603 videos, with static visual constraints and hard-negative near misses, and test both options.
The results show that splitting does not help search: per-constraint search loses to one whole-query search; on its top descriptions one LLM judgment of the whole query outperforms our per-constraint grading by about 0.10 nDCG@10, while the two perform comparably on frames.
The best-performing of three 2B cross-encoders, which also read the whole query, recovers about three quarters of the LLM's gain over whole-query search and outperforms our per-constraint grading.
Richer visual evidence does help: reranking the description reranker's top ten on video frames adds 0.08 nDCG@10 but needs nearly $9\times$ its input tokens, and a human check agrees with it on cases where the two rerankers disagree.
Longer descriptions recover two thirds of this gain at a fraction of the cost, while reranking only close calls on frames cuts input tokens by a fifth to a third.
We release the dataset and code.

\end{abstract}

\begin{links}
    \par\textbf{Code} --- \codeurl
    \par\textbf{Datasets} --- \dataurl
\end{links}

\section{Introduction}

Video platforms are searched by what is visibly present, and queries can combine several conditions, such as an object, a setting and an action \citep{awad2023trecvid2022}. Commonly, a vision-language model (VLM) writes a description of each video offline from sampled frames and a text embedder indexes the descriptions \citep{jia2025summa,huynh2025awsvideosearch,zhang2024omagent}; other systems embed the video directly \citep{miller2026condenast} or fuse several indexes \citep{samuel2025mmmorrf}. A description index runs on ordinary text search and keeps reranking cheap: a 44-word description costs a reranker about 40 times fewer tokens than the video's frames. Written before any query is known, however, a description may omit details a query depends on (Finding~\ref{sec:res-mechanism} and Appendix~\ref{app:length} measure what is missing; Section~\ref{sec:limitations} notes what we do not separate). A single search may then rank a video meeting three of four criteria above one meeting all four, as for ``a woman in a red apron stands by a sign reading `OPEN' in a brightly lit bakery''; image-text embedders, though they match text to images directly, fail similarly when a query combines several specific constraints \citep{kargi2026halftruths}. To mitigate this, a system built on such an index could split the query into one check per condition, judge the whole query with a stronger reranker, or pay to read the frames of its top candidates.
We investigate:
\textbf{RQ1:} where should decomposition act, at retrieval or at verification, when the description a verifier reads may lack the needed detail?
\textbf{RQ2:} does verifying against the frames beat verifying against the description, and how does the gap change with the number of constraints $k$ and with video length?
\textbf{RQ3:} what share of the frame gain does a trigger on the description reranker's score margin keep at each share of queries it sends to frames?
We study these questions on a benchmark with static visual facts, each checkable in a single frame (see Section~\ref{sec:limitations}).

In this paper, we make three contributions: (1)~a controlled comparison, over one description index, of whole-query search, decomposed retrieval, and description or frame verification for multi-constraint video retrieval. On the same candidates, when the verifier reads descriptions, one judgment of whether a video fits the whole query outperforms grading each constraint and averaging; when it reads frames, the two tie. This holds for our two prompts, not for decomposition in general (Section~\ref{sec:limitations}). The best of three off-the-shelf 2B cross-encoders, which also read the whole query, outperforms our per-constraint grading and gets about three quarters of the LLM's nDCG@10 gain over whole-query search (Finding~\ref{sec:res-local}). (2)~A benchmark of 722 queries over 8,603 videos, in a nested set, where each query adds a constraint to the last, and a near-miss set, where a hard-negative video meets all but one constraint. (3)~An analysis of reading frames only for the reranker's top candidates, and of a trigger that sends close calls to frames.

\section{Related Work}
\label{sec:related}

\paragraph{Decomposition for multi-condition retrieval.}
MultiConIR \citep{lu2025multiconir} shows retrievers degrade as a text query's conditions grow. \citet{yin2026stageaware} ask where decomposition should act: on MultiConIR, decomposing queries of four or more conditions can hurt first-stage retrieval, while checking each sub-query when reranking generally helps, more so as conditions grow. Their setting is text only and reports time, not token cost. \citet{shen2025digitaltwin} split implicit reasoning queries into sub-queries to retrieve videos by object-level representations, then have an LLM check the whole query and call vision models only when evidence is missing, but neither compare where to split nor report cost, so whether the text-only finding holds for video is untested. In video, a description written before the query may omit what it asks about, and frames show more at a higher cost. MUVR \citep{feng2025muvr} benchmarks untrimmed-video retrieval with multi-modal queries at several levels of visual correspondence and finds MLLMs weak at reranking many videos. \emph{We carry the question to video, where the queries include hard-negative near-misses, and report the per-query token cost of each verification strategy.}

\paragraph{Cross-encoder reranking.}
Cross-encoders encode a query and a candidate jointly and usually rerank after embedding retrieval; recent ones build on multimodal LLM backbones and rerank text or images \citep{li2026qwen3vlreranker,ananya2026lightonrerank}. MultiConIR finds that fine-tuned rerankers beat retrievers on single-condition queries but degrade more sharply as conditions are added \citep{lu2025multiconir}. Whether off-the-shelf rerankers degrade the same way is open. \emph{We run them zero-shot on whole queries over descriptions as a cheap local baseline, and ask how much of the LLM reranker's gain they recover.}

\paragraph{Video retrieval and the use of frames.}
MAVIS \citep{zhang2026mavis} filters candidates by embedding scores of offline LLM captions against scene, object and action sub-queries and re-scores survivors by ViCLIP frame similarity; no generative model reads the frames. Embedding similarity can rank a partial match above a full one \citep{kargi2026halftruths}. We did not test an embedding re-scorer on our index, so we cannot attribute our frame gain (RQ2) to generative verification over embedding similarity. Pointwise VLM rerankers judge whole multi-condition queries over product images and text \citep{lu2026mcmr}, an off-the-shelf VLM scores video shots 1--10 against the whole query at TRECVID \citep{goulas2025certh}, and a trained frame reranker beats a caption-reading reranker built on another model \citep{skow2026rankvideo}; none compares one verifier on descriptions and frames, or reports token cost. In image-text alignment, a single yes-probability check \citep{lin2024vqascore} outperforms decomposed question checks such as TIFA and DSG \citep{hu2023tifa,cho2024dsg}; averaged over sampled frames, it also beats a decomposed GPT-4V check of video--text alignment, but scores one video against its prompt. \citet{captiononce2026} route long-video questions to frames only when captions fall short, and frames fix a caption-only error on about 5\% of questions, but within one video. VideoSearch-R1 \citep{lee2026videosearchr1} refines its query only when its top-1 video fails a match check, but is trained, iterates and targets moment retrieval, rather than escalating once to frames over a fixed pool. RQ2 and RQ3 ask whether frames help in corpus retrieval, where many videos and near misses compete and the first stage caps the pool, and how the gain and its token cost vary with $k$ and with the share of queries sent to frames.

\section{Methodology}
\label{sec:method}

\paragraph{Setting.}
A query states $k\in\{1,\dots,4\}$ atomic constraints, each checkable by inspecting the video: an object, an attribute, on-screen text, a count, or the scene. We stop at four because a caption rarely yields more such facts (98 of our 100 test sources give a nested query at $k=1$, 73 at $k=4$), and four is where decomposing text queries starts to hurt first-stage retrieval \citep{yin2026stageaware}.

\paragraph{Description index.}
Gemini~3.5 Flash-Lite writes a description of each video from frames sampled uniformly at 0.5\,fps (the index frames), half the 1\,fps at which \citet{captiononce2026} caption video, to save cost. Descriptions average about 44 words, regardless of video length (Appendix~\ref{app:stats}).

\paragraph{Systems.}
S0 searches once with the whole query (Table~\ref{tab:systems}); S1 splits the query and searches once per constraint. S2 and S3 follow decomposed reranking \citep{yin2026stageaware}, but with our own integer grade from 0 to 10 per constraint, assigned by the LLM verifier \texttt{gpt-6-sol} (10 if the evidence clearly states it, 5 if it is not mentioned, 0 if it contradicts it; prompts in Appendix~\ref{app:prompts}), averaged over constraints, an aggregation fixed before the runs. We do not weight constraints, since we have no basis for ranking them, and five other aggregations, the minimum among them, leave S4@50 ahead of S2@50 (Appendix~\ref{app:results}, Table~\ref{tab:reagg}). S4 and S5 (cf.\ \citealp{lin2024vqascore,zhuang2024beyondyesno}) instead give one 0--10 score against the undecomposed query. We use one scale for S2--S5 so that granularity does not differ between them; graded scores beat yes/no answers in pointwise LLM ranking \citep{zhuang2024beyondyesno}, and binary per-constraint checks do no better here (Section~\ref{sec:res-ablations}). S6 returns S5's ranking when S4's top two scores differ by at most 2 points, and S4's otherwise, following margin-gated verification \citep{demidov2026beyond}. 

\paragraph{Cross-encoders.}
As baselines, three off-the-shelf cross-encoders score S0's top 50 descriptions against the whole query: Qwen3-VL-Reranker-2B \citep{li2026qwen3vlreranker}, LightOn-rerank-PW-2B and the list-wise LightOn-rerank-LW-2B \citep{ananya2026lightonrerank}, which ranks four descriptions per prompt (Qwen-2B, PW-2B and LW-2B in the tables). Each ran zero-shot, once, after the main runs.

\paragraph{Notation and defaults.}
A suffix @$d$ gives the rerank depth: S4@$d$ reranks S0's top $d$, and S2 and S4 without one mean depth 50. The frame systems (S3, S5, S6) rerank S4@50's top ten. All main results use one verifier (\texttt{gpt-6-sol}) and one description length; Section~\ref{sec:res-ablations} varies both. Details and prompts are in Appendix~\ref{app:systems}.

\begin{table}[t]\centering\small\setlength{\tabcolsep}{2pt}
\begin{tabular}{@{}lllll@{}}
\toprule
 & Query & Order by & Verifier sees & Candidates \\
\midrule
S0 & whole & similarity & none & all videos \\
S1 & split & RRF ranks & none & all videos \\
S2 & split & mean grade 0--10 & description & S0 top 50 \\
S3 & split & mean grade 0--10 & frames & S4@50 top 10 \\
S4 & whole & 0--10 score & description & S0 top 50 \\
CE & whole & model score & description & S0 top 50 \\
S5 & whole & 0--10 score & frames & S4@50 top 10 \\
S6 & whole & S5 or S4 & desc., then frames & S4@50 top 10 \\
\bottomrule
\end{tabular}
\caption{Systems, all over one index of descriptions. Query: the whole query, or split by an LLM into per-constraint searches (S1) or checks (S2, S3). CE: the three cross-encoders. Candidates: the videos each system ranks; RRF is reciprocal rank fusion.}
\label{tab:systems}
\end{table}

\section{Experimental Setup}
\label{sec:setup}

\paragraph{Corpus.}
We pool four parts of short web video, chosen because their detailed captions yield many visible facts per video (Appendix~\ref{app:corpus}), into one corpus of 8,603 videos: Shot2Story20K test \citep{han2023shot2story}, CaReBench \citep{xu2025carebench}, Vript short \citep{yang2024vript} (Shorts and TikTok videos up to 60\,s) and Vript long, spans of 30\,s to 4\,min cut from Vript's long YouTube videos. The source datasets are open-domain web video: Shot2Story and Vript come from YouTube (and TikTok, for Vript short) and cover YouTube's content categories, from entertainment and news to travel and gaming \citep{han2023shot2story,yang2024vript}, while CaReBench holds everyday human-activity clips from FineAction \citep{xu2025carebench}. Pooling the parts gives one index whose videos run from seconds to minutes and whose captions come from three sources (people, a model corrected by annotators, and GPT-4V), so results do not rest on a single video length or caption style.

\paragraph{Queries.}
Queries are built from dataset captions, which no system sees. We do not index the captions: queries are rewordings of facts mined from them and keep much of their vocabulary, so a caption index would let a system match the query's own words. Every video is indexed through a generated description instead, including the 580 Vript short videos that have no caption; these are never query sources or near-miss videos, but stay in the corpus as distractors. An LLM mines atomic facts from a source video's caption; a fact is kept only if visible in a single frame and confirmed on the video's frames by two VLMs from different families (the \emph{grounders}) \citep{min2023factscore,jing2024faithscore,verga2024poll}. An LLM then writes a sequence of nested queries, each extending the previous one, following the construction of \citet{lu2025multiconir}. The frame checks show the facts are visible, not that the queries read like user queries (Section~\ref{sec:limitations}).

\paragraph{Near-miss queries.}
Nested queries need not have a near match, so we add \emph{near-miss} queries. For a source $A$ we take a caption-embedding neighbour $B$ from another source, in the spirit of the nearest-neighbour hard negatives of \citet{xiong2020ance}, and verify on frames the facts the two share and one fact $d$ that holds only in $A$. The query states $k-1$ shared facts plus $d$, so $B$ is a hard-negative (HN) video meeting exactly $k-1$ of the $k$ constraints (a near miss). The 100 test sources, kept apart from 100 development sources, are stratified by corpus part and Vript length bin (15 Shot2Story, 15 CaReBench, 28 Vript short and 42 Vript long), so long videos are over-represented relative to the corpus. We oversample them because Shot2Story and CaReBench are almost all short clips (Appendix~\ref{app:corpus}) and length may affect both retrieval and description quality. Tables~\ref{tab:perk} and~\ref{tab:frames} give paired differences in nDCG@10 per length bin and corpus part; in Table~\ref{tab:perk} the S4 gains over S0 and S2b stay positive in every bin. The sources yield 350 nested and 372 near-miss test queries (Appendix~\ref{app:mining}).

\paragraph{Relevance judgments.}
Strict labels count only the source as relevant. Expanded labels (used unless noted) also count as relevant the other videos that two frame-based LLM judges, Gemini~3.5 Flash-Lite and \texttt{gpt-6-astra}, both accept on every fact the query states \citep{thomas2023llmpreferences,rahmani2024synthetic}. LLM judgments let us label deeply at corpus scale \citep{schall2026q2dweb}; they can preserve run-level conclusions but carry model and system biases \citep{upadhyay2024largescale,alaofi2026llmlabelling}. We therefore judge every top-10 candidate of every system (Appendix~\ref{app:mining}), report strict labels, which need no LLM judgment, and compare the judges with humans (Finding~\ref{sec:res-labels}).

\paragraph{Subgroups.}
We also report two post hoc, exploratory subgroups: the \emph{contested} near-miss queries, whose source and HN video are both in S4@50's top 10 (96 of 372), and a seeded sample of 56 nested Vript queries spread over $k$ and length.

\paragraph{Human check.}
An annotator who sees no system names or model answers labels the 100 test sources in four tasks (Appendix~\ref{app:human}). \emph{Task~A:} on the queries where S4@50 and S5 rank different videos first, does each top video meet each constraint? \emph{Task~B+D:} is each fact behind the queries on its source? \emph{Task~C:} for the near-miss sources Task~A does not cover, is each fact of the longest near-miss query met on the source and on its HN video? \emph{Task~E:} from text only, which facts does each index description state? A second annotator relabels part of Tasks~A, B+D and~E and all of Task~C.

\section{Results}
\label{sec:results}

\begin{figure*}[t]
\centering
\IfFileExists{Figures/fig_contrasts.pdf}{%
\includegraphics[width=\textwidth]{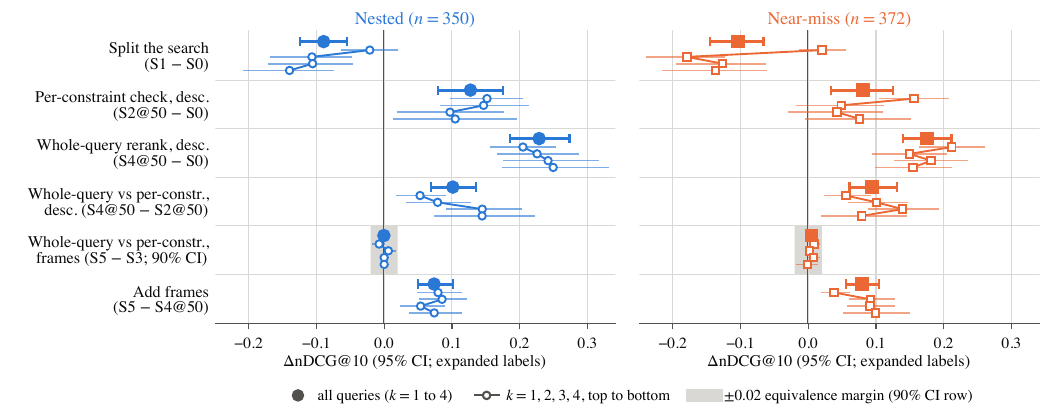}}{%
\fbox{\parbox[c][3.2cm][c]{0.92\textwidth}{\centering Figure pending (Figures/fig\_contrasts.pdf)}}}
\caption{The contrasts behind RQ1 and RQ2, in nDCG@10 (expanded labels) on all nested (left, $n=350$) and near-miss (right, $n=372$) test queries. Filled markers pool $k=1$ to 4, with 95\% intervals; the four hollow markers under each give $k=1$ to 4 from top to bottom ($n=98$, 94, 85, 73 nested and 98, 98, 98, 78 near-miss), each with its interval as a thin line. The frame row plots S5 $-$ S3 with 90\% intervals against the pre-set $\pm0.02$ equivalence margin (shaded; Table~\ref{tab:frames}).}
\label{fig:contrasts}
\end{figure*}

Table~\ref{tab:results} gives quality and cost for every system, and Figure~\ref{fig:contrasts} shows the contrasts behind RQ1 and RQ2. Intervals are 95\% bootstrap intervals on paired differences over the same queries, resampling source videos so the queries built from one video move together (Appendix~\ref{app:stats}); Holm-corrected tests for the description and S5 $-$ S4@50 contrasts are in Appendix~\ref{app:results}. We report the nested and near-miss sets together unless they point in different directions; per-set numbers, subgroup splits and post-hoc analyses are in Appendix~\ref{app:results}.

\IfFileExists{tables/table1.tex}{\begin{table}[t]
\centering
\small
\setlength{\tabcolsep}{2.4pt}
\begin{tabular}{@{}lccccccrr@{}}
\toprule
 & \multicolumn{2}{c}{Nested} & \multicolumn{2}{c}{Near-miss} & \multicolumn{2}{c}{R@1} & \multicolumn{2}{c}{Cost} \\
\cmidrule(lr){2-3}\cmidrule(lr){4-5}\cmidrule(lr){6-7}\cmidrule(l){8-9}
System & all & $k{=}4$ & all & $k{=}4$ & Nest. & N-miss & \#tok & $\times$S4 \\
\midrule
S0 & .408 & .485 & .450 & .447 & .294 & .401 & 0.0 & -- \\
S1 & .318 & .345 & .346 & .310 & .223 & .285 & 0.2 & 0.0 \\
S2@50 & .535 & .590 & .531 & .523 & .449 & .540 & 15.1 & 1.5 \\
S4@50 & \textbf{.636} & \textbf{.734} & \textbf{.626} & \textbf{.602} & \textbf{.586} & \textbf{.642} & 10.4 & 1.0 \\
\midrule
Qwen-2B & \textbf{.577} & \textbf{.698} & \textbf{.586} & \textbf{.570} & \textbf{.503} & \textbf{.591} & 7.1 & 0.7 \\
PW-2B & .519 & .648 & \textbf{.582} & \textbf{.565} & .417 & .575 & 6.5 & 0.6 \\
LW-2B & .512 & .627 & .548 & .545 & .434 & .575 & 7.0 & 0.7 \\
\midrule
S3 & \textbf{.711} & \textbf{.808} & \textbf{.700} & \textbf{.702} & \textbf{.749} & .782 & 92.1 & 8.9 \\
S5 & \textbf{.710} & \textbf{.808} & \textbf{.705} & \textbf{.701} & \textbf{.749} & \textbf{.801} & 90.7 & 8.7 \\
S6 & \textbf{.708} & -- & \textbf{.704} & -- & -- & -- & 71.3 & 6.9 \\
\bottomrule
\end{tabular}
\caption{Quality and cost per query (expanded labels). nDCG@10 over all queries ($n=350$ nested, 372 near-miss) and at $k=4$; R@1 over all queries. \#tok: input tokens per query (thousands), averaged over all 722 queries; for the frame systems the description rerank S4@50 (10.4k) plus the frame step. $\times$S4: total over S4@50's total. Bold is best per block or less than 0.01 below it. For the cross-encoders, which rerank S0's top 50 descriptions, \#tok counts the tokens they read; LW-2B also writes about 0.4k. S6 sends 77\% of queries to frames (S4@50 top-two margin $\le 2$); its $k=4$ and R@1 cells are not computed. Strict labels: Table~\ref{tab:results-strict}.}
\label{tab:results}
\end{table}
}{%
\begin{table}[t]\centering\fbox{\parbox[c][2.5cm][c]{0.92\columnwidth}{\centering Table pending (tables/table1.tex)}}\caption{Quality and cost per query (pending).}\label{tab:results}\end{table}}

\newcounter{finding}
\newcommand{\finding}{\refstepcounter{finding}(\thefinding)}

\subsection{Key findings}
\label{sec:res-findings}

\noindent\textbf{\finding\label{sec:res-rq1} Splitting hurts search; whole-query judging reranks better.} Over descriptions, searching once per constraint (S1) does worse than one whole-query search (S0). The S1--S0 difference is small at $k=1$, with 95\% intervals including zero, and negative at $k=2$--4 on both sets; none of five stronger fusions beats S0 (best: Table~\ref{tab:ablations-extra}). Reranking S0's top 50 descriptions shows the same pattern. With the same verifier, asking once whether a video fits the whole query (S4@50) beats grading each constraint and averaging the grades (S2@50) by $+0.101$ $[+0.069, +0.136]$ nDCG@10 on nested queries and $+0.094$ on near-miss queries, and it costs less. Both rerank the same 50 candidates; in a post hoc split, about $+0.04$ to $+0.05$ of the lead comes from which ten of the 50 each reranker lifts into its top ten and the rest from how it orders them, and both parts stay above zero, also under strict labels. The S2 and S4 prompts differ in how they frame the task, in how they treat a constraint the description does not mention (S2 grades it 5), and in how they combine scores. We do not separate the first two, but none of five re-aggregations of S2's grades closes the gap (Appendix~\ref{app:results}, Table~\ref{tab:reagg}), so the averaging rule alone does not explain it, and on frames, where the prompts differ in the same ways, the two checks tie (Finding~\ref{sec:res-mechanism}).

\smallskip\noindent\textbf{\finding\label{sec:res-mechanism} Grading each constraint loses only on descriptions, which often leave out what a query asks about.} On frames, the whole-query (S5) and per-constraint (S3) checks tie: they fall within the pre-set $\pm0.02$ margin on both sets and under both label sets, and both come within about 0.01 of the best possible reordering of their pool under expanded labels. S3 lifts nDCG@10 over S4@50 ($.711$ vs $.636$ nested, $.700$ vs $.626$ near-miss, expanded labels) but at 6.0 times its dollar cost (US\$0.207 vs 0.034 per query over all 722 test queries, at list prices), about what S5 costs, so grading each constraint buys nothing on frames. We report no wall-clock latency, which depends on provider load; S5 adds only one stage of ten parallel frame calls after S4, and the dollar figures are in Appendix~\ref{app:stats}. On descriptions, the evidence is consistent with missing evidence. An annotator found that a source's description states about a third of its facts, only implies 15\% and leaves out about half (Appendix~\ref{app:human}). S2 grades an implied fact 6.4 out of 10 on average, against 9.2 for a stated one (on queries whose split matches their $k$ constraints). Among queries with no fact left out, S4's lead is larger when the description only implies some fact ($+0.105$, $n=67$) than when it states every fact ($+0.047$, $n=101$), though the intervals overlap. Silence is common too: 44\% of S2's description grades are the 5 reserved for an unmentioned constraint, against 4\% on frames, and scoring those grades as zero or dropping them (post hoc re-aggregations) makes S2 worse, not better. This fits the per-constraint check having little to go on when a description is silent, however that silence is scored, though framing may also contribute (Finding~\ref{sec:res-rq1}).

\begin{figure}[!tb]
\centering
\IfFileExists{Figures/fig_trigger.pdf}{%
\includegraphics[width=\columnwidth]{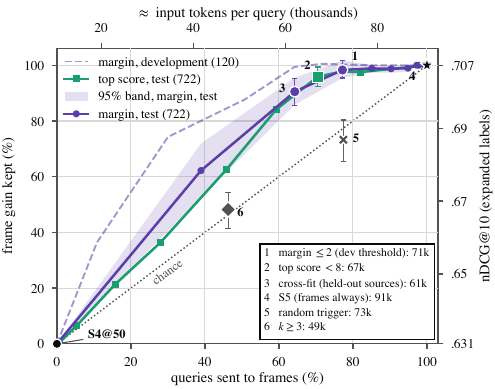}}{%
\fbox{\parbox[c][3.2cm][c]{0.92\columnwidth}{\centering Figure pending (Figures/fig\_trigger.pdf)}}}
\caption{Share of S5's gain over S4@50 kept against queries sent to frames (722 test queries) as the threshold of a trigger on S4@50's top-two margin or top score varies; dashed: development queries (120); dotted: random trigger. How to read: every curve runs from no queries sent (bottom left) to all queries sent and the full gain kept (top right); the further a curve sits above the dotted line, the better its trigger picks the queries frames help. Right axis: the same in nDCG@10 (exact). Top axis: approximate input tokens per query (10.4k plus 80.3k per query sent). Band and bars: 95\% cluster-bootstrap intervals (can exceed 100\%). Key: exact input tokens per query for each numbered point. Cross-fit: threshold chosen on held-out sources (five folds, target 80\% kept).}
\label{fig:trigger}
\end{figure}

\smallskip\noindent\textbf{\finding\label{sec:res-local} The best 2B cross-encoder recovers about three quarters of the LLM reranker's gain.} Of the three off-the-shelf cross-encoders we ran zero-shot on S0's top 50 descriptions, Qwen3-VL-Reranker-2B (Qwen-2B) gains $+0.170$ $[+0.124, +0.217]$ nDCG@10 over S0 on nested queries and $+0.136$ on near-miss ones, against $+0.23$ and $+0.18$ for the LLM reranker S4@50, so roughly three quarters of the gain on both sets (a ratio of the two gains, with no interval; Table~\ref{tab:local} gives intervals on both). It reads 7.1k input tokens per query (its own tokenizer) against S4@50's 10.4k. Consistent with Finding~\ref{sec:res-rq1}, it also ranks above S2, the LLM verifier that grades each constraint separately, by $+0.042$ $[+0.005, +0.081]$ and $+0.055$ $[+0.012, +0.101]$ (also under strict labels: $+0.048$ and $+0.078$). The two differ in model, prompt and scoring scale as well as in whether the query is split, so this gap is not due to splitting alone. S4@50 still leads Qwen-2B by $+0.06$ $[+0.035, +0.084]$ on nested and $+0.04$ $[+0.014, +0.066]$ on near-miss queries, but under strict labels its near-miss lead shrinks to $+0.011$ $[-0.017, +0.039]$, an interval that includes zero: what the LLM adds is clear on nested queries, while on near-miss ones it is small and, under strict labels, not separable from zero. The other two, LightOn's pointwise PW-2B and list-wise LW-2B, gain less on nested queries ($+0.11$ and $+0.10$); PW-2B is close to Qwen-2B on near-miss queries, and the list-wise model does no better than the pointwise one.

\smallskip\noindent\textbf{\finding\label{sec:res-frames} Reading the frames of the top ten adds about 0.08 nDCG@10 and nearly removes near-miss confusions within those ten.} Reranking S4@50's top ten on frames (S5) raises nDCG@10 by $+0.076$ $[+0.058, +0.096]$ over all 722 queries, at 8.7 times the input tokens (US\$0.198 against US\$0.034 per query at list prices; Appendix~\ref{app:stats}), and the gain holds under strict labels. Finding~\ref{sec:res-trigger} gives the quality--cost curve. Frames can only reorder the ten videos the description reranker puts on top, so that first stage caps the gain: the source is in S4@50's top 10 for 65.2\% of queries, and their gain makes up 88\% of the total gain. The other queries gain only $+0.02$ to $+0.03$ (Appendix~\ref{app:results}). S5 gets within 0.01 of the best possible ordering of those ten (expanded labels; 0.04 under strict). The gain is similar on nested and near-miss queries ($+0.074$ and $+0.079$) and largest where a near miss competes with the source. On the 96 contested queries (a post hoc subgroup), whose source and hard negative both reach the top ten, frames add $+0.134$, and the hard negative outranks the source for 1\% of them after S5 against 21\% after S4@50.

The gain has two limits. First, we picked S4@50's top ten as the frame pool after the runs. The pool we had planned, S0's top ten, shares on average only 5.2 videos with it, and there frames gain about half as much ($+0.039$ over S4@10 on the same videos). Second, on the same ten candidates, 195-word descriptions recover about two thirds of the frame gain. Frames still lead them on S4@50's pool ($+0.027$) and are within 0.02 on S0's, but they read about twenty times as much input per query, which is the price of that last margin.

\smallskip\noindent\textbf{\finding\label{sec:res-length} We detect no trend in frame gain with video length or constraint count, but length raises its cost.} In an exploratory, post hoc analysis within the Vript part of the corpus (videos up to 4\,min), the frame gain changes by $-0.011$ $[-0.043, +0.021]$ per doubling of duration. We cannot rule out slopes up to about $+0.02$ per doubling, and length is confounded with platform, scene density and caption length. Each added constraint changes the gain by $-0.003$ $[-0.013, +0.008]$ (leaving out near-miss queries at $k=1$, which have no near miss to resolve). The cost does grow: over the 507 Vript test queries, the frame step reads about 71k input tokens per query for clips under 30\,s and 105k for videos over a minute, with wide spread within each bin (Appendix~\ref{app:stats}). These analyses do not establish length or $k$ as reliable signals for targeting queries that benefit from frames. We next test a signal that might: the description reranker's own score margin.

\smallskip\noindent\textbf{\finding\label{sec:res-trigger} A close-call trigger keeps most of the gain but saves little.} The trigger S6 sends a query to frames when S4@50's top two scores differ by at most 2 points, a threshold set on development queries. It keeps 98.3\% of S5's gain but still sends 77.1\% of queries to frames (Figure~\ref{fig:trigger}). With the threshold instead chosen by five-fold cross-fitting over the test sources (each fold uses a threshold set on the other folds' sources to keep 80\% of the gain; Appendix~\ref{app:results}), it keeps 90.5\% and sends 64.3\%. This is a cheaper operating point, not a like-for-like drop, and on test data the margin separates the queries frames help less well. Since the trigger saves only a fifth to a third of input tokens, we report it as a quality--cost curve rather than as a way to save cost. Other triggers do no better: sending queries with three or more constraints keeps about what a random trigger keeps, and thresholding S4@50's top score works about as well as the margin.

\smallskip\noindent\textbf{\finding\label{sec:res-labels} Humans find almost every query's facts on its source, but the LLM judges reject many true sources at nested $k=4$.} One annotator finds every fact of a query on its source for 706 of the 715 test queries it can decide (98.7\%). The LLM judges accept the source less often as nested queries grow: at $k=4$ they accept it 51\% of the time, where humans accept 97\%, as per-fact misses (16\%, mostly one judge's ``no'', and 26--31\% on videos over a minute) compound; in a re-run, 64 frames instead of 16 raise it from 48\% to 64\% (Appendix~\ref{app:results}), so effect sizes at $k=4$ are the least certain. Off the source, where the judges decide relevance, expanded labels agree with humans on the videos humans checked, at precision 0.83 and recall 0.82 (Table~\ref{tab:human-agree-k}); at $k\ge3$ too few videos off the source are relevant (20) to tell whether this falls with $k$. Strict labels give the same system order, and dropping queries with a judge-rejected source changes no overall conclusion (Appendix~\ref{app:results}).

A separate human check backs the frame result. Where S4@50 and S5 put different videos first, the annotator judges S5's top video to meet the query for 66\% of queries and S4@50's for 18\%. Most of this is S5 recovering the source: 87 of its 115 wins are the source. Leaving out every query where either top video is the source, S5 still leads by $+0.20$ [$+0.11$, $+0.29$], winning 28 of 30 decisive queries (Table~\ref{tab:human-top1}). Covering only queries where the rerankers disagree, it supports the judge-based nDCG@10 rather than replacing it.

\begin{table*}[!t]
\centering
\small
\setlength{\tabcolsep}{3pt}
\renewcommand{\arraystretch}{1.08}
\begin{tabular}{@{}llllcc@{}}
\toprule
 & Ablation & Contrast & $n$ & Nested & Near-miss \\
\midrule
(a) & Strict labels & S4@50 $-$ S2@50 & 350 / 372 & $+0.097$ {\footnotesize[$+0.065$, $+0.131$]} & $+0.088$ {\footnotesize[$+0.051$, $+0.129$]} \\
 & Strict labels & S5 $-$ S4@50 & 350 / 372 & $+0.062$ {\footnotesize[$+0.036$, $+0.092$]} & $+0.070$ {\footnotesize[$+0.045$, $+0.096$]} \\
 & Source accepted by both judges & S4@50 $-$ S2b@50 & 243 / 334 & $+0.169$ {\footnotesize[$+0.128$, $+0.213$]} & $+0.127$ {\footnotesize[$+0.095$, $+0.160$]} \\
 & Source accepted by both judges$^s$ & S5 $-$ S4@50 & 38 / 88 & $+0.025$ {\footnotesize[$+0.005$, $+0.051$]} & $+0.123$ {\footnotesize[$+0.066$, $+0.189$]} \\
\midrule
(b) & Grading, descriptions & S2@50 $-$ S2b@50 & 350 / 372 & $+0.085$ {\footnotesize[$+0.051$, $+0.121$]} & $+0.031$ {\footnotesize[$-0.001$, $+0.061$]} \\
 & Whole-query against binary & S4@50 $-$ S2b@50 & 350 / 372 & $+0.187$ {\footnotesize[$+0.149$, $+0.226$]} & $+0.125$ {\footnotesize[$+0.096$, $+0.156$]} \\
 & Grading, frames$^s$ & S3 $-$ S3b & 56 / 96 & $+0.002$ {\footnotesize[$-0.000$, $+0.005$]} & $+0.007$ {\footnotesize[$-0.011$, $+0.032$]} \\
 & Binary vs.\ whole-query, frames$^s$ & S3b $-$ S5 & 56 / 96 & $-0.005$ {\footnotesize[$-0.013$, $+0.000$]} & $-0.020$ {\footnotesize[$-0.053$, $+0.003$]} \\
 & Minimum instead of mean grade & S2@50 (min) $-$ S2@50 & 350 / 372 & $-0.038$ {\footnotesize[$-0.067$, $-0.011$]} & $+0.012$ {\footnotesize[$-0.015$, $+0.041$]} \\
\midrule
(c) & S4@50's top 10 & S4L $-$ S4 & 350 / 372 & $+0.050$ {\footnotesize[$+0.029$, $+0.073$]} & $+0.050$ {\footnotesize[$+0.028$, $+0.074$]} \\
 & S4@50's top 10 & S5 $-$ S4L & 350 / 372 & $+0.024$ {\footnotesize[$+0.014$, $+0.035$]} & $+0.029$ {\footnotesize[$+0.016$, $+0.043$]} \\
 & S0's top 10 & S4L $-$ S4@10 & 350 / 372 & $+0.030$ {\footnotesize[$+0.017$, $+0.043$]} & $+0.027$ {\footnotesize[$+0.012$, $+0.045$]} \\
 & S0's top 10 & S5 $-$ S4L & 350 / 372 & $+0.007$ {\footnotesize[$+0.000$, $+0.013$]} & $+0.015$ {\footnotesize[$+0.006$, $+0.026$]} \\
\midrule
(d) & \texttt{gpt-5.5}, binary & S4@50 $-$ S2b@50 & 52 / 64 & $+0.230$ {\footnotesize[$+0.145$, $+0.315$]} & $+0.116$ {\footnotesize[$+0.058$, $+0.179$]} \\
 & \texttt{gpt-5.5} vs.\ \texttt{gpt-6-sol} & S2@50 $-$ S2@50 & 150 / 150 & $-0.025$ {\footnotesize[$-0.055$, $+0.002$]} & $-0.011$ {\footnotesize[$-0.033$, $+0.012$]} \\
 & \texttt{gpt-5.5}, graded (too few) & S4@50 $-$ S2@50 & 21 / 25 & $+0.106$ {\footnotesize[$-0.005$, $+0.228$]} & $+0.061$ {\footnotesize[$-0.022$, $+0.143$]} \\
 & \texttt{gemini-3.8-flash} on frames$^{s,a}$ & S5 $-$ S4@50 & 56 / 96 & $+0.043$ {\footnotesize[$+0.016$, $+0.073$]} & $+0.133$ {\footnotesize[$+0.076$, $+0.197$]} \\
\midrule
(e) & Depth 10 & S4@10 $-$ S2@10 & 350 / 372 & $+0.043$ {\footnotesize[$+0.023$, $+0.064$]} & $+0.041$ {\footnotesize[$+0.022$, $+0.061$]} \\
 & Depth 50 against 10 & S4@50 $-$ S4@10 & 350 / 372 & $+0.107$ {\footnotesize[$+0.073$, $+0.146$]} & $+0.098$ {\footnotesize[$+0.071$, $+0.126$]} \\
 & Same ten candidates$^s$ & S4@50 $-$ S2b$'$ & 56 / 96 & $+0.034$ {\footnotesize[$-0.002$, $+0.075$]} & $+0.090$ {\footnotesize[$+0.051$, $+0.133$]} \\
\midrule
(f) & First stage only & S0$_\mathrm{L}$ $-$ S0 & 350 / 372 & $+0.044$ {\footnotesize[$+0.005$, $+0.085$]} & $+0.015$ {\footnotesize[$-0.014$, $+0.045$]} \\
 & After frame rerank & S5$_\mathrm{L}$ $-$ S5 & 350 / 372 & $+0.021$ {\footnotesize[$-0.017$, $+0.059$]} & $+0.005$ {\footnotesize[$-0.026$, $+0.036$]} \\
 & Splitting at retrieval & S1$_\mathrm{L}$ $-$ S0$_\mathrm{L}$ & 350 / 372 & $-0.091$ {\footnotesize[$-0.131$, $-0.054$]} & $-0.132$ {\footnotesize[$-0.175$, $-0.089$]} \\
 & Whole-query lead & S4@50$_\mathrm{L}$ $-$ S2@50$_\mathrm{L}$ & 350 / 372 & $+0.097$ {\footnotesize[$+0.064$, $+0.134$]} & $+0.090$ {\footnotesize[$+0.052$, $+0.130$]} \\
 & Frame gain & S5$_\mathrm{L}$ $-$ S4@50$_\mathrm{L}$ & 350 / 372 & $+0.067$ {\footnotesize[$+0.041$, $+0.094$]} & $+0.067$ {\footnotesize[$+0.045$, $+0.090$]} \\
 & BM25: splitting & S1$_\mathrm{B}$ $-$ S0$_\mathrm{B}$ & 350 / 372 & $-0.087$ {\footnotesize[$-0.126$, $-0.050$]} & $-0.094$ {\footnotesize[$-0.131$, $-0.060$]} \\
 & BM25: whole-query lead & S4@50$_\mathrm{B}$ $-$ S2@50$_\mathrm{B}$ & 350 / 372 & $+0.065$ {\footnotesize[$+0.038$, $+0.093$]} & $+0.061$ {\footnotesize[$+0.030$, $+0.094$]} \\
 & BM25: frame gain & S5$_\mathrm{B}$ $-$ S4@50$_\mathrm{B}$ & 350 / 372 & $+0.048$ {\footnotesize[$+0.026$, $+0.072$]} & $+0.055$ {\footnotesize[$+0.034$, $+0.078$]} \\
\bottomrule
\end{tabular}
\caption{Ablations and robustness checks: paired differences in nDCG@10 with 95\% cluster-bootstrap intervals over source videos; expanded labels unless noted. (a) Labels. (b) Score scale: binary against graded checks. (c) Description length: about 195 against 44 words on the same ten candidates; S4L is S4 on the long descriptions. (d) Second verifiers. (e) Rerank depth and equal candidates. (f) First stage \texttt{text-embedding-3-large} or BM25 (strict labels); subscript L or B marks a system on that index. $^s$On the frame subgroups: the nested sample (Nested column) and the contested near-miss queries (Near-miss column). $^a$Called by an unpinned alias, so it may not reproduce; a supporting check only. Under \texttt{gpt-5.5} the graded contrast has too few queries to count as a replication. Details of every row: Appendices~\ref{app:results}, \ref{app:stats} and~\ref{app:length}.}
\label{tab:ablations}
\end{table*}

\subsection{Ablations and robustness}
\label{sec:res-ablations}

\noindent Each row block of Table~\ref{tab:ablations} changes one design choice (details in Appendices~\ref{app:results} and~\ref{app:length}). S2b and S3b are binary versions of S2 and S3 (yes, no or unclear per constraint, ranked by the number of yes answers); S3b ran only on the frame subgroups.
\par\noindent\emph{(a) Labels:} strict labels keep the whole-query lead and the frame gain, and keeping only queries whose source both judges accept moves no overall interval across zero.
\par\noindent\emph{(b) Score scale:} grading helps the per-constraint description check on nested queries but not clearly on near-miss ones, so about half of S4@50's lead over the binary check on nested queries comes from the score scale. On frames, grading changes little on the contested queries, with intervals that include zero.
\par\noindent\emph{(c) Description length:} on the same ten candidates, 195-word descriptions raise S4 by $+0.050$ on S4@50's top ten and $+0.028$ on S0's (Finding~\ref{sec:res-frames}). Longer descriptions also contradict more of a query's facts (a text-only judge marks a mined fact as contradicted when the description says something different about a detail that is there), but they state more too, so the ratio stays at about 0.6 contradicted per stated fact at every length. This ratio is descriptive: we did not test whether the extra stated facts themselves are what improves retrieval. The gain is positive for both query types: on the development set, S4 gains $+0.054$ on nested and $+0.068$ on near-miss queries at 202 words, and S2b's gain by number of constraints is in Appendix~\ref{app:length}.
\par\noindent\emph{(d) Second verifiers:} \texttt{gpt-5.5} reruns S2b and S4 on a seeded subset of 52 nested and 64 near-miss test queries and reproduces S4@50's lead over the binary S2b@50. Only 21 nested and 25 near-miss queries also have its graded S2@50 run, too few to test the graded contrast. \texttt{gemini-3.8-flash} regrades S5's candidates on 96 contested and 56 nested-sample queries and gives the same frame gain, but it is a Gemini model like the describer, so we treat it as a supporting check (Appendix~\ref{app:results}). Both checks cover subsets, so they support the S4 over S2b ordering and the frame gain across verifiers, not the graded contrast.
\par\noindent\emph{(e) Depth and candidates:} the whole-query lead holds at depth 10 and when S2b's stored verdicts rerank S4@50's own top ten (S2b$'$), though on the nested sample that interval just includes zero.
\par\noindent\emph{(f) First stage:} frames cannot recover a video outside the candidate pool, so we rerun every system, with strict labels, on two other first stages over the same descriptions: the stronger \texttt{text-embedding-3-large} and the weaker, lexical BM25 (Appendix~\ref{app:results}). The larger embedder barely changes which sources reach S5's pool, and every main contrast stays within 0.02 of its original value. BM25 lowers source recall@50 from 0.70 to 0.60, yet every contrast keeps its sign: over all 722 queries the whole-query lead falls from $+0.093$ to $+0.063$ and the frame gain from $+0.066$ to $+0.052$.
\par\noindent Stronger fusion, exact and gold splits, a repeat run with a fresh cache, dropping near-miss queries at $k=1$, and splits by fact type or corpus part change no conclusion either.

\section{Conclusion}
\label{sec:conclusion}

We asked whether multi-constraint video search over short descriptions is better served by splitting the query or by richer evidence, and built a benchmark of human-checked queries with near-miss videos to test both.
Splitting the query did not help search, our per-constraint grading reranked worse, and judging the whole query helped, whether with an LLM or the best of three small cross-encoders.
Richer evidence also helped: reading the frames of the top candidates improved ranking at a high token cost; longer descriptions recover much of that gain, but must be written for every video in the index.
Our findings suggest that, with the prompts, scoring and verifier we tested, which differ in framing as well as in decomposition, the limit for description-indexed video search lies more in what a description records than in how the query is decomposed, and that the open problem is deciding when a candidate is worth reading in full.
We release the dataset and code.

\section{Limitations}
\label{sec:limitations}

We did not test action, spatial, event or temporal constraints (Appendix~\ref{app:mining}); frames on human-written queries; videos over 4\,minutes or queries with more than four constraints; segment- or shot-level indexing, direct video embeddings or fused multi-index systems; descriptions from other describers against frames; learned or query-adaptive fusion; and per-constraint prompts other than ours, including other missing-evidence anchors. We also did not separate why descriptions omit detail (short length, a describer that misses fine detail even at length, or query-agnostic writing); only length is partly measured (Appendix~\ref{app:length}).

Queries are LLM-written from dataset captions that are partly model-written (Appendix~\ref{app:corpus}), so they are likely more explicit than user queries.
Real queries may leave constraints implicit or ambiguous, which could hurt splitting (it must infer them) or help it relative to whole-query judging. Our only human-written queries, 400 MultiVENT~2.0 news headlines run on descriptions only, mostly name entities, places and events rather than visual facts; S4 still leads S2b there (Appendix~\ref{app:results}), but they test neither our query style nor splitting under implicit or ambiguous constraints.
Every constraint was confirmed on frames by two VLMs, which could build in a frame advantage that neither strict labels nor the human check can rule out. The grounding step keeps facts that descriptions state more often, the fact-type filter does not, and the net effect is unclear (Appendix~\ref{app:mining}).

The human check covers little of the top 10; its top-1 contrast covers only disagreements, mostly source recovery.
Humans' 97\% at nested $k=4$ is 89\% with borderline facts absent.

One verifier (\texttt{gpt-6-sol}) drives the main results, so they may be specific to it (second verifiers: Section~\ref{sec:res-ablations}d).
Model families overlap across roles: Gemini describes, grounds and judges, and OpenAI models ground, judge and verify (Appendix~\ref{app:systems}).
The three 2B cross-encoders ran zero-shot on descriptions, once each, untuned; added post hoc, with three of five models reported (Appendix~\ref{app:results}), they are exploratory.

We chose the frame pool after the runs, on the same queries (Finding~\ref{sec:res-frames}).
The contested subgroup, the grade re-aggregations and the length and $k$ slopes are post hoc and exploratory: chosen after seeing the results, they may depend on those choices and should be read with caution.

\bibliography{refs}

\appendix

\section{Systems, Models and Prompts}
\label{app:systems}

\paragraph{Models.}
The describer is Gemini~3.5 Flash-Lite (\texttt{gemini-3.5-flash-lite}) at default media resolution; the embedder is \texttt{text-embedding-3-small}. The splitter, fact miner, near-miss miner and query writer are \texttt{gpt-6-luna} at low reasoning effort. The verifier for S2, S2b, S3, S3b, S4 and S5 is \texttt{gpt-6-sol} at low effort; the second verifier for S2, S2b and S4 is \texttt{gpt-5.5} at low effort. A second frame verifier for S5, reported separately, is \texttt{gemini-3.8-flash} at low thinking effort and medium media resolution, on the same prompt, frames and top-10 S4@50 candidates as \texttt{gpt-6-sol}. Grounding and the frame judge use Gemini~3.5 Flash-Lite at low media resolution together with \texttt{gpt-6-astra}. The main verifiers (\texttt{gpt-6-sol}, \texttt{gpt-5.5}) are never a grounder or a judge. \texttt{gemini-3.8-flash} is also a Gemini model, which adds to the family overlap noted in Section~\ref{sec:limitations}. Closed models are pinned by version and date, except \texttt{gemini-3.8-flash}, used by alias (Table~\ref{tab:models}). The 0.5\,fps index frames number at least 8 per video, and the index-description prompt asks the describer to include visible on-screen text.

\begin{table}[t]\centering\small\setlength{\tabcolsep}{2pt}
\begin{tabular}{@{}p{0.90in}p{0.88in}p{0.75in}p{0.60in}@{}}
\toprule
Model & Role & Snapshot & Setting \tabularnewline
\midrule
\texttt{gemini-3.5-}\newline\texttt{flash-lite} & \raggedright describer; grounder; judge & pinned & \raggedright media res.\ default (describe), low (ground, judge) \tabularnewline
\texttt{text-embed-}\newline\texttt{ding-3-small} & embedder & pinned & -- \tabularnewline
\texttt{gpt-6-luna} & \raggedright splitter; fact and near-miss miner; query writer & pinned & low effort \tabularnewline
\texttt{gpt-6-astra} & grounder; judge & pinned & -- \tabularnewline
\texttt{gpt-6-sol} & \raggedright verifier, S2, S2b, S3, S3b, S4, S5 & pinned & low effort \tabularnewline
\texttt{gpt-5.5} & \raggedright second verifier, S2, S2b and S4 & \texttt{gpt-5.5-}\newline\texttt{2026-04-23} & low effort \tabularnewline
\texttt{gemini-3.8-}\newline\texttt{flash} & \raggedright second frame verifier, S5 & \raggedright alias, unpinned & \raggedright low thinking; media res.\ medium \tabularnewline
\bottomrule
\end{tabular}
\caption{Models. Snapshot: \emph{pinned} models are called by a version and date recorded in the release (Appendix~\ref{app:release}); the dated id is shown where the API returned one (\texttt{gpt-5.5}); \texttt{gemini-3.8-flash} is called by alias, so no dated snapshot was recorded. Setting: reasoning effort for OpenAI models; media resolution for Gemini.}
\label{tab:models}
\end{table}

\paragraph{Query parsing.}
S1, S2 and S3, and the binary S2b and S3b, use one splitter call per query (the same split for all); systems never see the mined facts. On the development set, the splitter-check prompt checks how well the split covers the mined facts.

\paragraph{S1.}
The reciprocal rank fusion constant is 60.

\paragraph{S2 and S3 (graded per-constraint).}
For each candidate, the verifier grades each split constraint on the 0--10 scale of Section~\ref{sec:method}, from the description (S2) or the frames (S3), with the same strictness clause about quoted text, numbers and colours as the whole-query rerank prompt. Besides the mean grade, the minimum over constraints is reported as a secondary aggregation (Appendix~\ref{app:results}). An unparsable answer scores $-1$, like a failed S4 call, and a constraint the answer omits counts as 5. S3 sees the candidate's own 0.5\,fps index frames, with no cap, each labelled with its index and timestamp. S2 is rescored at depth 10 from the same calls (S2@10).

\paragraph{S2b and S3b (binary per-constraint; score-scale ablation).}
The same inputs and split as S2 and S3, but one call per candidate answers yes, no or unclear per constraint, with a short reason. The score is the number of yes answers, ties broken by S0 rank; candidates below the reranked depth keep their S0 order. S2b reranks S0's top 50 and S3b the top 10 of S4@50 (Section~\ref{sec:res-ablations}).

\paragraph{S2b example.}
For the nested $k=4$ query of Table~\ref{tab:examples}, the splitter returns four constraints: a man wears a black hair band; an explosion is visible; three men are present; a microphone is visible on the stage. S2b reads each of S0's top 50 descriptions once and answers yes, no or unclear per constraint. The source's description gets unclear (the band), yes, yes, yes, and so scores 3; the best of the other 49 candidates score 1 (for instance unclear, unclear, yes, unclear), so the source is ranked first.

\paragraph{S4 and S5.}
The same verifier, with the same inputs per candidate as S2 or S3, scores the candidate on its own, against the whole query, on a 0--10 scale in one call; ties are broken by S0 rank. They share the verifier, scale and candidates with S2 and S3 but differ in framing (the undecomposed query), in the missing-evidence anchor and in aggregation (Section~\ref{sec:method}); the binary S2b and S3b also change the score scale. The depth-10 variants S4@10 and S2b@10 are rescored from the same calls, to measure the effect of depth.

\paragraph{Frame systems.}
S3, S5 and S6 run on all 722 test queries and rerank the top 10 of S4@50. The pool was planned as S0's top 10; the switch to S4@50's top 10 is stated in the Limitations section.

\paragraph{S6.}
S6 applies the 2-point margin threshold of Section~\ref{sec:method} (chosen on the development set) unchanged to the test set, offline, from the saved S4 and S5 outputs, so it adds no model calls; its cost is that of S4 plus S5 on the queries it sends to frames. We report the share of the frame gain kept against the share of queries sent as the threshold varies (Figure~\ref{fig:trigger}), and three baselines: send iff $k\ge3$, which needs no score; send iff S4's top score is below 8, which reads the same S4 scores as the margin; and random at the same share.

\paragraph{Second verifier.}
\texttt{gpt-5.5} repeats S2b and S4 on a seeded subset of 52 nested and 64 near-miss test queries with the same S0 rankings and splits. We report the S4 minus S2b difference per verifier and Cohen's $\kappa$ between the verifiers' S2b answers (.77 on 8,750 nested and .82 on 8,000 near-miss constraint verdicts). It also repeats the graded S2@50 on a seeded subset of 150 queries per set (seed 0 inside the 100 test sources), with the same S0 rankings and splits; 21 nested and 25 near-miss of these queries fall in the 52 / 64 subset, so the graded S4@50 $-$ S2@50 contrast with both systems under \texttt{gpt-5.5} has $n=21$ / 25 (Appendix~\ref{app:results}). Per query, the two verifiers' S4@50 scores over the candidates both scored correlate at Spearman .787 $[.746, .823]$ on the 52 nested and .837 $[.805, .867]$ on the 64 near-miss queries (mean over queries, cluster bootstrap over sources), and their reranked top-1 video is the same for .731 $[.623, .843]$ and .828 $[.729, .918]$ of the queries.

\subsection{Prompts}
\label{app:prompts}

Each prompt is sent as one user message, with no system message; placeholders in braces are filled at run time, and frames are sent as images, each preceded by its label. Prompt texts are shown verbatim. Variants that differ from a shown prompt only in a phrase are noted in the text. All prompts are released (Appendix~\ref{app:release}).
\lstdefinestyle{prompt}{basicstyle=\scriptsize\ttfamily,numbers=none,xleftmargin=0pt,breaklines=true,breakatwhitespace=true,frame=none,aboveskip=4pt,belowskip=3pt,columns=fullflexible,keepspaces=false,showstringspaces=false}

\begin{promptbox}{Index descriptions (all systems).}
The describer receives the frames with this prompt. The longer-description variants (Appendix~\ref{app:length}) replace ``1 to 3 sentences'' by ``3 to 5'' and ``5 to 7''; the longest replaces the second paragraph by the one below.
\begin{lstlisting}[style=prompt]
These are {n} frames sampled at equal intervals, in order, from one video that may last from a few seconds to several minutes, so the time between two frames can be long and the scene may change completely between them.

Write 1 to 3 sentences of natural prose that describe only the genuinely distinct visual content and what would make the video findable: the on-screen activity or format (a skincare step, a wheelie, an ingredient graphic, a workout set, an unboxing), the products and objects shown and their category, the setting or location, notable on-screen text or graphics (quoted exactly), and the look of the video.

Describe what a viewer sees, not what anyone says: do not guess or restate speech, and never write "discusses", "explains" or "critiques". Do not narrate delivery ("gestures", "speaks to camera"). Name a brand, product or venue only when its name or logo is visible in these frames; describe incidental background generically. If the frames show nothing beyond a person talking to the camera, write one short sentence saying so, with the setting and anything clearly visible (for example "A woman talks to the camera in a bedroom with a ring light.").
\end{lstlisting}
\tcbline
\begin{lstlisting}[style=prompt]
Write 14 to 18 sentences of natural prose that describe the visual content in as much concrete detail as the frames show, covering what would make the video findable: the on-screen activity or format (a skincare step, a wheelie, an ingredient graphic, a workout set, an unboxing), the products and objects shown and their category, the setting or location, notable on-screen text or graphics (quoted exactly), and the look of the video. Every sentence must state a visible fact that no earlier sentence states. Do not write summary, transition or atmosphere sentences, for example about pacing, rhythm, mood, or what the video maintains or features throughout.
\end{lstlisting}
\end{promptbox}

\begin{promptbox}{Fact mining.}
The type rule that selects facts afterwards is a fixed rule in code (Appendix~\ref{app:mining}), not a prompt.
\begin{lstlisting}[style=prompt]
You extract atomic, visually checkable facts from the caption of one video. The caption is split into numbered scenes.

A fact is ONE property of the video that a person could confirm by looking at frames of the video. Leave out audio, speech content, thoughts, intentions and narrator commentary. Split compound statements: "a man in a red shirt holds two cups" gives three facts (a man wears a red shirt; a man holds cups; there are two cups). Do not repeat a fact. Each fact must be self-contained and understandable on its own, without the caption or the other facts: name its entity fully by what it is and how it looks (e.g. "the woman in the red apron wears a green scarf", not "the woman on the left", "the other character", "she" or "the glass"). Never write "the video" or "the clip".

For each fact give:
- "text": the fact, under 20 words;
- "type": exactly one of
  - action: what someone or something does (e.g. "a man lifts a barbell")
  - object: presence of a specific object, animal or kind of person (e.g. "a ukulele is visible")
  - attribute: colour, clothing, hair, appearance or material of an entity
  - on_screen_text: text visible in the video (titles, signs, labels, logos with words); quote it
  - count: an explicit number of entities (e.g. "three dogs are present")
  - scene: setting or location (e.g. "an indoor gym", "a beach at night")
  - spatial_relation: where things are relative to each other
  - temporal_order: an order of events (e.g. "after mixing, she pours the batter")
  - other: anything else visual (shot type, editing, lighting)
- "scene": the number of the scene the fact comes from;
- "single_frame": true if one frame could show the fact (an object, an attribute, a sign), false if it needs several frames (an order of events, a change, a motion that cannot be seen in one still);
- "self_contained": true if a reader who has not seen the caption knows which entity the fact is about; false if it relies on position or earlier mentions ("the woman on the left", "the other one", "the same man");
- "distinctive": true if the fact would help single out this video among about 8,600 diverse web videos (a named colour on a specific item, quoted text, an exact count, an uncommon object); false if it is true of many videos (e.g. "a man is present", "the scene is indoors").

Return JSON only: {"facts": [{"text": str, "type": str, "scene": int, "single_frame": bool, "self_contained": bool, "distinctive": bool}, ...]}

Caption:
"""{caption}"""
\end{lstlisting}
\end{promptbox}

\begin{promptbox}{Near-miss facts.}
\begin{lstlisting}[style=prompt]
You compare the captions of two different videos, A and B. Each caption is split into numbered scenes. A caption marked "[...]" was shortened.

List two kinds of facts:
1. "shared": facts that hold for BOTH videos according to their captions. Write each one so that, as written, it is true of both videos. When the two captions differ in a detail, use the wording that fits both, or leave the fact out.
2. "distinct": facts that hold for video A but not for video B: either B's caption contradicts the fact (it describes the same kind of thing differently: another colour, text, number, object or place), or B's caption describes its people, objects and setting in enough detail that the fact would have been mentioned and it is not.

Rules for every fact:
- one property only, checkable by looking at a single frame;
- type is exactly one of: object (presence of a specific object, animal or kind of person), attribute (colour, clothing, hair, appearance or material of an entity), on_screen_text (text visible in the video, quoted), count (an explicit number of entities), scene (setting or location);
- no actions, speech, sounds, thoughts, order of events, camera work, or positions of things relative to each other;
- self-contained: name the entity by what it is and how it looks, never "the same", "the other", "the one on the left", "she" or "the video"; under 20 words.
For distinct facts, prefer on-screen text, objects and attributes, whose absence can be checked by looking.

For each fact give "text", "type" and "scene_a" (the scene number in A's caption it comes from). For shared facts also give "scene_b" (the scene number in B's caption). For distinct facts also give "in_b": "contradicted" or "absent", and "b_evidence": the shortest passage of B's caption (at most 15 words) that contradicts it or shows the detail it would have been mentioned in, or "".

Give at most 6 shared and at most 4 distinct facts, the most specific ones first. If the videos share nothing visual, return an empty "shared" list.

Answer with JSON only: {"shared": [{"text": str, "type": str, "scene_a": int, "scene_b": int}, ...], "distinct": [{"text": str, "type": str, "scene_a": int, "in_b": str, "b_evidence": str}, ...]}

Caption of video A:
"""{caption_a}"""

Caption of video B:
"""{caption_b}"""
\end{lstlisting}
\end{promptbox}

\begin{promptbox}{Fact verification by two vision-language models.}
Used for both Gemini and \texttt{gpt-6-astra}, on each fact\textquoteright s scene frames; the same prompt, run on all index frames, logs whether a fact is visible in the index.
\begin{lstlisting}[style=prompt]
These are {n} frames from one video, in time order. They may cover only one part of the video. Each frame is preceded by a label "Frame i of {n} (t = ... s)", where t is the time in the video. Below is a numbered list of facts. For each fact, decide whether these frames show it.

Answer for each fact:
- "yes" only if at least one frame clearly shows the fact. Name the frame or frames where it is visible and say briefly what you see there that shows it.
- "no" if the frames clearly do not show it: what the fact is about appears in no frame, or it appears in a way that does not fit (another colour, number, text, object or place).
- "unsure" if you cannot tell: the relevant part is too small, blurred, cut off or hidden, or the fact cannot be judged from still frames.

Look at each frame separately; do not assume that something seen in one frame is also in the others. Use only what is visible, not what is likely or typical. Quoted text may differ in case or punctuation.

Facts:
{facts}

Answer with JSON only: {"visible": [{"fact": 1, "answer": "yes" or "no" or "unsure", "frames": [up to 3 frame numbers, as integers, that show the fact most clearly] or [], "evidence": "under 15 words: what is seen in the first of those frames, or why the answer is no" or ""}, ...]} with one entry per fact, in order.
\end{lstlisting}
\end{promptbox}

\begin{promptbox}{Query writing (nested and near-miss).}
\begin{lstlisting}[style=prompt]
Write one search query that a person could type to find a video that shows all of the facts below. Rules:
- cover every fact and add nothing that is not in the facts;
- write as a person searching from memory would: reword freely, and you may describe a quoted text or an exact detail loosely (a sign reading "OPEN" can become "a shop with an open sign"), as long as the fact stays checkable;
- the facts may come from different moments of the video; do not imply that they happen together or that their subjects are next to each other (no "beside", "while", "with" joining facts from different moments) unless the facts say so; it is fine to write "at one point ... later ..." or to list them as things the video shows;
- write one fluent phrase or sentence with articles and prepositions, not a list of keywords; about {words} words; no questions, no "video of", no quotation marks.

Facts:
{facts}

Answer with JSON only: {"query": "...", "phrases": {"1": "the part of the query that carries fact 1", "2": "..."}}
\end{lstlisting}
\end{promptbox}

\begin{promptbox}{Splitter (S1, S2, S2b, S3, S3b).}
\begin{lstlisting}[style=prompt]
Below is a search query for a collection of videos. Split it into its separate constraints: the individual things a video must show to match the query. Each constraint is one short, self-contained statement that can be checked on its own (repeat the subject where needed, so "a man in a red cap holding a sign that reads OPEN" becomes "a man wears a red cap" and "a man holds a sign that reads OPEN"). Keep exact wording of any quoted or on-screen text, numbers and colours. Do not add anything the query does not say, and do not drop anything it does say. A query that asks for one thing gives one constraint.

Query: {query}

Answer with JSON only: {"constraints": ["...", "..."]}
\end{lstlisting}
\end{promptbox}

\begin{promptbox}{S2, graded per-constraint description check.}
\begin{lstlisting}[style=prompt]
You check whether a video matches a search query. You cannot see the video; you get a written description of it, made from frames sampled over the whole video. The query has been split into numbered constraints. For each constraint, give one integer score from 0 to 10 from the description alone:
- 10: the description clearly states it.
- 5: the description neither confirms nor contradicts it (not mentioned).
- 0: the description clearly contradicts it (for example a different colour, number or text, or a setting that rules it out).
Use the values in between for partial or uncertain evidence. Judge each constraint separately and strictly; a near match on quoted text, a number or a colour does not score high.

Query: {query}

Constraints:
{constraints}

Description of the video:
{description}

Answer with JSON only: {"constraints": [{"constraint": 1, "score": integer from 0 to 10}, ...]}
\end{lstlisting}
\end{promptbox}

\begin{promptbox}{S2b, binary per-constraint description check.}
\begin{lstlisting}[style=prompt]
You check whether a video matches a search query. You cannot see the video; you get a written description of it, made from frames sampled over the whole video. The query has been split into numbered constraints. For each constraint, decide from the description alone:
- "yes": the description states it or clearly implies it.
- "no": the description contradicts it (for example a different colour, number or text, or a setting that rules it out).
- "unclear": the description neither confirms nor contradicts it.
Judge each constraint separately and strictly; a near match on quoted text, a number or a colour is not a "yes".

Query: {query}

Constraints:
{constraints}

Description of the video:
{description}

Answer with JSON only: {"constraints": [{"constraint": 1, "answer": "yes" or "no" or "unclear"}, ...]}
\end{lstlisting}
\end{promptbox}

\begin{promptbox}{S3, graded per-constraint frame check.}
\begin{lstlisting}[style=prompt]
You check whether a video matches a search query. These are {n} frames sampled at equal intervals, in order, over the whole video; each frame is preceded by a label "Frame i of {n} (t = ... s)". The query has been split into numbered constraints. For each constraint, give one integer score from 0 to 10 from the frames:
- 10: at least one frame clearly shows it; name the frame(s).
- 5: you cannot tell from these frames (neither shown nor ruled out).
- 0: the frames show something that rules it out (for example a different colour, number or text), or the video clearly lacks it.
Use the values in between for partial or uncertain evidence. Judge each constraint separately and strictly; a near match on quoted text, a number or a colour does not score high. Look at each frame separately; do not assume that something seen in one frame is also in the others.

Query: {query}

Constraints:
{constraints}

Answer with JSON only: {"constraints": [{"constraint": 1, "score": integer from 0 to 10, "frames": [frame numbers that show it, as integers] or []}, ...]}
\end{lstlisting}
\end{promptbox}

\begin{promptbox}{S3b, binary per-constraint frame check.}
\begin{lstlisting}[style=prompt]
You check whether a video matches a search query. These are {n} frames sampled at equal intervals, in order, over the whole video; each frame is preceded by a label "Frame i of {n} (t = ... s)". The query has been split into numbered constraints. For each constraint, decide from the frames:
- "yes": at least one frame clearly shows it; name the frame(s).
- "no": the frames show something that rules it out (for example a different colour, number or text), or the video clearly lacks it.
- "unclear": you cannot tell from these frames.
Judge each constraint separately and strictly; a near match on quoted text, a number or a colour is not a "yes". Look at each frame separately; do not assume that something seen in one frame is also in the others.

Query: {query}

Constraints:
{constraints}

Answer with JSON only: {"constraints": [{"constraint": 1, "answer": "yes" or "no" or "unclear", "frames": [frame numbers that show it, as integers] or []}, ...]}
\end{lstlisting}
\end{promptbox}

\begin{promptbox}{S4 and S5, whole-query rerank.}
\texttt{\{evidence\}} is the first text below for S4 (description) and the second for S5 (frames).
\begin{lstlisting}[style=prompt]
You judge how well a video matches a search query. {evidence}

Query: {query}

Give one score from 0 to 10 for how well the whole video matches the whole query: 10 means it clearly matches every part of the query, 0 means it clearly does not match at all. A video that matches only part of the query scores in between. Be strict about quoted text, numbers and colours.

Answer with JSON only: {"score": integer from 0 to 10}
\end{lstlisting}
\tcbline
\begin{lstlisting}[style=prompt]
You cannot see the video; you get a written description of it, made from frames sampled over the whole video.

Description of the video:
{description}
\end{lstlisting}
\tcbline
\begin{lstlisting}[style=prompt]
These are {n} frames sampled at equal intervals, in order, over the whole video; each frame is preceded by a label "Frame i of {n} (t = ... s)".
\end{lstlisting}
\end{promptbox}

\begin{promptbox}{Frame relevance judge (expanded labels).}
Run with both judge models on at most 16 index frames per candidate.
\begin{lstlisting}[style=prompt]
These are {n} frames sampled at equal intervals, in time order, from one video. Each frame is preceded by a label "Frame i of {n} (t = ... s)", where t is the time in the video. Below is a numbered list of facts. For each fact, decide whether the video shows it, judging only from these frames. A fact holds if any frame shows it; the facts may come from different moments and need not appear together.

Answer for each fact:
- "yes" only if at least one frame clearly shows the fact. Name the frame or frames and say briefly what you see there that shows it.
- "no" if the frames clearly do not show it: what the fact is about appears in no frame, or it appears in a way that does not fit (another colour, number, text, object or place).
- "unsure" if you cannot tell: the relevant part is too small, blurred, cut off or hidden, or the fact cannot be judged from still frames.

Look at each frame separately; do not assume that something seen in one frame is also in the others. Use only what is visible, not what is likely or typical. Quoted text may differ in case or punctuation.

Facts:
{facts}

Answer with JSON only: {"facts": [{"fact": 1, "answer": "yes" or "no" or "unsure", "frames": [up to 3 frame numbers, as integers] or [], "evidence": "under 15 words: what is seen in the first of those frames, or why the answer is no" or ""}, ...]} with one entry per fact, in order.
\end{lstlisting}
\end{promptbox}

\begin{promptbox}{Text-only judge, \texttt{gpt-6-sol}.}
Marks each fact stated (``yes''), contradicted (``no'') or unclear; it gives the stated labels of the fact-type check (Appendix~\ref{app:mining}) and the description ablation.
\begin{lstlisting}[style=prompt]
Below is a written description (caption) of one video, possibly split into numbered scenes, and a numbered list of facts. Using only the caption, decide for each fact whether the video shows it.

Answer for each fact:
- "yes" if the caption states the fact or something that plainly implies it. Paraphrases and synonyms count; quoted on-screen text may differ in case, punctuation or a small spelling slip.
- "no" if the caption contradicts the fact: it describes the same kind of thing (the person, object, text, number or setting the fact is about) in a way that does not fit, for example a different colour, count, text or place.
- "unclear" if the caption does not mention what the fact is about, or mentions it too vaguely to decide. Do not answer "no" only because something is not mentioned.

Judge each fact on its own; a fact holds if any part of the video shows it. Use only the caption, not what is likely or typical. A caption marked "[...]" was shortened; do not treat the missing part as evidence either way.

For "yes" and "no", give a quote: the shortest passage of the caption, copied word for word (at most 20 words), that supports your answer. For "unclear", give the closest passage or "".

Caption:
{caption}

Facts:
{facts}

Answer with JSON only: {"facts": [{"fact": 1, "answer": "yes" or "no" or "unclear", "quote": "..."}, ...]} with one entry per fact, in order.
\end{lstlisting}
\end{promptbox}

\begin{promptbox}{Splitter check.}
\begin{lstlisting}[style=prompt]
A search query for a video was written from a list of facts about the video (the gold facts). An automatic splitter then split the query into constraints. Compare the two lists.

For each gold fact, say how the split constraints cover it:
- "full": one constraint, or several together, state the whole fact (wording may differ; exact text, numbers and colours must agree).
- "partial": the constraints state only part of the fact, or change a detail.
- "none": no constraint states it.
For each split constraint, give the number of the gold fact it states (or part of), or null when it states none of them (generic, added, temporal-only or about something else).

Gold facts:
{facts}

Split constraints:
{constraints}

Answer with JSON only: {"facts": [{"fact": 1, "match": "full" or "partial" or "none", "constraints": [constraint numbers]}, ...], "constraints": [{"constraint": 1, "fact": fact number or null}, ...]}
\end{lstlisting}
\end{promptbox}

\begin{promptbox}{Constraints per MultiVENT~2.0 query.}
\begin{lstlisting}[style=prompt]
Below is a search query for a collection of news and event videos. List its constraints: the atomic, checkable conditions a video must meet to match the query. A condition is one of: an entity (a named person, organisation, team, product or object), an event or action, an attribute of an entity or event (a quality, role, amount, outcome), a place, a date or time, or text that must appear on screen.

Rules:
- Each condition is one short, self-contained statement that can be checked on its own (repeat the subject where needed).
- Split conjunctions: "Putin and Zelensky meet in Istanbul" gives "Putin is present", "Zelensky is present", "they meet", "the meeting is in Istanbul".
- Do not add anything the query does not say, and do not drop anything it does say.
- Do not split a single name or fixed phrase ("2016 Summer Olympics" is one condition; "Event Horizon Telescope" is one entity).
- A query that asks for one thing gives one condition.

Query: {query}

Answer with JSON only: {"constraints": ["...", "..."]}
\end{lstlisting}
\end{promptbox}

\section{Corpus}
\label{app:corpus}

\begin{figure}[t]
\centering
\includegraphics[width=\columnwidth]{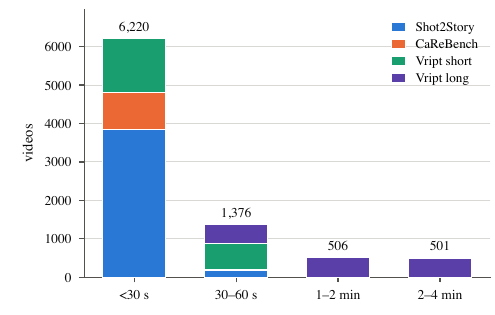}
\caption{The 8,603 corpus videos by length bin, stacked by corpus part; totals above the bars.}
\label{fig:corpus-lengths}
\end{figure}

\paragraph{Video lengths and captions.}
The corpus pools four parts (Section~\ref{sec:setup}); Table~\ref{tab:corpus-lengths} and Figure~\ref{fig:corpus-lengths} give their lengths. Captions are written by people for CaReBench, drafted per shot by a model and corrected by annotators for Shot2Story, and written by GPT-4V for Vript. We chose these datasets for caption detail: in a 50-video sample per dataset, an LLM reading the captions found at least four distinctive fine-detail facts for 44\% (Shot2Story), 76\% (CaReBench) and 82\% (Vript short) of videos, against 8\% for ActivityNet, which we dropped. CaReBench, the one part with human-written captions, shows the same system ordering as the rest: S4 leads S2b by $+0.086$ $[+0.021, +0.152]$ on nested and $+0.106$ $[+0.050, +0.171]$ on near-miss queries (Table~\ref{tab:perk}), though with 15 test sources this is a small check. Most videos are short: 6,220 of the 8,603 (72\%) are under 30\,s, and Shot2Story and CaReBench together supply 4,813 of them. Videos over one minute come almost entirely from Vript long (1,000 of 1,007), which is why length is analysed within Vript (Section~\ref{sec:setup}). Vript long is built to be uniform over 30--60\,s, 1--2\,min and 2--4\,min (500 spans each), and Vript short supplies the two shortest bins. The 70 Vript test sources used for the length analysis are 14 under 30\,s, 28 of 30--60\,s (14 short, 14 long), 15 of 1--2\,min and 13 of 2--4\,min (Appendix~\ref{app:mining}).

\begin{table}[t]
\centering
\small
\setlength{\tabcolsep}{3pt}
\resizebox{\columnwidth}{!}{%
\begin{tabular}{@{}lrrrrr@{\hspace{9pt}}rrr@{}}
\toprule
 & & \multicolumn{4}{c}{Videos by length} & \multicolumn{3}{c}{Duration (s)} \\
\cmidrule(lr){3-6}\cmidrule(l){7-9}
Part & Videos & $<$30\,s & 30--60\,s & 1--2\,min & 2--4\,min & Min & Median & Max \\
\midrule
Shot2Story & 4,023 & 3,847 & 176 & 0 & 0 & 10.0 & 14.9 & 39.8 \\
CaReBench & 1,000 & 966 & 32 & 1 & 1 & 1.1 & 12.7 & 124 \\
Vript short & 2,080 & 1,407 & 668 & 5 & 0 & 3.6 & 18.8 & 60.1 \\
Vript long & 1,500 & 0 & 500 & 500 & 500 & 31.6 & 93.6 & 240 \\
\midrule
All & 8,603 & 6,220 & 1,376 & 506 & 501 & & & \\
\bottomrule
\end{tabular}}
\caption{The 8,603 corpus videos by part and length bin, with duration in seconds. Bins use the study's rule: under 30\,s, up to 60\,s, up to 120\,s and up to 240\,s.}
\label{tab:corpus-lengths}
\end{table}

\section{Query Construction and Labels}
\label{app:mining}

\paragraph{Sources and splits.}
Sources must have a caption; Vript videos marked as failing its integrity check are excluded, and Vript long spans are English only. The 100 development sources (25 per corpus part) set the prompts, rerank depth and the S6 threshold. The 100 test sources (Section~\ref{sec:setup}) are drawn with a fixed seed, stratified by corpus part and Vript length bin with a fixed number per stratum (not proportional to corpus size, so Vript long is over-represented); Vript short splits into 14 under 30\,s and 14 of 30--60\,s, and Vript long into 14 of 30--60\,s, 15 of 1--2\,min and 13 of 2--4\,min. Every drawn source has a non-empty description and yields queries in both sets.

\paragraph{Mining.}
An LLM reads a source's caption, scene by scene (shots for Shot2Story, scenes for Vript), and lists atomic, self-contained facts \citep{min2023factscore,chen2023densex}. Each fact carries a type (attribute, object, on-screen text, count, scene, action, spatial relation, temporal order, other), its scene number, and two flags: visible in a single frame, and distinctive within a corpus of about 8,600 videos.

\paragraph{Selection.}
A fixed rule, not the LLM, keeps distinctive single-frame facts of the types attribute, object, on-screen text, count and scene, and drops actions, spatial relations, temporal order and other. These are the types a VLM can verify from one sampled frame (Section~\ref{sec:setup}); actions, spatial relations and temporal order need several frames, and grounder agreement on them was not measured. The miner prompt excludes audio and speech. At most one on-screen-text or logo fact enters a query, since a second one would make the query a near-unique text key. For Vript spans, facts are drawn from different scenes, spread across the span, because longer spans have longer captions. Table~\ref{tab:fact-funnel} gives the counts by type for the 100 test sources. Of the 3,812 mined facts, the type rule removes 1,678 (44\%), mostly actions and spatial relations, before any VLM sees them, and the single-frame and self-contained flags remove 7 more; the two-VLM filter then drops 294 of the 836 grounded candidates (35\%). Counts are dropped most often (.522) and attributes least (.298), with scenes, objects and on-screen text between .31 and .39. The types are the miner's own labels and were not validated.

\begin{table}[t]
\centering
\small
\setlength{\tabcolsep}{2pt}
\begin{tabular}{@{}lrrrrrr@{}}
\toprule
Type & Mined & Select. & Grounded & Kept & Dropped & Rate \\
\midrule
Attribute & 757 & 754 & 329 & 231 & 98 & .298 \\
Object & 791 & 790 & 281 & 174 & 107 & .381 \\
On-screen text & 269 & 269 & 110 & 67 & 43 & .391 \\
Count & 95 & 92 & 46 & 22 & 24 & .522 \\
Scene & 222 & 222 & 70 & 48 & 22 & .314 \\
\midrule
Action & 848 & 0 & -- & -- & -- & -- \\
Spatial relation & 646 & 0 & -- & -- & -- & -- \\
Temporal order & 38 & 0 & -- & -- & -- & -- \\
Other & 146 & 0 & -- & -- & -- & -- \\
\midrule
All & 3,812 & 2,127 & 836 & 542 & 294 & .352 \\
\bottomrule
\end{tabular}
\caption{Mined facts by type through the fact-mining funnel, for the 100 test sources. \emph{Mined}: facts listed by the LLM, labelled with its own type. \emph{Selectable}: facts that pass the fixed rule (type attribute, object, on-screen text, count or scene, visible in a single frame, self-contained); the other types are removed by the rule and never reach the VLMs. \emph{Grounded}: selectable facts chosen as candidates and checked by both VLMs (distinctive facts first, with a cap per source). \emph{Kept}: both VLMs answer yes. \emph{Dropped}: not confirmed by both VLMs (a no or unsure from either); drop rate is dropped over grounded. How often the descriptions state the facts of each type, including the excluded types, is in Table~\ref{tab:fact-types}.}
\label{tab:fact-funnel}
\end{table}

\paragraph{Do the kept facts favour what descriptions say?}
A text-only judge read each source's short indexed description and marked every fact of the source, kept and dropped shuffled together and unlabelled, as stated, contradicted or unclear (Table~\ref{tab:fact-judge}; 98 of the 100 test sources have a kept or dropped fact). Kept facts are stated more often than dropped ones (.131 $[.094, .171]$ on 542 facts against .044 $[.008, .086]$ on 135; per source $+.080$ $[+.029, +.133]$ over the 65 sources with both kinds) and contradicted less often (.026 against .067), so this check finds no sign that the filter favours facts the descriptions miss; it does not rule the concern out. Humans who read the same descriptions (Task E, App.~\ref{app:human}) agree with the judge on which of the 350 kept nested facts are stated ($\kappa=0.71$; the judge's stated has precision 0.85 and recall 0.67 against theirs). Humans mark more of them stated (51, or 15\%, against the judge's 40) and another 13\% implied; the judge has no \emph{implied} answer and marks most of those unclear, so that part of the difference comes from the label sets. The describer is also one of the two grounders, so shared blind spots could in principle produce this gap, but facts that Gemini accepted and \texttt{gpt-6-astra} rejected are stated far less often than kept facts (.042, 3 of 72) and are contradicted more often (.111 against .026), whereas shared blind spots would predict a stated rate near that of kept facts. One possible reading, which we have not verified, is that the filter removes facts that are not in the video, which the description also fails to state or contradicts; facts that only Gemini rejected are stated less often than kept ones (.048, 3 of 63, against .131), consistent with some correlation with the describer. Most answers are unclear (.84 to .94), the split cells hold few stated facts (3 and 3), the judge is checked against one annotator only, on 350 kept facts, and the filter selects single-frame, distinctive facts of five types by construction, which a single frame can verify; the check cannot show whether frame systems find dropped facts easy or hard. As a control on 98 development sources, mixed and kept-only judging agree on 97\% of 325 facts ($\kappa$ .91).

\begin{table}[t]
\centering
\small
\setlength{\tabcolsep}{3pt}
\resizebox{\columnwidth}{!}{%
\begin{tabular}{@{}lrrccr@{}}
\toprule
Group & Facts & Sources & Stated [95\% CI] & Contradicted [95\% CI] & Unclear \\
\midrule
Kept (both grounders yes) & 542 & 98 & .131 [.094, .171] & .026 [.009, .047] & .843 \\
Dropped (one grounder yes) & 135 & 65 & .044 [.008, .086] & .067 [.022, .121] & .889 \\
\addlinespace[2pt]
\quad of which Gemini yes, \texttt{gpt-6-astra} no & 72 & 45 & .042 [.000, .105] & .111 [.038, .206] & .847 \\
\quad of which Gemini no, \texttt{gpt-6-astra} yes & 63 & 38 & .048 [.000, .109] & .016 [.000, .053] & .937 \\
\bottomrule
\end{tabular}}
\caption{Does the indexed description state the facts the grounding filter kept? A text-only judge (\texttt{gpt-6-sol}, low effort) read each source's short indexed description (about 44 words), never the video, and marked every fact of that source as stated, contradicted or unclear (not mentioned). Kept and dropped facts of a source were shuffled together, unlabelled, in one call per source; 98 test sources. \emph{Dropped}: exactly one of the two grounders (Gemini~3.5 Flash-Lite, \texttt{gpt-6-astra}) said yes; facts both rejected are left out. Intervals are 10,000-draw cluster bootstraps over sources. Per-source stated rate, kept minus dropped: +.080 [+.029, +.133] over 65 sources with both groups.}
\label{tab:fact-judge}
\end{table}

\paragraph{Does the type filter favour descriptions?}
Table~\ref{tab:fact-types} runs the same text-only judge on all 3,812 mined facts of the 100 test sources, so that kept and excluded types can be compared. The 1,678 facts of the excluded types (actions, spatial relations, temporal order, other) are stated in 4.8\% of cases $[.034, .066]$, against 6.2\% $[.046, .082]$ for kept-type facts that were also never grounded. The intervals overlap, so the excluded types are not stated more often than the kept types. The grounder-confirmed filter is the step that favours description-stated facts: 13.7\% $[.102, .177]$ of facts both grounders confirmed are stated, against about 6\% for ungrounded kept-type facts. Whether that step favours frames or descriptions more is unclear, because facts a VLM finds salient may also help a frame verifier. Descriptions are sparse for every type, with about 90\% of facts of each type unclear.
Three limits apply. First, the judge was validated against humans only on kept facts, and it may be stricter on paraphrase-heavy action facts. Second, \emph{stated} is a judge label, not a human label or a retrieval outcome. Third, the run does not measure the frame gain on excluded types, and the temporal ($n=38$), count ($n=95$) and other ($n=146$) cells are too small to compare.

\begin{table}[t]
\centering
\small
\setlength{\tabcolsep}{3pt}
\resizebox{\columnwidth}{!}{%
\begin{tabular}{@{}llrcc@{}}
\toprule
Type & In query set & $n$ & Stated (\%) & 95\% CI \\
\midrule
Attribute & type kept & 757 & 4.8 & [.029, .068] \\
Object & type kept & 791 & 7.2 & [.049, .103] \\
On-screen text & type kept & 269 & 11.5 & [.069, .176] \\
Count & type kept & 95 & 10.5 & [.049, .168] \\
Scene & type kept & 222 & 12.2 & [.079, .180] \\
\addlinespace[2pt]
Action & type excluded & 848 & 5.4 & [.035, .079] \\
Spatial relation & type excluded & 646 & 3.4 & [.016, .058] \\
Temporal order & type excluded & 38 & 2.6 & [.000, .086] \\
Other & type excluded & 146 & 8.2 & [.029, .145] \\
\midrule
Excluded types, pooled & type excluded & 1,678 & 4.8 & [.034, .066] \\
Kept types, pooled & type kept & 2,134 & 7.5 & [.059, .096] \\
\midrule
Excluded type & not grounded & 1,678 & 4.8 & [.034, .066] \\
Kept type, never grounded & not in queries & 1,298 & 6.2 & [.046, .082] \\
Kept type, both grounders yes & in queries & 542 & 13.7 & [.102, .177] \\
Kept type, grounders split yes/no & dropped & 135 & 3.7 & [.008, .067] \\
Kept type, grounded, other outcome & dropped & 159 & 1.3 & [.000, .031] \\
\bottomrule
\end{tabular}}
\caption{Does the type filter favour descriptions? A text-only judge (\texttt{gpt-6-sol}, low effort, the Table~\ref{tab:fact-judge} prompt) read each source's indexed description (about 44 words), never the video, and marked every mined fact of the 100 test sources, 3,812 in all, as stated, contradicted or unclear. \emph{Stated} is the share marked stated. Unlike Table~\ref{tab:fact-judge}, facts of all types were shuffled together (seed fixed) and sent in chunks of at most 40, 145 calls in all, not one call per source; on the 677 facts both runs cover, the two runs agree on 95.4\% (646). 95\% intervals are 2,000-draw cluster bootstraps over sources. The last block splits the facts by grounding outcome: the three kept-type groups are not the same population as the excluded types. Types are the miner's own labels.}
\label{tab:fact-types}
\end{table}

\paragraph{Grounding.}
Each selected fact is checked on about six frames sampled from its own scene or shot window, at least 0.5\,s away from every index timestamp, at most 768\,px on the longest side. Where the caption has no scene times (Vript short), the window is the whole video at 1\,fps, at most 30 frames. Gemini (at low media resolution) and \texttt{gpt-6-astra} each answer yes, no or unsure; a fact is kept only if both answer yes \citep{jing2024faithscore,verga2024poll}. A separate Gemini call on all index frames logs whether each kept fact is also visible in the index frames, which measures how much the index frames themselves miss. We report the share of facts dropped by the agreement rule, per type and per model, and the agreement between the two grounders.

\paragraph{Nested queries.}
Each fact set extends the last by one fact for $k=1..4$ (Section~\ref{sec:setup}), so each source contributes up to four queries. An LLM writes one natural query per set, rewording quoted text loosely, and returns a map from each fact to the phrase that states it. The map gives the gold constraints for the gold-split variant and for per-fact analysis.

\paragraph{Near-miss queries.}
For source $A$ we embed all captions in a mining-only index (never seen by systems) and take $A$'s five nearest neighbours $B$ from other source videos. An LLM reading both captions lists facts that hold in both and facts of $A$ that are absent from or contradicted in $B$, preferring on-screen text, objects and attributes for the distinguishing fact $d$, since their absence is easier to verify. Shared facts are grounded, by the agreement rule, on local frames of both videos; $d$ must be confirmed in $A$ and confirmed absent in $B$ on up to 24 frames sampled at 1\,fps over the whole of $B$. Neighbours are tried in order of similarity, up to three, until one pair yields three shared facts and $d$. The queries are then written as in Section~\ref{sec:setup}; at $k=1$ the query states $d$ alone. Unlike MultiConIR \citep{lu2025multiconir} and caption-foil benchmarks, nothing is edited; unlike earlier natural pairs \citep{li2024naturalbench,krojer2025mvp,ventura2024covr2,zhang2024vinoground}, the shared and missing facts are verified one by one and the condition count is varied. We report the yield of HN videos per $k$ and per corpus part.

\paragraph{Examples.}
Table~\ref{tab:examples} shows the queries for one source, a Shot2Story clip. Its HN video shows a man on a stage and a microphone but no explosion.

\begin{table}[t]\centering\small
\begin{tabular}{@{}c p{0.42\columnwidth} p{0.45\columnwidth}@{}}
\toprule
$k$ & Nested query & near-miss query \\
\midrule
1 & A man wearing a black hair band appears in the footage. & Footage clearly shows a visible \underline{explosion} taking place in the scene. \\
2 & A man wearing a black hair band appears, and an explosion is shown at another point. & A man is on a stage at one point, and an \underline{explosion} is visible at another point in the video. \\
3 & At one point a man wears a black hair band; at another point an explosion is visible; at another point three men are present. & A man stands on a stage; at another point, an \underline{explosion} is visible, and the video also shows a microphone. \\
4 & I remember footage showing a man wearing a black hair band, an explosion visible, three men present, and a microphone visible on the stage. & A clip shows an \underline{explosion}; at another point, a man is on a stage, a microphone is visible, and a man wears red clothing. \\
\bottomrule
\end{tabular}
\caption{Queries built from one source video. The distinguishing fact $d$ of the near-miss queries is the explosion (underlined), the one constraint the HN video does not meet; it meets the other $k-1$.}
\label{tab:examples}
\end{table}

\paragraph{Frame judge.}
Per source, the pool is the union of the top 10 of S0, S1, S2b@50 and S4@50 over all of the source's queries, plus the source and its HN video. Each model makes one call per (source, candidate) that answers every fact of the source's queries on at most 16 of the candidate's index frames at 512\,px. Labels apply the rule of Section~\ref{sec:setup} fact by fact, so one rejected fact rejects every nested query that contains it. Candidates reached only by queries with $k\ge3$ are judged on a seeded 25\% sample; unjudged candidates count as non-relevant. The HN video is judged like any other candidate, but the expanded labels count it as non-relevant by rule, since it lacks $d$ by construction; the judges' verdicts on it serve only as a check (they accept it on 4 of the 1,026 near-miss queries checked). Videos that later runs ranked into the top 10 outside this pool were judged the same way in two supplementary label fills, by the same two judges and the same rule: 2,625 (source, candidate) pairs over 613 queries, then 1,996 pairs for the top 10 of four local rerankers (Appendix~\ref{app:results}).

\paragraph{Label quality.}
Besides the human check in App.~\ref{app:human}, we report (i) source acceptance, the share of queries for which both judges accept every fact on the source's own frames; (ii) the judges' per-fact and per-video precision, recall and Cohen's $\kappa$ against the human check (App.~\ref{app:human}); and (iii) the number of extra positives per $k$ and set. Because the judges reject some true facts, expanded labels miss some relevant videos; they also count some that humans reject (App.~\ref{app:human}), so they are approximate in both directions rather than a lower bound. On the development set, a re-run of the judge and re-judging at default resolution changed source acceptance by no more than run-to-run noise, so the pool is not re-judged at a higher resolution. The judge pool of the test set holds 8,712 (source, candidate) pairs, and with the pairs of the rerankers of Appendix~\ref{app:results} 10,708 are judged. The main judging pass planned 6,889 of them (the 2,458 candidates reached only through $k\ge3$ queries entered on a seeded 25\% sample, 635 of them) and judged 6,087; the first supplementary fill then judged the remaining 2,625 pairs (613 queries; 1,576 of the pairs are S2@50 top-10 slots), so all 8,712 pairs are judged; a second fill judged 1,996 further pairs, the top-10 slots of four off-the-shelf rerankers, so no system in the main results has an unjudged top-10 slot (Appendix~\ref{app:results}). The first fill added 349 relevant (query, video) pairs and the second 228 (1,958 in all, from 1,381 before the first); strict labels are unchanged, the first fill moved every expanded-label difference by at most 0.011 and the second by at most 0.013, and two intervals with a bound within 0.002 of zero changed side.

\section{Statistics and Cost Model}
\label{app:stats}

\paragraph{Metrics.}
We report nDCG@10 and R@1 under both label sets; on the near-miss set, also the rate at which the HN video ranks above the source when both are in the reranked pool. Cost per query is calls, input tokens and images; index cost is reported separately.

\paragraph{Statistics.}
The unit is the query; queries built from the same source video form a cluster. Every comparison is paired on the same queries. Intervals (95\%) come from a bootstrap that resamples source videos with replacement (10,000 resamples) and recomputes the mean difference. We report them overall and per $k$, per set (nested, near-miss), per corpus part and per Vript length bin. Planned comparisons are S1 vs S0, S4 vs S2, S3 and S5 vs S4, S5 vs S3 (equivalence at the pre-set $\pm0.02$), and S6 vs S4 and S5, and, for the score-scale ablation, S2b vs S0, S4 vs S2b and S2 vs S2b. On the near-miss set we also report the HN-above-source rate, restricted to queries with both videos in the reranked pool, since a reranker cannot recover a source that S0 missed; we report how often the source is outside S0's top 10 and top 50.

\paragraph{Length.}
Length is analysed within Vript only (short and long), so that it is not confounded with the corpus part. Besides the four bins, we regress each system's per-query gain (over S0, or over S4@50 for the frame systems) on log duration, with $k$, query set (nested or near-miss), Vript part (short or long) and scenes per minute as covariates, and report the slope with a cluster-bootstrap interval and the minimal slope the design detects at 80\% power. Caption length is not a covariate: over the 70 Vript test sources the correlation of caption words per minute with scenes per minute is 0.997 (0.996 on the 28 short, 0.998 on the 42 long), because Vript captions run a near-constant length per scene (120 to 132 words per scene between the 10th and 90th percentiles), and with both in the model each has a variance inflation factor of 169, so length and scene rate cannot be separated by the regression. Long spans also come from YouTube and short videos from Shorts and TikTok, so the slope describes an association only (\S\ref{sec:res-length}).

\paragraph{Runs.}
In the main results each stochastic component runs once at fixed settings, with responses cached. A repeat run (fresh cache keys, expanded labels) on 52 nested and 64 near-miss test queries changes nDCG@10 by $+0.000$ $[+0.000, +0.001]$ for S2b@50 and $-0.018$ $[-0.042, +0.000]$ for S4@50 on the nested set, by $-0.019$ $[-0.042, -0.003]$ for S2b@50 and $+0.018$ $[-0.013, +0.057]$ for S4@50 on the near-miss set, and by $-0.005$ $[-0.016, +0.000]$ for S5 (44 contested queries); the top-1 video is unchanged for 100\% (S2b@50) and 75\% (S4@50) of the nested queries, 97\% and 83\% of the near-miss queries, and 98\% of S5's queries. Verifier dependence is measured by the second verifier (Appendix~\ref{app:systems}).

\paragraph{Online cost.}
Per query we log input and output tokens, calls, images and searches for every model call; we report cost as calls, input and output tokens (output tokens include the verifier's reasoning tokens) and images, not wall-clock time, which depends on provider load. Latency is set by the number of sequential model stages: S1 and S4 have one, S2 and S2b two (split, then verify), and S5 adds one stage of ten parallel frame calls after S4. Per-call costs measured in the development run (input tokens) are:
a description-based verification call (S2, S2b, S4), about 200--290 tokens per candidate, so S4 at depth 20 is about 4.1k tokens per query;
a frame call (S3, S3b, S5) on a candidate's own 0.5\,fps index frames, about 10.8k tokens and 26 images on average (389 tokens per frame), so S5 on the top 10 adds about 106k tokens per query in the development run, about 26 times S4. In the test run, over all 722 queries, S5's frame step alone is 80.3k tokens (81.7k for S3), about $7.7\times$ S4@50's 10.4k; with the S4@50 pass included, the full pipeline is 90.7k, or $8.7\times$ (Table~\ref{tab:results});
a splitter call, about 180 input and 120 output tokens.
A search costs one query embedding. By Vript length bin, S5's frame-step input on the contested queries is 102k tokens under 30\,s ($n=11$), 98k at 30--60\,s ($n=22$), 116k at 1--2\,min ($n=20$) and 138k at 2--4\,min ($n=12$), 112k over the 65 Vript contested queries and 96.0k over all 96 (which include 31 non-Vript queries); on the nested sample it is 78k, 82k, 98k and 113k by bin ($n=17$, 15, 10, 14; 56 in all, against a full-set mean of 80.3k over 722 queries). Weighted by bin size, the nested-sample mean is about 91k, so the sample runs above the full-set mean and overstates average cost. Per query the spread is wide: on the nested sample the standard deviation is 29k, 36k, 22k and 66k tokens by bin (ranges 45k--155k, 37k--150k, 50k--139k and 46k--264k), and over all Vript test queries ($n=102$, 193, 111, 101) the means are 71k, 80k, 105k and 105k with standard deviations of 33k, 40k, 58k and 55k. Duration thus predicts average frame cost but little of any one query's cost. These counts are the frame step only; with the S4@50 pass the total on the contested queries is 106.5k. Per $+0.01$ of S5 $-$ S4@50 (expanded labels) the contested frame step is about 9.4k, 5.8k, 10.4k and 10.3k tokens by bin, against 7.2k over all 96 contested queries; the per-bin gains have wide intervals, so we do not read a trend into them.

\paragraph{Offline cost.}
Describing the full corpus at 0.5\,fps took 8,129 calls (the remaining videos were described in development), 147,040 frames, 165.2M input and 0.46M output tokens, about 20k input tokens and 18 frames per video (US\$51 with Gemini). On the development mix, with more long Vript videos, it used about 32k input tokens and 29 frames per video (median 8 frames for Shot2Story and CaReBench, 15.5 for Vript short and 46 for Vript long, at most 120), at about 1.1k tokens per frame at default resolution. Building both query sets for 100 sources (mining excluded) took 2,463 calls, 18,637 images and 7.9M input tokens; grounding at low resolution costs 262 tokens per frame. A frame-judge call costs about 3.3k input tokens (at most 16 frames, 512\,px). Frame storage for S3b and S5 is about 6\,GB of JPEG frames at 0.5\,fps for the 8,603 videos.

\paragraph{Description statistics.}
Table~\ref{tab:descstats} summarises the generated descriptions of the 100 test sources. Descriptions are short and do not grow with video length: the mean is 41--48 words in every Vript bin (33--48 across parts; 44 overall, range 19--103), 1.5--1.9 sentences (1--4) and 39--61 output tokens (20--125), while the mean number of input frames rises from 9--10 for Shot2Story and CaReBench clips to 94 (73--112) for Vript spans of 2--4\,min. Across the whole corpus (8,589 non-empty descriptions) the mean is 44.7 words (median 44, range 9--231) and 56.9 output tokens, and 155 descriptions (1.8\%) exceed the three-sentence limit. The describer returned an empty description for 14 videos, none of them a test source. Sentences are counted with a simple punctuation rule, and output tokens are those billed by the describer API.
\begin{table}[t]
\centering
\footnotesize
\setlength{\tabcolsep}{3pt}
\resizebox{\columnwidth}{!}{%
\begin{tabular}{@{}lrcccc@{}}
\toprule
 & $n$ & Words & Sentences & Output tok. & Frames \\
\midrule
Shot2Story & 15 & 43 (22--69) & 1.9 (1--4) & 53 (24--95) & 9 (8--13) \\
CaReBench & 15 & 33 (21--51) & 1.8 (1--3) & 39 (24--71) & 10 (8--28) \\
Vript short & 28 & 48 (26--83) & 1.9 (1--3) & 61 (33--125) & 18 (8--30) \\
Vript long 30--60\,s & 14 & 44 (19--60) & 1.9 (1--3) & 56 (27--91) & 25 (19--29) \\
Vript long 1--2\,min & 15 & 47 (19--103) & 1.5 (1--3) & 60 (21--118) & 47 (32--60) \\
Vript long 2--4\,min & 13 & 41 (19--68) & 1.5 (1--3) & 50 (20--95) & 94 (73--112) \\
\midrule
All & 100 & 44 (19--103) & 1.8 (1--4) & 54 (20--125) & 31 (8--112) \\
\bottomrule
\end{tabular}}
\caption{Index descriptions of the evaluated test sources: mean (min--max) per source.}
\label{tab:descstats}
\end{table}

\paragraph{Money.}
The only verifier price we recorded is the list price of \texttt{gpt-6-sol}, US\$2 per 1M input and US\$10 per 1M output tokens (standard tier, noted on 2026-10-26 and not re-checked). At that price, counting output tokens (including reasoning tokens), a query over all 722 test queries costs US\$0.034 for S4@50, \$0.069 for S2@50, \$0.159 for S6, \$0.198 for S5 and \$0.207 for S3 (S4@50: 10.4k input and 1.3k output tokens; S5: 90.7k and 1.7k). The description reranks spend relatively more on output, so in dollars the frame premium shrinks from $8.7\times$ to $5.8\times$ S4@50 for S5 ($6.0\times$ for S3, $4.6\times$ for S6), while S2@50 rises from $1.5\times$ to $2.0\times$. These figures exclude the splitter (\texttt{gpt-6-luna}; about 175 input and 94 output tokens per query on the test set, used by S2 and S3) and the query embedding (\texttt{text-embedding-3-small}, one call), for which no price is recorded. Prices change, so the token counts (Table~\ref{tab:results}) are the primary cost measure.

\section{Reproducibility and Release}
\label{app:release}

\paragraph{Dataset.}
The dataset is available at \dataurl. It holds the corpus with its index descriptions, length bins, video identifiers and Vript span boundaries; the nested and near-miss queries with their facts, $k$ and fact-to-phrase maps, each HN video with its shared and distinguishing facts, and the time at which the two grounders agreed on each nested fact; and the strict, expanded and human labels with both judges' per-fact answers. It includes the Shot2Story and CaReBench videos. Vript's terms do not allow redistribution, so for Vript it gives pointers and a script that rebuilds the clips and spans from the user's own copy. The code is available at \codeurl. All prompts are in Appendix~\ref{app:prompts} and the exact model identifiers with versions and dates in Table~\ref{tab:models}.

\paragraph{Upon acceptance.}
We will release the cached per-candidate scores, per-query outputs and cost logs of every system.

\paragraph{What we generate.}
Descriptions, facts, groundings, HN videos, queries, judge labels and verifier outputs are all model-generated. The human check (\S\ref{sec:setup}, App.~\ref{app:human}) covers every fact, the HN pair of every near-miss source (on 261 of 372 near-miss queries) and the S4@50 vs S5 top-1 disagreements, not the rest of the judge labels. Everything else comes from the released dataset.

\paragraph{Determinism.}
Open models run with fixed seeds and greedy decoding where possible. Closed models are pinned by version and date, except \texttt{gemini-3.8-flash}, which we called by alias, so no dated snapshot was recorded for it and its results may not reproduce exactly. They run once for the main results (with a repeat of S2b, S4 and S5 on a subset), and their responses are cached. Frame sampling is deterministic (the index frames of \S\ref{sec:method}, at least 8 frames, fixed resolution).

\paragraph{Code.}
The pipeline has three stages that communicate through files: build the index, run the systems with cached S0 rankings shared by all systems, and evaluate.

\section{Full Results}
\label{app:results}

Tables~\ref{tab:perk} and~\ref{tab:frames} give the paired differences behind the main text, and Table~\ref{tab:results-subsets} the frame systems on the two subgroups where the binary S3b also ran. Under strict labels (source only; Table~\ref{tab:results-strict}) the order of systems is the same as under expanded labels, apart from the frame pair: S5 $-$ S3 is $+0.005$ $[-0.004, +0.013]$ on all near-miss and $-0.003$ $[-0.011, +0.004]$ on all nested queries under strict labels, and on the contested subgroup S5 leads the binary S3b by 0.023 $[0.001, 0.052]$ (not significant after a Holm correction), while under expanded labels no frame-pair interval excludes zero. Source recall in S0's top 10, 20 and 50 is .526, .600 and .694 on the nested set and .516, .583 and .704 on the near-miss set. So the source is outside S0's top 50 for 31\% of nested queries (95\% interval 24--38\%) and 30\% of near-miss queries (23--37\%), which no reranker can recover. Table~\ref{tab:desc_samples} shows one index description per Vript length bin next to the longest nested query built from the same source.

\begin{table}[t]
\centering
\small
\setlength{\tabcolsep}{2.4pt}
\begin{tabular}{@{}lcccccc@{}}
\toprule
 & \multicolumn{2}{c}{Nested} & \multicolumn{2}{c}{Near-miss} & \multicolumn{2}{c}{R@1} \\
\cmidrule(lr){2-3}\cmidrule(lr){4-5}\cmidrule(l){6-7}
System & all & $k{=}4$ & all & $k{=}4$ & Nest. & N-miss \\
\midrule
S0 & .375 & .485 & .397 & .471 & .249 & .290 \\
S1 & .293 & .345 & .284 & .311 & .183 & .180 \\
S2b@50 & .414 & .476 & .420 & .511 & .283 & .301 \\
S2@50 & .485 & .590 & .431 & .540 & .366 & .325 \\
S4@50 & \textbf{.582} & \textbf{.734} & \textbf{.519} & \textbf{.610} & \textbf{.494} & \textbf{.409} \\
\midrule
Qwen-2B & \textbf{.532} & \textbf{.698} & \textbf{.509} & \textbf{.611} & \textbf{.420} & \textbf{.414} \\
PW-2B & .478 & .648 & .502 & .582 & .346 & .403 \\
LW-2B & .474 & .627 & .485 & .578 & .371 & .401 \\
\midrule
S3 & \textbf{.648} & \textbf{.808} & .584 & \textbf{.710} & \textbf{.623} & .516 \\
S5 & .644 & \textbf{.808} & \textbf{.589} & .703 & .617 & \textbf{.530} \\
\bottomrule
\end{tabular}
\caption{Strict-label results (source only) in the layout of Table~\ref{tab:results}: nDCG@10 over all queries ($n=350$ nested, 372 near-miss) and at $k=4$, and R@1 over all queries; bold is best per block. Cost does not depend on the labels and is in Table~\ref{tab:results}, which has no S2b@50 row: S2b@50 makes 51 calls and 14.6k input tokens per query, and S3 and S5 read 199 frame images per query. S6 is evaluated on expanded labels only.}
\label{tab:results-strict}
\end{table}

\begin{table}[t]
\centering
\small
\setlength{\tabcolsep}{3pt}
\begin{tabular}{@{}lcccccc@{}}
\toprule
 & \multicolumn{2}{c}{Nested sample} & \multicolumn{2}{c}{Contested} & \multicolumn{2}{c}{R@1} \\
\cmidrule(lr){2-3}\cmidrule(lr){4-5}\cmidrule(l){6-7}
System & all & $k{=}4$ & all & $k{=}4$ & Nest. & Cont. \\
\midrule
\multicolumn{7}{@{}l}{\emph{Frame subsets} ($n=56$ / $96$ contested)} \\
S0 & .380 & .414 & .635 & .672 & .232 & .552 \\
S4@50 & .656 & .668 & .809 & .796 & .625 & .740 \\
S3b & .698 & \textbf{.750} & .923 & .931 & \textbf{.732} & .948 \\
S3 & .699 & \textbf{.750} & .930 & .946 & \textbf{.732} & .958 \\
S5 & \textbf{.703} & \textbf{.750} & \textbf{.943} & \textbf{.953} & \textbf{.732} & \textbf{1.000} \\
\bottomrule
\end{tabular}
\caption{Frame systems on the frozen frame subsets of the test set (expanded labels): the nested Vript sample ($n=56$) and the contested near-miss queries ($n=96$, source and near miss both in S4@50's top 10), the only queries S3b (per-constraint yes/no verifier on frames) was run on. nDCG@10 over all queries of the subset and at $k=4$, R@1 over all queries of the subset; bold is best per column. S3, S5 (and S6) on every test query are in Table~\ref{tab:results}.}
\label{tab:results-subsets}
\end{table}

\paragraph{Binary against graded per-constraint checks (ablation of the score scale).} On descriptions (the grading gain and S4@50's lead are in Section~\ref{sec:results}), at depth 10 S4@10 leads S2b@10 by $+0.094$ / $+0.055$, and S2@50 gains $+0.127$ $[+0.079, +0.176]$ and $+0.081$ $[+0.033, +0.125]$ over S0. On frames, where S3b ran only on the subgroups (Table~\ref{tab:results-subsets}), S3 $-$ S3b is $+0.007$ $[-0.011, +0.032]$ on the contested queries and $+0.002$ $[-0.000, +0.005]$ on the nested sample, and S3b $-$ S5 is $-0.020$ $[-0.053, +0.003]$ and $-0.005$ $[-0.013, +0.000]$ (Table~\ref{tab:rc-frames}). Under strict labels, S2@50 $-$ S2b@50 is $+0.071$ $[+0.034, +0.108]$ and $+0.011$ $[-0.022, +0.042]$. Taking the minimum grade over constraints instead of the mean (the secondary aggregation) lowers S2@50 by $-0.038$ $[-0.067, -0.011]$ on the nested set and moves it by $+0.012$ $[-0.015, +0.041]$ on the near-miss set, and keeps S5 $-$ S3 at or below $+0.005$ on both sets.

\paragraph{Re-aggregating the per-constraint grades does not close the description gap.} The graded checks store one 0--10 grade per split constraint for every candidate they score (S2: all 50 S0 candidates; S3: S4@50's top 10), so we re-rank the same candidates by five post-hoc aggregations of the same grades, with ties kept in pool order: the minimum; the mean after setting every 5 (unmentioned) to 0; the mean over grades other than 5 (5 if all are 5); the count of grades $\ge8$; and that count minus the count of grades $\le2$. On descriptions 18.8\% of candidates have only 5s, and on frames 0.6\% (44\% of S2's description grades are 5s, against 4\% on frames). On queries whose split matches their $k$ constraints, S2 grades a fact the description only implies 6.4 on average and a stated one 9.2; among queries with no fact left out, S4@50 $-$ S2@50 is $+0.105$ $[+0.056, +0.165]$ where some fact is only implied ($n=67$) and $+0.047$ $[+0.014, +0.091]$ where every fact is stated ($n=101$). Every aggregation leaves S4@50 ahead of S2@50 on both sets under both label sets, every interval above zero ($p<.001$; Table~\ref{tab:reagg}). Against the paper's mean (pooled, expanded labels), the minimum ($-0.012$ $[-0.031, +0.007]$) and stated minus contradicted ($+0.001$ $[-0.015, +0.018]$) do about as well, and unmentioned as 0 ($-0.038$), the mean over grades other than 5 ($-0.072$) and the count of grades $\ge8$ ($-0.043$) do worse. On frames, S5 $-$ S3 stays within $\pm0.02$ for every aggregation and both sets except the mean over grades other than 5 on near-miss queries ($+0.014$, 90\% interval $[+0.006, +0.022]$; strict $[+0.004, +0.023]$); under strict labels the count of grades $\ge8$ on near-miss queries sits on the margin ($+0.010$, $[+0.001, +0.020]$). Every frame aggregation still gains over S4@50 (pooled $+0.067$ to $+0.074$ expanded, $+0.059$ to $+0.065$ strict). A Holm correction over these aggregations would not move any description interval across zero.

\begin{table*}[t]
\centering
\footnotesize
\setlength{\tabcolsep}{3pt}
\resizebox{\textwidth}{!}{%
\begin{tabular}{@{}lcccccc@{}}
\toprule
 & \multicolumn{4}{c}{Descriptions: S4@50 $-$ S2@50 (95\% interval)} & \multicolumn{2}{c}{Frames: S5 $-$ S3 (90\% interval), expanded} \\
\cmidrule(lr){2-5}\cmidrule(l){6-7}
Aggregation & Nested, exp. & Near-miss, exp. & Nested, strict & Near-miss, strict & Nested & Near-miss \\
\midrule
Mean (paper) & $+0.101$ $[+0.069, +0.136]$ & $+0.094$ $[+0.061, +0.131]$ & $+0.097$ $[+0.065, +0.131]$ & $+0.088$ $[+0.051, +0.129]$ & $-0.000$ $[-0.004, +0.003]$ & $+0.005$ $[-0.001, +0.011]$ \\
Minimum & $+0.139$ $[+0.101, +0.180]$ & $+0.082$ $[+0.047, +0.118]$ & $+0.136$ $[+0.096, +0.177]$ & $+0.068$ $[+0.031, +0.107]$ & $+0.005$ $[+0.000, +0.011]$ & $+0.004$ $[-0.002, +0.010]$ \\
Unmentioned $\to$ 0 & $+0.156$ $[+0.118, +0.197]$ & $+0.117$ $[+0.086, +0.151]$ & $+0.149$ $[+0.109, +0.190]$ & $+0.104$ $[+0.070, +0.143]$ & $-0.000$ $[-0.004, +0.003]$ & $+0.006$ $[-0.000, +0.012]$ \\
Mean over $\ne5$ & $+0.142$ $[+0.106, +0.180]$ & $+0.197$ $[+0.156, +0.238]$ & $+0.136$ $[+0.101, +0.174]$ & $+0.172$ $[+0.128, +0.218]$ & $+0.004$ $[-0.001, +0.009]$ & $+0.014$ $[+0.006, +0.022]$ \\
Count $\ge8$ & $+0.164$ $[+0.128, +0.203]$ & $+0.119$ $[+0.092, +0.149]$ & $+0.150$ $[+0.114, +0.190]$ & $+0.095$ $[+0.064, +0.128]$ & $+0.006$ $[+0.001, +0.012]$ & $+0.010$ $[+0.001, +0.019]$ \\
\#$\ge8$ $-$ \#$\le2$ & $+0.108$ $[+0.076, +0.143]$ & $+0.087$ $[+0.059, +0.117]$ & $+0.097$ $[+0.066, +0.132]$ & $+0.072$ $[+0.040, +0.105]$ & $+0.003$ $[-0.001, +0.008]$ & $+0.003$ $[-0.002, +0.009]$ \\
\bottomrule
\end{tabular}}
\caption{Re-aggregations of the cached per-constraint grades (post hoc). Left: S4@50's lead over the graded description check S2@50 with each aggregation, nDCG@10, all test queries ($n=350$ nested, 372 near-miss); S4@50 scores .636 / .626 (expanded). Right: S5 $-$ S3 on frames with the pre-set equivalence margin $\pm0.02$ (expanded labels); only the mean over grades $\ne5$ on near-miss queries falls outside it.}
\label{tab:reagg}
\end{table*}

\paragraph{The same-candidate lead does not depend on who chose the candidates (post hoc).} Table~\ref{tab:w1-pools} ranks the same ten candidates by S4@50's stored grades and by S2's stored grades (mean; ties in S0 order), for the ten chosen by S2 (its own top 10), by S0 and by S4@50, on all 722 queries. Every gap is positive with its interval above zero on both sets, and under strict labels (pooled $+0.046$, $+0.042$, $+0.057$). On S4@50's top 10 the gap is $+0.054$ $[+0.032, +0.076]$ where the splitter returns $k$ constraints ($n=443$) and $+0.050$ $[+0.025, +0.078]$ where they also match the gold facts ($n=330$; no finished judge labels exist for splits), and $+0.079$ $[+0.039, +0.122]$ where some fact is only implied (first annotator, $n=201$) against $+0.027$ $[+0.005, +0.058]$ where all are stated ($n=101$). Frames: S5 $-$ S3 is $+0.002$ $[-0.001, +0.006]$ there, and both reach 0.707 and 0.705 of the 0.716 oracle. Splitting the full S4@50 $-$ S2@50 lead per query into selection (swapping in the other reranker's top-10 set under a fixed order rule) and ordering, and averaging the two orders of the swap, gives selection $+0.041$ $[+0.022, +0.062]$ and ordering $+0.061$ $[+0.041, +0.081]$ on nested queries, and $+0.050$ $[+0.031, +0.071]$ and $+0.044$ $[+0.025, +0.066]$ on near-miss queries (expanded labels; strict labels give $+0.037$ and $+0.060$, $+0.045$ and $+0.043$). The two orders of the swap differ by at most 0.022, and the two top tens share 5.1 and 6.4 of ten videos on average. S2's mean grades tie often; breaking those ties by S4's order instead of S0's lowers the lead to $+0.076$ and $+0.075$, with both parts still above zero.

\begin{table}[t]
\centering
\footnotesize
\setlength{\tabcolsep}{3pt}
\resizebox{\columnwidth}{!}{%
\begin{tabular}{@{}lcccc@{}}
\toprule
Pool of ten & Nested & Near-miss & Pooled & Oracle \\
\midrule
S2 top 10 & $+0.060$ $[+0.041, +0.081]$ & $+0.033$ $[+0.016, +0.052]$ & $+0.046$ $[+0.033, +0.060]$ & 0.642 \\
S0 top 10 & $+0.043$ $[+0.023, +0.064]$ & $+0.041$ $[+0.022, +0.061]$ & $+0.042$ $[+0.026, +0.059]$ & 0.575 \\
S4@50 top 10 & $+0.061$ $[+0.037, +0.086]$ & $+0.056$ $[+0.031, +0.083]$ & $+0.058$ $[+0.039, +0.079]$ & 0.716 \\
\bottomrule
\end{tabular}}
\caption{Same ten candidates ranked by S4@50's grades minus by S2's grades (nDCG@10, expanded labels, 95\% interval); oracle: best reordering of the pool, all 722 queries.}
\label{tab:w1-pools}
\end{table}

\paragraph{S3 and S5 are equivalent within $\pm0.02$ on a margin set before the run; the binary S3b is not shown equivalent on the contested subgroup.} The 90\% interval of S5 minus S3 is $[-0.001, +0.011]$ on all near-miss and $[-0.004, +0.003]$ on all nested queries, both within the margin, under expanded labels and under strict labels alike; on the contested subgroup alone it is $[-0.001, +0.033]$, not within the margin, and on the nested sample $[+0.000, +0.009]$. The margins for the binary S3b against S5 were not pre-registered, so for that pair we report the smallest margin the 90\% interval of S3b $-$ S5 excludes: $\pm0.047$ on the contested queries and $\pm0.011$ on the nested sample under expanded labels, and $\pm0.047$ and $\pm0.020$ under strict labels, where S5 leads on the contested queries.

\paragraph{Restricting to sources both judges accept changes no overall conclusion.} On the 577 of 722 queries (80\%) where both judges accept the source (Table~\ref{tab:rc-accepted}), two subgroup cells under strict labels move from significant to including zero: S3b $-$ S5 on the contested queries ($-0.023$ to $-0.015$ on the 88 accepted) and S5 $-$ S4@50 on the nested sample ($+0.038$ to $+0.002$ on the 38 accepted).

\paragraph{On contested queries S5 stays near 1 at every $k$ while S4@50 does not.} On the contested queries S4@50 is .982 at $k=1$ (3 queries, too few for an interval) and .751, .840 and .796 at $k=2$ to 4, while S5 stays between .899 and 1.000 (Table~\ref{tab:rc-perk}); we run no trend test.

\paragraph{No top-10 candidate is unjudged after the label fill.} Before the supplementary fill (Appendix~\ref{app:mining}), 9--13\% of the other systems' top-10 slots and 25--28\% of S2@50's were unjudged and counted as non-relevant; the fill judged every such slot (2,625 pairs over 613 queries, 1,576 of them S2@50 slots) with the same two judges and rule. The fill moved every expanded-label difference by at most 0.011 and no interval across zero; strict labels, which need no judging, are unchanged and give the same ordering (above).

\paragraph{Splitter errors add to the loss from decomposing but do not explain it.}
On the nested set the splitter returns more than $k$ constraints for 52\% of queries (70\% at $k=1$; $n=350$ and 98) and fewer than $k$ for 3\% (10\% at $k=4$), while keeping the content (token-overlap recall of the facts .93--.97 per $k$); near-miss queries are over-split less often (14\%; 23\% at $k=1$). With an exact split S1 $-$ S0 is $+0.032$ $[-0.002, +0.073]$, $-0.086$ $[-0.166, -0.013]$, $-0.073$ $[-0.154, -0.000]$ and $-0.085$ $[-0.163, -0.016]$ for $k=1$--4 ($n=29$, 54, 41 and 34), against $-0.043$ $[-0.101, +0.013]$, $-0.135$ $[-0.235, -0.041]$, $-0.142$ $[-0.247, -0.046]$ and $-0.208$ $[-0.338, -0.090]$ when over-split ($n=69$, 40, 42 and 32), and $-0.021$, $-0.107$, $-0.106$ and $-0.140$ over all nested queries (Table~\ref{tab:perk}); under strict labels the exact-split cells are $+0.031$, $-0.083$, $-0.059$ and $-0.085$ and the over-split cells $-0.021$, $-0.131$, $-0.142$ and $-0.208$. Because exactly split queries are not a random subset, a gold-split run is the cleaner test; we ran it for S2b and S3b on the development set (366 nested queries, top 20, strict labels), passing the mined facts as the constraints. Gold constraints raise S2b by $+0.030$ $[+0.010, +0.053]$ (Table~\ref{tab:ablations-extra}), level with S0 (0.657 against 0.656) and far below S4 (0.770) (by $+0.060$ at $k=4$), and change S3b by only $+0.006$ $[-0.006, +0.020]$; on frames, S3b with gold constraints still beats S2b with gold constraints by $+0.090$ $[+0.059, +0.124]$. We did not run S1 with gold constraints.

\paragraph{Additional ablations: stronger fusion does not rescue decomposition at retrieval, and a repeat run moves little.} Table~\ref{tab:ablations-extra} completes Table~\ref{tab:ablations} with the retrieval variants, the repeat run and the MultiVENT~2.0 check. From cached embeddings, with RRF constant 60 and depth 500 as in S1, we add the full query as one more list (S1+Q), with weight 2 or $k$ (S1+2Q, S1+$k$Q), or replace RRF by the mean cosine over constraints, without or with the full query. Against S0 these give $-0.063$, $-0.055$, $-0.035$, $-0.069$ and $-0.034$ nDCG@10 on the nested set and $-0.065$, $-0.052$, $-0.043$, $-0.059$ and $-0.021$ on the near-miss set. The closest variant on both sets is the mean cosine with the full query (Table~\ref{tab:ablations-extra}). Every overall interval excludes zero under strict labels, and under expanded labels all but the mean cosine with the full query on near-miss queries (upper bound $+0.000$), but the per-$k$ intervals do not at $k=1$ (S1+$k$Q $-$ S0 is $-0.006$ $[-0.032, +0.018]$ nested and $+0.015$ $[-0.008, +0.035]$ on near-miss queries). A repeat run with a fresh cache moves nDCG@10 by at most 0.019 (Table~\ref{tab:ablations-extra}, details in Appendix~\ref{app:stats}).

\paragraph{The first-stage embedder check in detail.} The \texttt{text-embedding-3-large} index of Table~\ref{tab:ablations}f, which ranks above \texttt{text-embedding-3-small} on MTEB(eng,~v2) \citep{enevoldsen2025mmteb,chung2025maintainingmteblongterm}, keeps the descriptions, the S0 and S1 settings, the splits, the prompts and the verifier of the main runs. Its expanded labels would be incomplete: the judge pool came from the original systems, so 13.2--15.1\% of the top-10 slots of S4@50$_\mathrm{L}$ and S5$_\mathrm{L}$ (18--22\% for S1$_\mathrm{L}$ and S2@50$_\mathrm{L}$) were never judged. The two first stages share about half their top 10 (48--55\%). S5's pool holds the source for 68.6\% of nested queries against 66.9\% on the main index, and for 63.7\% of near-miss queries on both. After reranking, the embedder changes S2@50, S4@50, S3 and S5 by $+0.005$ to $+0.021$, every interval including zero, and on frames the whole-query and per-constraint checks still tie (S5$_\mathrm{L}$ $-$ S3$_\mathrm{L}$ is $+0.000$ and $+0.002$). The BM25 first stage (Okapi BM25 with $k_1=1.2$, $b=0.75$, no stemming, all fixed in advance) keeps the same descriptions, splits, prompts and verifier. Its source recall@10 and @50 are 0.44 and 0.60, against 0.52 and 0.70 for the main embedder and 0.52 and 0.72 for the large one, and S5's pool holds the source for 56\% of queries against 65\%. Splitting at retrieval loses as much as on the main index (difference $+0.008$ $[-0.027, +0.043]$, pooled); the whole-query lead is $0.030$ $[0.011, 0.049]$ smaller and the frame gain $0.015$ $[0.001, 0.030]$ smaller, but both stay positive, and with the source in S5's pool the frame gain is nearly unchanged ($+0.091$ against $+0.101$), so it shrinks mostly because fewer sources reach the pool.

\begin{table*}[tb]
\centering
\small
\setlength{\tabcolsep}{3pt}
\renewcommand{\arraystretch}{1.08}
\begin{tabular}{@{}lllcc@{}}
\toprule
Ablation & Contrast & $n$ & Nested & Near-miss \\
\midrule
\multicolumn{5}{@{}l}{\emph{Decomposition at retrieval} (Section~\ref{sec:res-rq1})} \\
Best of five fusion variants & S1 (mean cos.\,$+$\,Q) $-$ S0 & 350 / 372 & $-0.034$ {\footnotesize[$-0.063$, $-0.005$]} & $-0.021$ {\footnotesize[$-0.044$, $+0.000$]} \\
Query added with weight $k$ & S1$+k$Q $-$ S0 & 350 / 372 & $-0.035$ {\footnotesize[$-0.056$, $-0.015$]} & $-0.043$ {\footnotesize[$-0.069$, $-0.018$]} \\
Exact split, $k=2$, 3, 4 & S1 $-$ S0 & 54, 41, 34 / -- & $-0.086$, $-0.073$, $-0.085$ & -- \\
Gold constraints (dev.)$^d$ & S2b@20 gold $-$ S2b@20 & 366 / -- & $+0.030$ {\footnotesize[$+0.010$, $+0.053$]} & -- \\
\addlinespace[2pt]
\multicolumn{5}{@{}l}{\emph{Repeat run} (fresh cache)} \\
Description rerank & S4@50 rerun $-$ S4@50 & 52 / 64 & $-0.018$ {\footnotesize[$-0.042$, $+0.000$]} & $+0.018$ {\footnotesize[$-0.013$, $+0.057$]} \\
Frame rerank (contested) & S5 rerun $-$ S5 & -- / 44 & -- & $-0.005$ {\footnotesize[$-0.016$, $+0.000$]} \\
\addlinespace[2pt]
\multicolumn{5}{@{}l}{\emph{Human-written queries: MultiVENT~2.0, descriptions only, depth 20} (Appendix~\ref{app:results}); 400 queries} \\
Whole-query rerank & S4 $-$ S0 & 400 & \multicolumn{2}{c}{$+0.071$ {\footnotesize[$+0.054$, $+0.089$]}} \\
Binary per-constraint check & S2b $-$ S0 & 400 & \multicolumn{2}{c}{$-0.001$ {\footnotesize[$-0.012$, $+0.011$]}} \\
Whole-query rerank, $\ge4$ constraints & S4 $-$ S0 & 88 & \multicolumn{2}{c}{$+0.029$ {\footnotesize[$-0.006$, $+0.063$]}} \\
\bottomrule
\end{tabular}
\caption{Additional ablations, in the format of Table~\ref{tab:ablations}: paired differences in nDCG@10 with 95\% cluster-bootstrap intervals over source videos (MultiVENT block: over queries), expanded labels unless noted. $^d$Development set, strict labels, rerank depth 20: all 366 nested development queries. The exact-split cells are point estimates (intervals in the paragraph on splitter errors).}
\label{tab:ablations-extra}
\end{table*}

\paragraph{The ordering of S4 over S2b holds with a second verifier.} The \texttt{gpt-5.5} verifier repeats S2b and S4 on a seeded subset of 52 nested and 64 near-miss test queries, with the same S0 rankings and splits. Section~\ref{sec:results} reports S4@50 $-$ S2b@50 on each set; per $k$ ($n=16$, 8, 14, 14 and 18, 14, 15, 17) it is $+0.119$, $+0.264$, $+0.308$, $+0.258$ and $+0.070$, $+0.116$, $+0.160$, $+0.127$, every interval above zero but wide. The two verifiers' S2b verdicts agree at $\kappa=.77$ (nested) and $.82$ (near-miss).

\paragraph{The graded verifier under the second verifier.} \texttt{gpt-5.5} also repeats S2@50 on a seeded 150-query subset of each set (seed 0 inside the 100 test sources), with the same S0 rankings and splits. On these 150 queries S2@50 under \texttt{gpt-5.5} scores $-0.025$ $[-0.055, +0.002]$ (nested) and $-0.011$ $[-0.034, +0.012]$ (near-miss) nDCG@10 relative to S2@50 under \texttt{gpt-6-sol}, and on the 25,350 and 18,500 (query, candidate, constraint) grades the two verifiers agree at Spearman .80 / .90, exactly for .77 / .78 of the grades and within $\pm1$ for .83 / .86, with $\kappa=.92$ / .94 on ``stated'' ($\ge8$) and .75 / .78 on ``contradicted'' ($\le2$). Across verifiers, S4@50 under \texttt{gpt-6-sol} leads S2@50 under \texttt{gpt-5.5} by $+0.141$ $[+0.093, +0.193]$ and $+0.081$ $[+0.049, +0.113]$, and S2@50 under \texttt{gpt-5.5} leads S2b@50 under \texttt{gpt-6-sol} by $+0.062$ $[+0.003, +0.124]$ and $+0.048$ $[+0.015, +0.082]$. S4@50 and S2b@50 under \texttt{gpt-5.5} exist only on the 52 / 64 subsets above, of which 21 nested and 25 near-miss queries fall in the 150-query subsets, so the same-verifier graded contrasts have $n=21$ / 25: S4@50 $-$ S2@50 is $+0.106$ $[-0.005, +0.228]$ and $+0.061$ $[-0.022, +0.143]$ and S2@50 $-$ S2b@50 is $+0.115$ $[+0.023, +0.204]$ and $+0.025$ $[-0.040, +0.094]$ (Holm $p=.44$, $.89$, $.15$ and $.92$ over the 13 primary graded contrasts; under strict labels $+0.064$ $[-0.066, +0.197]$, $+0.045$ $[-0.036, +0.126]$, $+0.157$ $[+0.040, +0.288]$ and $+0.018$ $[-0.043, +0.085]$).

\paragraph{Reranking the top 50 beats the top 10, at a higher cost per point of gain.} S4@50 beats S4@10 by $+0.107$ $[+0.073, +0.146]$ (nested) and $+0.098$ $[+0.071, +0.126]$ (near-miss), roughly half of S4's gain over S0. Per $+0.01$ nDCG@10 over S0 (expanded labels), S4@10 needs about 172 (nested) and 267 (near-miss) input tokens (est.), S4@50 about 457 and 590, S2b@10 about 1,152 and 1,322 (est.), and S2b@50 about 3,545 and 2,840, so the shallow rerank is about 2.2 to 2.7 times as efficient as the deep one, and S2b@50 costs about 7.8 (nested) and 4.8 (near-miss) times as much as S4@50 per point of gain. The S4@10 and S2b@10 token counts (marked est.) are estimates: S4@10 and S2b@10 rescore the depth-50 calls at depth 10, so their cost is taken as 10 of the 50 per-candidate calls, since per-call tokens are not saved.

\paragraph{S2b does not lower the HN-above-source rate.} On the 96 contested queries the HN video ranks above the source in .292 $[.17, .42]$ of S2b@10's rankings against .219 $[.11, .34]$ for S0; S4@50 gives .208 $[.11, .32]$, S3b .042 and S5 and S3 .010 $[.00, .03]$. Over all 372 near-miss queries the rate is .234 $[.170, .299]$ for S2b@50 and .212 $[.157, .271]$ for S2@50 against .196 $[.135, .261]$ for S0, .177 $[.119, .240]$ for S4@50 and .126 $[.078, .179]$ for S5 and S3. The rate counts a video missing from the saved ranking as last.

\paragraph{A second frame verifier gives the same frame gain (supporting check only).} Replacing the S5 verifier \texttt{gpt-6-sol} by \texttt{gemini-3.8-flash} (low thinking effort, medium media resolution) on the same candidates gives S5 $-$ S4@50 of $+0.133$ $[+0.076, +0.197]$ on the 96 contested test queries against $+0.134$ $[+0.074, +0.200]$ for \texttt{gpt-6-sol} (expanded labels); the two verifiers differ by $-0.001$ $[-0.008, +0.005]$. On the 56 nested-sample queries the gains are $+0.043$ $[+0.016, +0.073]$ and $+0.047$ $[+0.019, +0.078]$, a difference of $-0.004$ $[-0.013, +0.000]$; under strict labels the contested gains are $+0.138$ and $+0.129$ (difference $+0.009$ $[-0.001, +0.024]$). Their top-1 choice (the same rank-1 video) agrees for 96.9\% $[92.0, 100]$ of the contested queries and 80.4\% $[69.1, 90.3]$ of the nested-sample queries. This is a cross-model check on the same candidates, not a repeat run. The second verifier is a Gemini model, like the describer and one of the grounders and judges, so it is not independent of the index or of the labels. It was called by the unpinned alias \texttt{gemini-3.8-flash} and may not reproduce (Appendix~\ref{app:release}), so it is a supporting check only.

\paragraph{MultiVENT~2.0 check.} As a check on human-written queries \citep{kriz2024multivent}, we run two description rerankers; this is a description-only ablation (no frame system). The corpus is a 2,547-video English subset of MultiVENT~2.0; the index is the dataset's shipped Qwen3-Omni descriptions, not ours, and S0 is the Jasper-Token-Compression-600M first stage. S2b and S4 rerank S0's top 20 with the splitter, binary description-check and whole-query rerank prompts (\texttt{gpt-6-luna} splitter, \texttt{gpt-6-sol} verifier at low effort). The 400 queries are the 100 test and 300 development queries; nDCG@10 uses the graded judgments (0--3), the constraint count $k$ is counted by \texttt{gpt-6-luna}, and intervals are 95\% bootstrap intervals over queries (5,000 resamples). nDCG@10 is .749 $[.719, .777]$ for S0, .748 $[.718, .776]$ for S2b and .820 $[.795, .844]$ for S4. S4 $-$ S0 is $+0.071$ $[+0.054, +0.089]$ and S2b $-$ S0 $-0.001$ $[-0.012, +0.011]$ (Table~\ref{tab:ablations-extra}); S2b $-$ S4 is $-0.072$ $[-0.091, -0.054]$. By constraint count ($n=69$, 85, 158 and 88 for $k=1$, 2, 3 and 4+), S4 $-$ S0 is $+0.099$ $[+0.058, +0.147]$, $+0.082$ $[+0.049, +0.120]$, $+0.077$ $[+0.048, +0.108]$ and $+0.029$ $[-0.006, +0.063]$, and S2b $-$ S0 is $+0.036$ $[+0.011, +0.072]$, $+0.012$ $[+0.003, +0.024]$, $-0.006$ $[-0.025, +0.011]$ and $-0.032$ $[-0.062, -0.008]$. The queries are short news-event headlines (dates and places name the event; Section~\ref{sec:limitations}), and $k$ here counts entities, events, places and dates as well, so it is not comparable to our $k$. These constraints are often carried by speech or on-screen text rather than by a single frame, and S0's nDCG@10 does not rise with $k$ here (Spearman $+0.09$), so the check does not test queries like ours. It keeps the direction of the S4 over S2b ordering (RQ1) and says nothing about S1 or frames. The comparison is also confounded with score scale (S2b is binary), the prompts were written for visual queries, all 400 queries were used in pilot runs of an earlier routing study (not to tune these prompts), and judged@10 is only about .33--.36 with judgments pooled from other systems, so we treat it as a weak external check and do not use it as evidence in the main text.

\paragraph{Cross-encoders.} We first tried \texttt{bge-reranker-v2-m3} \citep{chen2024bgem3}, a 568M-parameter cross-encoder. It does not move S0 under strict labels ($+0.002$ $[-0.051, +0.056]$ nested, $-0.030$ $[-0.078, +0.017]$ near-miss), so we added the three stronger models of Table~\ref{tab:results}, Qwen's and LightOn's rerankers at 2B, the largest we could run locally on a laptop. We also tried the late-interaction encoder pplx-embed-v2-late-0.6b \citep{krimmel2026pplxembed}, which encodes query and description apart and scores them by MaxSim over token vectors; it gains less ($+0.063$ $[+0.017, +0.110]$ and $+0.094$ $[+0.061, +0.126]$ over S0) and is not a cross-encoder, so it is left out of the main results. We tuned none of the five. All read the descriptions that S4 reads, on S0's top 50, zero-shot; Qwen-2B uses its default instruction. We took the 2B list-wise model rather than the 0.8B because the LightOn family improves with size. LW-2B ranks four descriptions per prompt in a window that slides up the list with stride 2, as its model card prescribes, so it reads 24 prompts per query; Qwen-2B and PW-2B score each description on its own. Counted with each model's own tokenizer, Qwen-2B reads 7.1k input tokens per query, PW-2B 6.5k and LW-2B 7.0k, plus about 0.4k it writes, against 10.4k for S4@50. A second label fill judged every top-10 slot of the three and of the late-interaction encoder (Appendix~\ref{app:mining}); bge was not filled, so for it we quote strict labels only. Only Qwen-2B ranks above S2 on both sets (Table~\ref{tab:local}). PW-2B and LW-2B are level with S2 on nested queries, and the list-wise LW-2B is level with PW-2B on nested queries and below it on near-miss ones ($-0.034$ $[-0.055, -0.014]$). Under strict labels S4@50 leads all three on nested queries, but on near-miss ones it is level with Qwen-2B ($+0.011$ $[-0.017, +0.039]$) and PW-2B ($+0.017$ $[-0.015, +0.049]$). These contrasts were added after the main runs and are not in the Holm family.

\begin{table*}[!ht]
\centering
\footnotesize
\setlength{\tabcolsep}{2pt}
\begin{tabular}{@{}lcccccc@{}}
\toprule
 & \multicolumn{2}{c}{Model $-$ S0} & \multicolumn{2}{c}{S4@50 $-$ model} & \multicolumn{2}{c}{S2 $-$ model} \\
\cmidrule(lr){2-3}\cmidrule(lr){4-5}\cmidrule(l){6-7}
Model & Nested & Near-miss & Nested & Near-miss & Nested & Near-miss \\
\midrule
Qwen-2B & $+.170$\,{\scriptsize[$+.124$, $+.217$]} & $+.136$\,{\scriptsize[$+.097$, $+.176$]} & $+.059$\,{\scriptsize[$+.035$, $+.084$]} & $+.040$\,{\scriptsize[$+.014$, $+.066$]} & $-.042$\,{\scriptsize[$-.081$, $-.005$]} & $-.055$\,{\scriptsize[$-.101$, $-.012$]} \\
PW-2B & $+.112$\,{\scriptsize[$+.067$, $+.159$]} & $+.132$\,{\scriptsize[$+.094$, $+.170$]} & $+.117$\,{\scriptsize[$+.087$, $+.151$]} & $+.044$\,{\scriptsize[$+.012$, $+.075$]} & $+.016$\,{\scriptsize[$-.022$, $+.053$]} & $-.051$\,{\scriptsize[$-.097$, $-.007$]} \\
LW-2B & $+.104$\,{\scriptsize[$+.067$, $+.144$]} & $+.098$\,{\scriptsize[$+.068$, $+.129$]} & $+.125$\,{\scriptsize[$+.093$, $+.159$]} & $+.078$\,{\scriptsize[$+.046$, $+.109$]} & $+.023$\,{\scriptsize[$-.018$, $+.063$]} & $-.017$\,{\scriptsize[$-.065$, $+.028$]} \\
\bottomrule
\end{tabular}
\caption{Cross-encoders on S0's top 50 descriptions: paired differences in nDCG@10 (expanded labels) with 95\% cluster-bootstrap intervals over sources (10,000 resamples), $n=350$ nested and 372 near-miss queries.}
\label{tab:local}
\end{table*}

\paragraph{Dropping the near-miss queries at $k=1$ changes no conclusion.} At $k=1$ the HN video meets none of the constraints, so it is not a near miss; without these queries ($n=274$) S4@50 $-$ S0 is $+0.162$ $[+0.122, +0.203]$, S4@50 $-$ S2b@50 $+0.134$ $[+0.099, +0.170]$ and S1 $-$ S0 $-0.148$ $[-0.202, -0.098]$; on the contested subgroup at $k\ge2$ ($n=93$) S5 $-$ S4@50 is $+0.138$ $[+0.076, +0.204]$, and the per-$k$ cells on all near-miss queries are in Table~\ref{tab:frames}.

\paragraph{The frame gain does not rest on one fact type or corpus part.} Nested facts are 50\% attributes, 30\% objects, 10\% on-screen text, 6\% scenes and 4\% counts (833 fact instances); near-miss facts are 19\%, 53\%, 13\%, 14\% and 2\% (900). On the contested queries, S5 $-$ S4@50 is $+0.249$ $[+0.119, +0.395]$ when the HN video's distinct fact is an attribute ($n=31$), $+0.077$ $[+0.015, +0.158]$ for an object ($n=33$) and $+0.092$ $[+0.019, +0.195]$ for on-screen text ($n=29$; counts $n=3$). By corpus part, over all near-miss queries, S5 $-$ S4@50 is $+0.086$ $[+0.033, +0.149]$ on CaReBench ($n=59$), $+0.065$ $[-0.004, +0.157]$ on Shot2Story (57), $+0.079$ $[+0.046, +0.117]$ on Vript long (151) and $+0.083$ $[+0.043, +0.131]$ on Vript short (105), and over all nested queries $+0.096$, $+0.045$, $+0.087$ and $+0.051$ (Table~\ref{tab:frames}); on the contested queries alone it is $+0.166$ $[+0.021, +0.326]$ on CaReBench ($n=21$), $+0.062$ $[-0.027, +0.288]$ on Shot2Story (10), $+0.143$ $[+0.056, +0.248]$ on Vript long (45) and $+0.116$ $[+0.015, +0.233]$ on Vript short (20), every interval but Shot2Story's above zero. The gain appears on the two corpora with human-written captions as on Vript, so it is not specific to GPT-4V captions, but all four share the Gemini descriptions, grounders and judges.

\paragraph{Frame gain on stated and omitted facts.} We split the per-constraint grades of the graded checks by whether the text-only judge marks the constraint's fact as stated in the source's description. S3 recovers constraints that the description does not state: S2 meets $.29$ of them and S3 $.94$ ($n=509$ query-fact rows, 241 facts, 78 sources; difference $+.644$ $[.587, .703]$, and $+.631$ $[.569, .691]$ with one row per fact). On the few stated constraints S2 is already at ceiling ($.99$, $n=84$ rows, 38 facts, 28 sources), so the data cannot show whether frames add anything there (difference $+.012$ $[.000, .039]$). Five limits apply. The analysis covers nested queries whose source is in S4@50's top 10 only, and a constraint is met when its grade is above 5; at a grade of 8 the stated difference is $+.060$ $[.000, .143]$ and the omitted one $+.849$ $[.805, .891]$. Nearly all omitted constraints are \emph{unclear} in the description (98\%), so recovery from a description that gets a detail wrong is not tested. The line between stated and unclear is a text-only judge label and is soft: using the Option~A labels moves about 2\% of rows. Constraints were matched to mined facts by token overlap, because the planned LLM matcher was not run on the test set. The near-miss split is not reported, because the judge was never run on near-miss facts or on hard-negative descriptions.

\paragraph{Source acceptance falls with $k$ on nested queries.} Nested 89, 71, 61 and 51\% for $k=1$--4, near-miss 93, 92, 88 and 86\%; 100 of the 107 rejected nested queries fail on a fact they share with another nested query of the same source, and 90 fail only on such a shared fact, so one judge error on one fact rejects every nested query that contains it. The fall with $k$ is mostly this compounding: at least one judge misses 57 of the 350 nested test facts (16\%), at a similar rate whether a fact enters at low or high $k$, so a query that needs all $k$ facts accepted passes about $0.84^k$ of the time ($0.49$ at $k=4$, against $0.51$ observed). Most misses are one judge's ``no'' (Gemini more often than \texttt{gpt-6-astra}), not ``unsure'' (counting unsure as yes lifts $k=4$ only to 53\%), and they cluster on long videos: 10--12\% of facts are missed on sources under a minute, 26--31\% on longer ones, where 16 sampled frames are sparse. Part of the strictness is the prompt: on pilot pairs, a softer judge prompt that asks for the evidence first and accepts synonyms raised source acceptance from 0.84 to 0.91 (nested 0.75 to 0.86) but lowered the judges' agreement on other videos ($\kappa$ 0.86 to 0.83), so we kept the original prompt. A follow-up ablation on the 100 test sources shows that the frame budget explains part of the gap. With 64 instead of 16 frames, nested $k=4$ source acceptance rises from 0.48 (a same-settings re-run; 0.51 in the labelling pass) to 0.64 ($+0.16$ $[+0.07, +0.27]$), and per-fact acceptance on videos over a minute rises by $+0.06$ $[+0.01, +0.11]$, while the judges' precision against humans off the source stays within run-to-run noise (0.83 against 0.85). Sending 768-pixel frames gains nothing, and the softer prompt, though it raises acceptance most, lowers that precision by 0.05. The labels therefore keep the 16-frame judge with the original prompt, and its source rejections remain a conservative error. The annotator finds 55 of the 56 rejected nested facts they could decide present on the source, but the second annotator's overlap holds too few of them to check this, so the claim that most rejections are judge errors rests on one annotator. Off the source, the human items are each disagreement query's two top videos and the HN video, not a sample of the judged pool (Appendix~\ref{app:human}).

\paragraph{The main contrasts survive a Holm correction; the subgroup cells are exploratory.} With a Holm correction over the ten main contrasts of the binary systems, the six description contrasts and both S5 $-$ S4@50 contrasts (contested, nested sample) stay significant at Holm $p=.002$, and neither S3b $-$ S5 contrast is (contested $p=.206$; nested sample $p=.084$, Holm $.168$; strict contested $p=.031$, Holm $.062$). Over the primary contrasts of the graded systems, S4@50 $-$ S2@50 on both sets and S2@50 $-$ S2b@50 on the nested set are significant at $p<.001$, S2@50 $-$ S2b@50 on the near-miss set is not ($p=.054$, Holm $.430$ over the 13 primary graded contrasts, which include the four second-verifier contrasts above), S5 $-$ S3 is not significant anywhere, and S5 $-$ S4@50 pooled over all 722 queries is ($p<.001$). The 40 subgroup cells are S5 $-$ S4@50 and S3b $-$ S5 per $k$ and per Vript length bin on contested queries and on the nested sample (32 cells) and S4@50 $-$ S2b@50 per $k$ on each set (8 cells; the length-bin cells of Table~\ref{tab:perk} were not included). Of these, 11 stay significant under expanded labels: the eight S4@50 $-$ S2b@50 cells and S5 $-$ S4@50 on contested queries at $k=2$, 3 and 4 (no length-bin cell), so the subgroup cells should be read as exploratory.

\paragraph{Output tokens are small next to the frame input tokens.} Output tokens, which include the verifiers' reasoning tokens, are 1.3--1.4k per query for S4@50, 3.6--4.6k for S2b@50, 3.4--4.3k for S2@50 and 0.3k (S5), 0.9k (S3) or 0.9--1.1k (S3b) for the frame step.

\paragraph{S0's score rises with $k$ on nested queries and not on near-miss queries.} Nested S0 nDCG@10 goes from .339 at $k=1$ to .485 at $k=4$, likely because each added fact makes the generated query more specific to its source; on the near-miss set, where the HN video meets $k-1$ constraints, S0 stays between .436 and .461 (.455 at $k=1$, .447 at $k=4$). This is why the near-miss set is added to the nested set.

\paragraph{No detectable trend of the frame gain with length or with $k$.} We regress the per-query S5 $-$ S4@50 nDCG@10 on log duration within Vript (507 test queries, 70 sources), with the covariates of Appendix~\ref{app:stats}, and report the slope per doubling with a 95\% cluster-bootstrap interval (10,000 resamples of sources) and the minimal slope detectable at 80\% power ($2.8\times$ the bootstrap standard error). The expanded-label slope is in Finding~\ref{sec:res-length} (standard error 0.016, detectable slope 0.045); under strict labels it is $-0.013$ $[-0.045, +0.019]$ (0.016, 0.046). Slopes above about $+0.02$ per doubling are thus outside both intervals, but smaller ones, up to about $+0.06$ over the three doublings from 30\,s to 4\,min, are not excluded, and the design has 80\% power only for slopes of about 0.05. S3 $-$ S4@50 gives $-0.011$ $[-0.044, +0.020]$ and S4@50 $-$ S0 $-0.032$ $[-0.076, +0.014]$ (expanded). For the number of constraints, an OLS of the same per-query gain on $k$ (with a set indicator when pooled; cluster bootstrap as above) gives, per added constraint, $+0.006$ $[-0.006, +0.017]$ under strict labels over all queries except the near-miss ones at $k=1$ ($n=624$; expanded in Finding~\ref{sec:res-length}); nested alone $-0.005$ $[-0.018, +0.007]$ (strict $+0.005$ $[-0.009, +0.019]$). With the near-miss $k=1$ queries kept the pooled slope is $+0.007$ $[-0.002, +0.016]$ under expanded and $+0.013$ $[+0.003, +0.023]$ under strict labels, and the near-miss slope alone $+0.019$ $[+0.004, +0.034]$; this rise comes from $k=1$, since on near-miss queries at $k\ge2$ it is $+0.004$ $[-0.021, +0.028]$ (strict $+0.007$ $[-0.020, +0.034]$).

\paragraph{The description leaves more detail unconfirmed in longer videos, but the trend is only suggestive.} The index frames show 93\% $[91.6, 95.1]$ of grounded facts overall (941 facts), 95\% under 30\,s, 96\% at 30--60\,s, 90\% at 1--2\,min and 91\% at 2--4\,min, as checked by Gemini. On the nested set, the description leaves 72\% of a query's conditions unconfirmed under 30\,s, 79\% at 30--60\,s, 90\% at 1--2\,min and 94\% at 2--4\,min (79\% $[73.1, 84.1]$ overall; almost all ``unclear''; counted only when the source is in S0's top 50); the conditions of near-miss queries are coarser and stay at 46--53\% (47\% overall). After controlling for $k$ and corpus part, the duration slope of this drop is $+0.047$ $[-0.016, +0.117]$, which is not significant. Humans reading the descriptions (Task E, App.~\ref{app:human}) give the same picture: on average 85\% $[80, 90]$ of a nested query's facts and 53\% $[47, 59]$ of a near-miss query's are not stated in the description (implied or absent); on the same per-query measure for nested Vript queries, humans' share rises like the verifier's, from 81\% under 30\,s to 87\% at 30--60\,s, 93\% at 1--2\,min and 97\% at 2--4\,min, with overlapping intervals, so humans show the same suggestive rise (with implied counted as not stated).

\paragraph{The margin trigger separates less well on the test set than on the development set.} Sweeping the threshold traces the share of S5's gain kept against the share of queries sent to frames (Figure~\ref{fig:trigger}); a random trigger gives an area of 0.5. The area under the S4 top-two margin curve is .66 over all test queries (.64 nested, .67 near-miss, .73 on the contested subgroup) against .78 on the development set.

\paragraph{The trigger over all queries, and three simpler rules.} Over all 722 queries, the margin trigger at the development threshold sends 81.4\% of nested and 73.1\% of near-miss queries to frames and keeps 97.3\% $[93.4, 99.9]$ and 99.2\% $[94.8, 105.2]$ of S5's gain (117 of 120 and 110 of 115 top-1 changes caught), at 71.3k input tokens per query against 90.7k for S5 on every query. $k\ge3$, which needs no score, sends 45.1\% of nested and 47.3\% of near-miss queries and keeps 38.6\% and 56.6\% of the gain; a threshold on S4@50's top score (below 8), which reads the same S4@50 scores as the margin, sends 87.1\% and 54.8\% and keeps 97.5\% and 94.4\%, level with the margin rule at the same pooled share (Finding~\ref{sec:res-trigger}; over the whole sweep its area is .63 against the margin's .66 over all queries and .70 against .67 on near-miss queries); a random trigger at the margin rule's share (77\%) keeps 73.3\% $[65.5, 80.2]$ pooled. On the contested subgroup the margin trigger fires on 60 of 96 queries (62.5\%) and catches all 27 top-1 changes; on the nested sample it fires on 46 of 56 (82.1\%) and catches all 16; on the 570 queries outside the two subgroups it sends 79.1\% and keeps 97.7\% $[93.9, 102.3]$ (184 of 192 top-1 changes). Pooled, the development threshold's shares in Finding~\ref{sec:res-trigger} have intervals $[71.6, 82.3]$ sent and $[95.4, 101.7]$ kept (pointwise cluster-bootstrap intervals; kept can exceed 100\% when the trigger beats S5 on a resample). The cross-fit threshold (five folds over sources, 20 repetitions with different fold assignments, all giving the same point) is picked per fold as the cheapest threshold that keeps the target share of S5's gain on the other four folds, then applied to the held-out fold, so no source helps set the threshold applied to it; the held-out decisions are pooled. At the 80\% target all 100 fold fits pick the same threshold, a margin of 1 point; per held-out fold of about 20 sources the trigger sends 64.3\% of queries on average (standard deviation 5.7 points, range 48.6--76.4\%) and keeps 91\% of the gain (90.5\% pooled over all 722 queries; standard deviation 5.6 points; range 78.5--114\%, since a small fold's gain is noisy). So the drop from the development point (98.3\% kept, 77.1\% sent, at a margin of 2) is a cheaper operating point chosen by the 80\% target, not instability; at a 90\% target the folds split between margins 1 and 2 (65 and 35 of 100 fits). The resampling intervals below show the spread across sources. It reads 60.8k input tokens per query at a kept-gain target of 80\%, and at a 90\% target it sends 68.5\% and keeps 92.1\%. Repeating the cross-fit inside each of 2,000 resamples of sources gives 95\% intervals of 58.5--70.2\% sent, 85.6--95.4\% kept and 54.6k--68.2k tokens at the 80\% target, and 60.4--80.1\% sent and 90.7--98.7\% kept at 90\%. A fixed budget that sends the $x\%$ of queries with the smallest S4@50 margin (ties to the lower top score, then at random) keeps the same gain as the threshold rule at the rule's own shares, because the margin is integer-valued, and can hit any share in between: it keeps 77.5\% $[69, 86]$ at 50\% sent and 88.0\% $[82, 93]$ at 60\%, and at 30\% the top-score tie-break keeps 60.8\% against 48.0\% for the interpolated rule. It needs the margin distribution of the batch, or a cutoff carried over from an earlier batch. Because it orders queries by the same S4@50 margin, it is a variant of the margin trigger, not an independent baseline. The intervals on the absolute means of S4@50 and S5 in Figure~\ref{fig:trigger} overlap even where paired differences exclude zero (Figure~\ref{fig:contrasts}, Table~\ref{tab:results}).

\paragraph{Frame gain by where the source sits.} The source is outside S4@50's top 10 for 135 of the 372 near-miss queries (36\%) and 116 of the 350 nested queries (33\%), where the strict-label gain is zero by construction. Where the source is in the top 10 ($n=234$ nested, 237 near-miss), S5 $-$ S4@50 is $+0.097$ $[+0.060, +0.139]$ and $+0.110$ $[+0.074, +0.150]$ (expanded labels); where it is not, $+0.028$ $[+0.014, +0.044]$ and $+0.024$ $[+0.014, +0.036]$, from other relevant videos (only 105 of these 251 queries have any relevant video in the pool). The in-pool queries supply 87.6\% $[77.6, 94.2]$ and 89.0\% $[82.4, 93.9]$ of the gain (pooled 88.4\% $[83.3, 92.5]$). Sorting S4@50's top 10 by label, the best any reranker of this pool can do, gives .717 (nested) and .715 (near-miss) against S5's .710 and .705 (pooled .716 against .707), so S5 takes 91.3\% $[84.3, 95.9]$ and 88.7\% $[81.3, 93.8]$ of the headroom over S4@50. Under strict labels this ceiling is .652 against .616 pooled (64.4\% $[53.5, 73.8]$ of the headroom; 71.9\% nested, 59.1\% near-miss), so a better frame verifier could add at most $+0.009$ (expanded) or $+0.037$ (strict) on this pool. Unjudged candidates count as non-relevant in the expanded-label ceiling. The 237 near-miss queries with the source in the top 10 are the contested queries (96, 26\% of near-miss queries) plus the 141 (38\%) with the source but not the HN video there. Queries that S4 did not select give smaller gains than contested queries: $+0.060$ $[+0.038, +0.084]$ on the 276 unselected near-miss queries, $+0.079$ $[+0.051, +0.110]$ on the 294 unselected nested queries and $+0.047$ $[+0.019, +0.078]$ on the nested sample; in development the gain was $+0.028$ $[+0.010, +0.049]$ on a random 150-query nested subset (strict, against S4@10) and $+0.029$ $[-0.010, +0.076]$ on 40 near-miss queries matched to nested ones (expanded, against S4@10, S0's top 10).

\paragraph{Frame gain on the development queries (semi-held-out).} The 120 development near-miss queries on which the trigger threshold was set (51 development sources, disjoint from the test sources; queries whose source and HN video are both in S0's top 10) have saved S4@50 and S5 rankings. On them S5 $-$ S4@50 is $+0.059$ $[+0.022, +0.104]$ under expanded and $+0.046$ $[+0.003, +0.093]$ under strict labels (95\% cluster-bootstrap intervals over sources, 10,000 resamples), against $+0.134$ $[+0.074, +0.200]$ and $+0.129$ $[+0.071, +0.195]$ on the 96 contested test queries. Two caveats apply. First, on these queries S5 reranked S0's top 10 (the planned pool), not S4@50's, so the contrast mixes the two pools; S4@50 has the source in its top 10 for 119 of the 120 queries and differs from S4@10 by $-0.002$ $[-0.009, +0.003]$. Second, S4@50 scores .902 here (expanded), so the headroom is smaller than on test. Reproduction check: the same code gives the development report's strict S5 $-$ S4@10 of $+0.043$ (interval $[+0.001, +0.090]$ here against $[+0.003, +0.093]$ there, from a different bootstrap draw).

\paragraph{Frames on the planned pool.} We ran S5 on S0's top 10, the pool planned before the runs, for all 722 test queries (nested 350, near-miss 372); S4@10 is S4's reordering of the same ten videos, so the primary contrast S5 $-$ S4@10 compares two rerankers of one pool that neither chose (Table~\ref{tab:s0pool}). S0's and S4@50's top tens share 5.2 videos on average. S5 $-$ S4@10 is $+0.039$ $[+0.028, +0.053]$ pooled under expanded labels ($+0.036$ $[+0.023, +0.051]$ nested, $+0.043$ $[+0.026, +0.062]$ near-miss) and $+0.035$ $[+0.022, +0.050]$ under strict labels. S5 takes 85\% $[77, 90]$ of the headroom over S4@10 up to the pool's oracle (S4@50's pool: 90\%). The source is in S0's top 10 for 376 of the 722 queries, which supply 83\% of the gain ($+0.063$ $[+0.041, +0.088]$ there, $+0.014$ $[+0.006, +0.021]$ elsewhere). On the 69 contested queries with source and HN video both in S0's top 10 (96 in S4@50's), S5 $-$ S4@10 is $+0.093$ $[+0.045, +0.154]$, and the HN video ranks above the source for 14.5\% of queries after S4@10 and 0\% after S5. The development check ($+0.059$, S5 on S0's top 10 minus S4@50, on 120 development contested queries) has the like-for-like test number $+0.083$ $[+0.018, +0.161]$ on these 69. The run paid for 24.1M input tokens (about \$49); 3,761 of its 7,220 frame calls were cache hits (3,740 from the S4@50 run, 21 from a five-query dry run).

\begin{table}[t]
\centering
\footnotesize
\setlength{\tabcolsep}{4pt}
\begin{tabular}{@{}lccc@{}}
\toprule
Ranking (nDCG@10) & Nested & Near-miss & Pooled \\
\midrule
S0 & .408 & .450 & .430 \\
S4@10 & .529 & .528 & .528 \\
S5 on S0's top 10 & .565 & .570 & .568 \\
Oracle, S0's top 10 & .570 & .580 & .575 \\
\midrule
S4@50 & .636 & .626 & .631 \\
S5 (S4@50's top 10) & .710 & .705 & .707 \\
Oracle, S4@50's top 10 & .717 & .715 & .716 \\
\bottomrule
\end{tabular}
\caption{Frames on the planned pool (expanded labels, all 722 test queries); S4@10 reorders S0's top 10.}
\label{tab:s0pool}
\end{table}

\begin{table}[t]
\centering
\small
\setlength{\tabcolsep}{1.5pt}
\begin{tabular}{@{}lcc@{}}
\toprule
 & Contested & Nested sample \\
 & ($n=96$) & ($n=56$) \\
\midrule
\multicolumn{3}{l}{\emph{nDCG@10}} \\
S2b$'$ & $0.720$ & $0.622$ \\
S4@50 & $0.809$ & $0.656$ \\
S3b & $0.923$ & $0.698$ \\
S5 & $0.943$ & $0.703$ \\
\midrule
\multicolumn{3}{l}{\emph{Paired difference, 95\% CI}} \\
S3b $-$ S2b$'$ & $+0.203$\,{\scriptsize[$+0.143$, $+0.271$]} & $+0.076$\,{\scriptsize[$+0.029$, $+0.126$]} \\
S4@50 $-$ S2b$'$ & $+0.090$\,{\scriptsize[$+0.051$, $+0.133$]} & $+0.034$\,{\scriptsize[$-0.002$, $+0.075$]} \\
S5 $-$ S4@50 & $+0.134$\,{\scriptsize[$+0.074$, $+0.200$]} & $+0.047$\,{\scriptsize[$+0.019$, $+0.078$]} \\
S3b $-$ S5 & $-0.020$\,{\scriptsize[$-0.053$, $+0.003$]} & $-0.005$\,{\scriptsize[$-0.013$, $+0.000$]} \\
\midrule
\multicolumn{3}{l}{\emph{Equivalence of S3b and S5 (paired TOST)}} \\
90\% CI & {\scriptsize[$-0.047$, $+0.002$]} & {\scriptsize[$-0.011$, $-0.000$]} \\
Smallest margin & $\pm0.047$ & $\pm0.011$ \\
\bottomrule
\end{tabular}
\caption{Like-for-like comparison on the frame candidates (expanded labels). S2b$'$ applies S2b's per-constraint description verdicts (from the S2b@50 run, no new calls) to S4@50's top 10, with S2b's yes-count scoring and tie-break. Intervals are paired cluster-bootstrap intervals over source videos. Equivalence (two one-sided tests at $\alpha=.05$): the 90\% interval of S3b $-$ S5 lies inside $\pm$ the margin; we give the smallest such margin, since no margin was set before the main run.}
\label{tab:rc-frames}
\end{table}

\begin{table*}[t]
\centering
\small
\setlength{\tabcolsep}{4pt}
\begin{tabular}{lrcccc}
\toprule
& & \multicolumn{2}{c}{All queries} & \multicolumn{2}{c}{Source accepted by both judges} \\
\cmidrule(lr){3-4}\cmidrule(lr){5-6}
Comparison & $n$ & Expanded & Strict & Expanded & Strict \\
\midrule
\multicolumn{6}{l}{\emph{Nested, all queries}} \\
S1 $-$ S0 & 243/350 & $-0.089$\,{\scriptsize[$-0.125$, $-0.055$]} & $-0.082$\,{\scriptsize[$-0.118$, $-0.048$]} & $-0.075$\,{\scriptsize[$-0.112$, $-0.039$]} & $-0.065$\,{\scriptsize[$-0.103$, $-0.030$]} \\
S2b@50 $-$ S0 & 243/350 & $+0.042$\,{\scriptsize[$+0.007$, $+0.080$]} & $+0.039$\,{\scriptsize[$+0.003$, $+0.077$]} & $+0.052$\,{\scriptsize[$+0.011$, $+0.097$]} & $+0.045$\,{\scriptsize[$+0.002$, $+0.091$]} \\
S4@50 $-$ S0 & 243/350 & $+0.229$\,{\scriptsize[$+0.186$, $+0.274$]} & $+0.207$\,{\scriptsize[$+0.162$, $+0.254$]} & $+0.221$\,{\scriptsize[$+0.174$, $+0.271$]} & $+0.194$\,{\scriptsize[$+0.147$, $+0.245$]} \\
S4@50 $-$ S2b@50 & 243/350 & $+0.187$\,{\scriptsize[$+0.149$, $+0.226$]} & $+0.168$\,{\scriptsize[$+0.129$, $+0.209$]} & $+0.169$\,{\scriptsize[$+0.128$, $+0.213$]} & $+0.150$\,{\scriptsize[$+0.108$, $+0.193$]} \\
\addlinespace[2pt]
\multicolumn{6}{l}{\emph{Near-miss, all queries}} \\
S1 $-$ S0 & 334/372 & $-0.104$\,{\scriptsize[$-0.145$, $-0.066$]} & $-0.113$\,{\scriptsize[$-0.156$, $-0.073$]} & $-0.107$\,{\scriptsize[$-0.151$, $-0.067$]} & $-0.117$\,{\scriptsize[$-0.163$, $-0.075$]} \\
S2b@50 $-$ S0 & 334/372 & $+0.050$\,{\scriptsize[$+0.017$, $+0.082$]} & $+0.024$\,{\scriptsize[$-0.013$, $+0.060$]} & $+0.054$\,{\scriptsize[$+0.019$, $+0.089$]} & $+0.024$\,{\scriptsize[$-0.016$, $+0.064$]} \\
S4@50 $-$ S0 & 334/372 & $+0.175$\,{\scriptsize[$+0.139$, $+0.211$]} & $+0.123$\,{\scriptsize[$+0.089$, $+0.157$]} & $+0.181$\,{\scriptsize[$+0.142$, $+0.220$]} & $+0.122$\,{\scriptsize[$+0.086$, $+0.158$]} \\
S4@50 $-$ S2b@50 & 334/372 & $+0.125$\,{\scriptsize[$+0.096$, $+0.156$]} & $+0.099$\,{\scriptsize[$+0.066$, $+0.133$]} & $+0.127$\,{\scriptsize[$+0.095$, $+0.160$]} & $+0.097$\,{\scriptsize[$+0.062$, $+0.135$]} \\
\addlinespace[2pt]
\multicolumn{6}{l}{\emph{Contested (frame subset)}} \\
S5 $-$ S4@50 & 88/96 & $+0.134$\,{\scriptsize[$+0.074$, $+0.200$]} & $+0.129$\,{\scriptsize[$+0.071$, $+0.195$]} & $+0.123$\,{\scriptsize[$+0.066$, $+0.189$]} & $+0.115$\,{\scriptsize[$+0.062$, $+0.179$]} \\
S3b $-$ S5$^\dagger$ & 88/96 & $-0.020$\,{\scriptsize[$-0.053$, $+0.003$]} & $-0.023$\,{\scriptsize[$-0.052$, $-0.001$]} & $-0.011$\,{\scriptsize[$-0.041$, $+0.005$]} & $-0.015$\,{\scriptsize[$-0.038$, $+0.000$]} \\
S3b $-$ S2b$'$ & 88/96 & $+0.203$\,{\scriptsize[$+0.143$, $+0.271$]} & $+0.209$\,{\scriptsize[$+0.145$, $+0.280$]} & $+0.196$\,{\scriptsize[$+0.135$, $+0.264$]} & $+0.200$\,{\scriptsize[$+0.134$, $+0.271$]} \\
\addlinespace[2pt]
\multicolumn{6}{l}{\emph{Nested sample (frame subset)}} \\
S5 $-$ S4@50$^\dagger$ & 38/56 & $+0.047$\,{\scriptsize[$+0.019$, $+0.078$]} & $+0.038$\,{\scriptsize[$+0.010$, $+0.069$]} & $+0.025$\,{\scriptsize[$+0.005$, $+0.051$]} & $+0.002$\,{\scriptsize[$+0.000$, $+0.006$]} \\
S3b $-$ S5 & 38/56 & $-0.005$\,{\scriptsize[$-0.013$, $+0.000$]} & $-0.007$\,{\scriptsize[$-0.022$, $+0.000$]} & $-0.003$\,{\scriptsize[$-0.008$, $+0.000$]} & $+0.000$\,{\scriptsize[$+0.000$, $+0.000$]} \\
S3b $-$ S2b$'$ & 38/56 & $+0.076$\,{\scriptsize[$+0.029$, $+0.126$]} & $+0.058$\,{\scriptsize[$+0.011$, $+0.108$]} & $+0.064$\,{\scriptsize[$+0.021$, $+0.113$]} & $+0.042$\,{\scriptsize[$+0.002$, $+0.087$]} \\
\bottomrule
\end{tabular}
\caption{Main paired comparisons (nDCG@10, 95\% cluster-bootstrap intervals) restricted to the queries whose own source both frame judges accept (577 of 722, 80\%); $n$ is accepted/all. S2b$'$: S2b's description verdicts applied to S4@50's top 10. $^\dagger$: the interval's position relative to zero differs from all queries.}
\label{tab:rc-accepted}
\end{table*}

\begin{table}[t]
\centering
\small
\setlength{\tabcolsep}{4pt}
\begin{tabular}{lccccc}
\toprule
System & All & $k=1$ & $k=2$ & $k=3$ & $k=4$ \\
\midrule
\multicolumn{6}{l}{\emph{Contested} ($n$ = 96/3/21/37/35)} \\
S0 & $0.635$ & $0.812$ & $0.561$ & $0.627$ & $0.672$ \\
S4@10 & $0.718$ & $0.922$ & $0.632$ & $0.725$ & $0.744$ \\
S2b$'$ & $0.720$ & $0.978$ & $0.672$ & $0.739$ & $0.705$ \\
S4@50 & $0.809$ & $0.982$ & $0.751$ & $0.840$ & $0.796$ \\
S3b & $0.923$ & $1.000$ & $0.888$ & $0.929$ & $0.931$ \\
S5 & $0.943$ & $1.000$ & $0.899$ & $0.954$ & $0.953$ \\
\midrule
\multicolumn{6}{l}{\emph{Nested sample} ($n$ = 56/12/16/12/16)} \\
S0 & $0.380$ & $0.188$ & $0.322$ & $0.605$ & $0.414$ \\
S4@10 & $0.562$ & $0.314$ & $0.505$ & $0.750$ & $0.664$ \\
S2b$'$ & $0.622$ & $0.472$ & $0.542$ & $0.844$ & $0.647$ \\
S4@50 & $0.656$ & $0.482$ & $0.580$ & $0.917$ & $0.668$ \\
S3b & $0.698$ & $0.541$ & $0.599$ & $0.917$ & $0.750$ \\
S5 & $0.703$ & $0.549$ & $0.610$ & $0.917$ & $0.750$ \\
\midrule
\multicolumn{6}{l}{\emph{All near-miss queries} ($n$ = 372/98/98/98/78)} \\
S0 & $0.450$ & $0.455$ & $0.461$ & $0.436$ & $0.447$ \\
S1 & $0.346$ & $0.476$ & $0.282$ & $0.310$ & $0.310$ \\
S2b@50 & $0.500$ & $0.567$ & $0.485$ & $0.465$ & $0.481$ \\
S4@10 & $0.528$ & $0.541$ & $0.524$ & $0.523$ & $0.521$ \\
S4@50 & $0.626$ & $0.667$ & $0.611$ & $0.617$ & $0.602$ \\
\midrule
\multicolumn{6}{l}{\emph{All nested queries} ($n$ = 350/98/94/85/73)} \\
S0 & $0.408$ & $0.339$ & $0.358$ & $0.475$ & $0.485$ \\
S1 & $0.318$ & $0.318$ & $0.251$ & $0.370$ & $0.345$ \\
S2b@50 & $0.450$ & $0.419$ & $0.397$ & $0.520$ & $0.476$ \\
S4@10 & $0.529$ & $0.415$ & $0.505$ & $0.603$ & $0.627$ \\
S4@50 & $0.636$ & $0.544$ & $0.584$ & $0.717$ & $0.734$ \\
\bottomrule
\end{tabular}
\caption{Absolute nDCG@10 by number of constraints $k$ (expanded labels); $n$ is all/$k{=}1$/$k{=}2$/$k{=}3$/$k{=}4$. S2b$'$ applies S2b's description verdicts to S4@50's top 10.}
\label{tab:rc-perk}
\end{table}

\begin{table*}[t]
\centering
\small
\setlength{\tabcolsep}{3pt}
\begin{tabular}{lrcccc}
\toprule
Group & $n$ & S1 $-$ S0 & S2b $-$ S0 & S4 $-$ S0 & S4 $-$ S2b \\
\midrule
\multicolumn{6}{l}{\emph{Nested set}} \\
All & 350 & $-.089$ \,[$-.125$, $-.055$] & $+.042$ \,[$+.007$, $+.080$] & $+.229$ \,[$+.186$, $+.274$] & $+.187$ \,[$+.149$, $+.226$] \\
\addlinespace[2pt]
$k=1$ & 98 & $-.021$ \,[$-.064$, $+.020$] & $+.079$ \,[$+.044$, $+.115$] & $+.205$ \,[$+.156$, $+.254$] & $+.126$ \,[$+.086$, $+.166$] \\
$k=2$ & 94 & $-.107$ \,[$-.168$, $-.048$] & $+.039$ \,[$-.005$, $+.089$] & $+.226$ \,[$+.166$, $+.288$] & $+.186$ \,[$+.133$, $+.243$] \\
$k=3$ & 85 & $-.106$ \,[$-.172$, $-.046$] & $+.045$ \,[$-.013$, $+.107$] & $+.242$ \,[$+.175$, $+.317$] & $+.197$ \,[$+.140$, $+.257$] \\
$k=4$ & 73 & $-.140$ \,[$-.209$, $-.074$] & $-.009$ \,[$-.070$, $+.055$] & $+.250$ \,[$+.173$, $+.331$] & $+.258$ \,[$+.179$, $+.342$] \\
\addlinespace[2pt]
Vript $<$30\,s & 48 & $-.133$ \,[$-.249$, $-.021$] & $+.048$ \,[$-.052$, $+.158$] & $+.187$ \,[$+.119$, $+.260$] & $+.139$ \,[$+.052$, $+.233$] \\
Vript 30--60\,s & 97 & $-.088$ \,[$-.156$, $-.027$] & $+.062$ \,[$-.008$, $+.150$] & $+.228$ \,[$+.152$, $+.313$] & $+.165$ \,[$+.106$, $+.228$] \\
Vript 1--2\,min & 54 & $-.114$ \,[$-.216$, $-.023$] & $-.026$ \,[$-.094$, $+.042$] & $+.296$ \,[$+.187$, $+.414$] & $+.322$ \,[$+.220$, $+.420$] \\
Vript 2--4\,min & 52 & $+.002$ \,[$-.054$, $+.066$] & $+.071$ \,[$-.004$, $+.161$] & $+.374$ \,[$+.246$, $+.496$] & $+.302$ \,[$+.180$, $+.422$] \\
\addlinespace[2pt]
Shot2Story & 42 & $-.090$ \,[$-.134$, $-.046$] & $+.076$ \,[$-.032$, $+.200$] & $+.188$ \,[$+.084$, $+.346$] & $+.112$ \,[$+.045$, $+.184$] \\
CaReBench & 57 & $-.115$ \,[$-.208$, $-.031$] & $+.015$ \,[$-.054$, $+.066$] & $+.100$ \,[$+.050$, $+.153$] & $+.086$ \,[$+.021$, $+.152$] \\
Vript short & 94 & $-.103$ \,[$-.178$, $-.033$] & $+.054$ \,[$-.005$, $+.120$] & $+.185$ \,[$+.131$, $+.242$] & $+.130$ \,[$+.073$, $+.193$] \\
Vript long & 157 & $-.072$ \,[$-.127$, $-.022$] & $+.035$ \,[$-.021$, $+.099$] & $+.313$ \,[$+.241$, $+.384$] & $+.277$ \,[$+.216$, $+.339$] \\
\midrule
\multicolumn{6}{l}{\emph{Near-miss set}} \\
All & 372 & $-.104$ \,[$-.145$, $-.066$] & $+.050$ \,[$+.017$, $+.082$] & $+.175$ \,[$+.139$, $+.211$] & $+.125$ \,[$+.096$, $+.156$] \\
\addlinespace[2pt]
$k=1$ & 98 & $+.020$ \,[$-.013$, $+.056$] & $+.111$ \,[$+.078$, $+.147$] & $+.212$ \,[$+.164$, $+.261$] & $+.100$ \,[$+.064$, $+.142$] \\
$k=2$ & 98 & $-.179$ \,[$-.239$, $-.123$] & $+.024$ \,[$-.024$, $+.068$] & $+.150$ \,[$+.094$, $+.205$] & $+.126$ \,[$+.082$, $+.171$] \\
$k=3$ & 98 & $-.126$ \,[$-.196$, $-.063$] & $+.029$ \,[$-.022$, $+.079$] & $+.181$ \,[$+.126$, $+.236$] & $+.152$ \,[$+.103$, $+.204$] \\
$k=4$ & 78 & $-.137$ \,[$-.216$, $-.061$] & $+.034$ \,[$-.033$, $+.095$] & $+.155$ \,[$+.099$, $+.213$] & $+.121$ \,[$+.069$, $+.181$] \\
\addlinespace[2pt]
Vript $<$30\,s & 54 & $-.110$ \,[$-.189$, $-.041$] & $+.059$ \,[$-.026$, $+.145$] & $+.190$ \,[$+.076$, $+.301$] & $+.131$ \,[$+.054$, $+.211$] \\
Vript 30--60\,s & 96 & $-.073$ \,[$-.143$, $-.017$] & $+.030$ \,[$-.025$, $+.090$] & $+.181$ \,[$+.124$, $+.240$] & $+.151$ \,[$+.085$, $+.224$] \\
Vript 1--2\,min & 57 & $-.170$ \,[$-.300$, $-.055$] & $+.045$ \,[$-.036$, $+.129$] & $+.199$ \,[$+.111$, $+.290$] & $+.153$ \,[$+.076$, $+.238$] \\
Vript 2--4\,min & 49 & $-.082$ \,[$-.227$, $+.022$] & $+.022$ \,[$-.056$, $+.095$] & $+.107$ \,[$+.027$, $+.182$] & $+.085$ \,[$+.022$, $+.153$] \\
\addlinespace[2pt]
Shot2Story & 57 & $-.121$ \,[$-.236$, $-.026$] & $+.048$ \,[$-.041$, $+.118$] & $+.150$ \,[$+.055$, $+.234$] & $+.102$ \,[$+.028$, $+.187$] \\
CaReBench & 59 & $-.086$ \,[$-.176$, $-.001$] & $+.107$ \,[$+.008$, $+.208$] & $+.213$ \,[$+.114$, $+.325$] & $+.106$ \,[$+.050$, $+.171$] \\
Vript short & 105 & $-.097$ \,[$-.166$, $-.039$] & $+.046$ \,[$-.004$, $+.096$] & $+.157$ \,[$+.090$, $+.223$] & $+.111$ \,[$+.060$, $+.162$] \\
Vript long & 151 & $-.109$ \,[$-.180$, $-.048$] & $+.032$ \,[$-.018$, $+.083$] & $+.184$ \,[$+.132$, $+.237$] & $+.151$ \,[$+.099$, $+.205$] \\
\bottomrule
\end{tabular}
\caption{Paired differences in nDCG@10 (expanded labels) for the systems that read only the index, by number of constraints $k$, Vript length bin and corpus part, with 95\% cluster-bootstrap intervals. S2b and S4 rerank S0's top 50. $n$ is the number of queries.}
\label{tab:perk}
\end{table*}

\begin{table*}[t]
\centering
\small
\setlength{\tabcolsep}{4pt}
\begin{tabular}{lrccc}
\toprule
Group & $n$ & S5 $-$ S4 & S3 $-$ S4 & S5 $-$ S3 \\
\midrule
\multicolumn{5}{l}{\emph{Near-miss set, all queries}} \\
All & 372 & $+.079$ \,[$+.055$, $+.104$] & $+.074$ \,[$+.052$, $+.099$] & $+.005$ \,[$-.002$, $+.012$] \\
\addlinespace[2pt]
$k=1$ & 98 & $+.038$ \,[$+.019$, $+.061$] & $+.030$ \,[$+.014$, $+.047$] & $+.009$ \,[$+.000$, $+.019$] \\
$k=2$ & 98 & $+.092$ \,[$+.059$, $+.128$] & $+.090$ \,[$+.057$, $+.126$] & $+.002$ \,[$-.003$, $+.008$] \\
$k=3$ & 98 & $+.091$ \,[$+.057$, $+.128$] & $+.083$ \,[$+.048$, $+.121$] & $+.008$ \,[$-.001$, $+.019$] \\
$k=4$ & 78 & $+.099$ \,[$+.052$, $+.150$] & $+.100$ \,[$+.058$, $+.146$] & $-.001$ \,[$-.023$, $+.017$] \\
\addlinespace[2pt]
Vript $<$30\,s & 54 & $+.086$ \,[$+.028$, $+.157$] & $+.069$ \,[$+.013$, $+.136$] & $+.016$ \,[$+.001$, $+.036$] \\
Vript 30--60\,s & 96 & $+.084$ \,[$+.042$, $+.133$] & $+.082$ \,[$+.039$, $+.131$] & $+.002$ \,[$-.003$, $+.007$] \\
Vript 1--2\,min & 57 & $+.058$ \,[$+.014$, $+.123$] & $+.058$ \,[$+.014$, $+.123$] & $+.000$ \,[$+.000$, $+.000$] \\
Vript 2--4\,min & 49 & $+.096$ \,[$+.040$, $+.154$] & $+.090$ \,[$+.035$, $+.150$] & $+.006$ \,[$-.000$, $+.017$] \\
\addlinespace[2pt]
Shot2Story & 57 & $+.065$ \,[$-.004$, $+.157$] & $+.078$ \,[$+.017$, $+.166$] & $-.014$ \,[$-.036$, $-.001$] \\
CaReBench & 59 & $+.086$ \,[$+.033$, $+.149$] & $+.067$ \,[$+.029$, $+.113$] & $+.019$ \,[$-.002$, $+.054$] \\
Vript short & 105 & $+.083$ \,[$+.043$, $+.131$] & $+.074$ \,[$+.035$, $+.121$] & $+.009$ \,[$+.000$, $+.020$] \\
Vript long & 151 & $+.079$ \,[$+.046$, $+.117$] & $+.076$ \,[$+.043$, $+.115$] & $+.003$ \,[$-.001$, $+.007$] \\
\midrule
\multicolumn{5}{l}{\emph{Nested set, all queries}} \\
All & 350 & $+.074$ \,[$+.049$, $+.101$] & $+.074$ \,[$+.050$, $+.102$] & $-.000$ \,[$-.005$, $+.004$] \\
\addlinespace[2pt]
$k=1$ & 98 & $+.080$ \,[$+.048$, $+.114$] & $+.087$ \,[$+.054$, $+.124$] & $-.007$ \,[$-.020$, $+.003$] \\
$k=2$ & 94 & $+.086$ \,[$+.052$, $+.122$] & $+.079$ \,[$+.047$, $+.115$] & $+.006$ \,[$-.002$, $+.020$] \\
$k=3$ & 85 & $+.054$ \,[$+.023$, $+.090$] & $+.054$ \,[$+.023$, $+.090$] & $+.000$ \,[$+.000$, $+.000$] \\
$k=4$ & 73 & $+.074$ \,[$+.036$, $+.115$] & $+.074$ \,[$+.036$, $+.115$] & $+.000$ \,[$+.000$, $+.000$] \\
\addlinespace[2pt]
Vript $<$30\,s & 48 & $+.067$ \,[$+.016$, $+.139$] & $+.077$ \,[$+.023$, $+.149$] & $-.010$ \,[$-.028$, $+.001$] \\
Vript 30--60\,s & 97 & $+.082$ \,[$+.038$, $+.145$] & $+.086$ \,[$+.041$, $+.149$] & $-.003$ \,[$-.013$, $+.003$] \\
Vript 1--2\,min & 54 & $+.063$ \,[$+.011$, $+.145$] & $+.063$ \,[$+.011$, $+.145$] & $+.001$ \,[$-.000$, $+.002$] \\
Vript 2--4\,min & 52 & $+.073$ \,[$+.009$, $+.158$] & $+.061$ \,[$+.006$, $+.140$] & $+.012$ \,[$-.000$, $+.034$] \\
\addlinespace[2pt]
Shot2Story & 42 & $+.045$ \,[$+.007$, $+.090$] & $+.048$ \,[$+.008$, $+.094$] & $-.003$ \,[$-.008$, $+.000$] \\
CaReBench & 57 & $+.096$ \,[$+.043$, $+.166$] & $+.094$ \,[$+.040$, $+.165$] & $+.002$ \,[$-.002$, $+.007$] \\
Vript short & 94 & $+.051$ \,[$+.021$, $+.093$] & $+.061$ \,[$+.029$, $+.104$] & $-.010$ \,[$-.022$, $-.001$] \\
Vript long & 157 & $+.087$ \,[$+.045$, $+.139$] & $+.082$ \,[$+.041$, $+.134$] & $+.005$ \,[$+.000$, $+.012$] \\
\midrule
\multicolumn{5}{l}{\emph{Near-miss queries, source and HN video in the pool (contested)}} \\
All & 96 & $+.134$ \,[$+.074$, $+.200$] & $+.120$ \,[$+.063$, $+.184$] & $+.013$ \,[$-.001$, $+.037$] \\
\addlinespace[2pt]
$k=1$ & 3 & $+.018$ \,[$+.000$, $+.053$] & $+.018$ \,[$+.000$, $+.053$] & $+.000$ \,[$+.000$, $+.000$] \\
$k=2$ & 21 & $+.147$ \,[$+.060$, $+.248$] & $+.136$ \,[$+.052$, $+.237$] & $+.011$ \,[$-.000$, $+.032$] \\
$k=3$ & 37 & $+.114$ \,[$+.051$, $+.184$] & $+.092$ \,[$+.024$, $+.164$] & $+.022$ \,[$-.002$, $+.052$] \\
$k=4$ & 35 & $+.157$ \,[$+.081$, $+.241$] & $+.149$ \,[$+.078$, $+.230$] & $+.008$ \,[$-.013$, $+.036$] \\
\addlinespace[2pt]
Vript $<$30\,s & 11 & $+.109$ \,[$-.003$, $+.229$] & $+.071$ \,[$-.082$, $+.227$] & $+.038$ \,[$-.000$, $+.110$] \\
Vript 30--60\,s & 22 & $+.169$ \,[$+.042$, $+.307$] & $+.169$ \,[$+.042$, $+.307$] & $+.000$ \,[$+.000$, $+.000$] \\
Vript 1--2\,min & 20 & $+.112$ \,[$+.002$, $+.282$] & $+.112$ \,[$+.002$, $+.282$] & $+.000$ \,[$+.000$, $+.000$] \\
Vript 2--4\,min & 12 & $+.134$ \,[$+.012$, $+.349$] & $+.134$ \,[$+.012$, $+.349$] & $+.000$ \,[$+.000$, $+.000$] \\
\midrule
\multicolumn{5}{l}{\emph{Nested sample (Vript only)}} \\
All & 56 & $+.047$ \,[$+.019$, $+.078$] & $+.043$ \,[$+.014$, $+.076$] & $+.003$ \,[$-.000$, $+.010$] \\
\addlinespace[2pt]
$k=1$ & 12 & $+.068$ \,[$+.009$, $+.144$] & $+.065$ \,[$+.007$, $+.142$] & $+.002$ \,[$-.000$, $+.006$] \\
$k=2$ & 16 & $+.031$ \,[$+.000$, $+.085$] & $+.020$ \,[$-.022$, $+.077$] & $+.011$ \,[$-.000$, $+.032$] \\
$k=3$ & 12 & $+.000$ \,[$+.000$, $+.000$] & $+.000$ \,[$+.000$, $+.000$] & $+.000$ \,[$+.000$, $+.000$] \\
$k=4$ & 16 & $+.082$ \,[$+.000$, $+.176$] & $+.082$ \,[$+.000$, $+.176$] & $+.000$ \,[$+.000$, $+.000$] \\
\addlinespace[2pt]
Vript $<$30\,s & 17 & $+.067$ \,[$+.009$, $+.136$] & $+.067$ \,[$+.009$, $+.136$] & $+.000$ \,[$+.000$, $+.000$] \\
Vript 30--60\,s & 15 & $+.020$ \,[$+.000$, $+.066$] & $+.009$ \,[$-.030$, $+.066$] & $+.011$ \,[$-.000$, $+.032$] \\
Vript 1--2\,min & 10 & $+.116$ \,[$+.044$, $+.188$] & $+.114$ \,[$+.042$, $+.187$] & $+.003$ \,[$-.000$, $+.006$] \\
Vript 2--4\,min & 14 & $+.000$ \,[$+.000$, $+.000$] & $+.000$ \,[$+.000$, $+.000$] & $+.000$ \,[$+.000$, $+.000$] \\
\bottomrule
\end{tabular}
\caption{Paired differences in nDCG@10 (expanded labels) for the frame systems, which rerank the top 10 of S4@50, by $k$, Vript length bin and corpus part, with 95\% cluster-bootstrap intervals: S5 (whole-query score) and S3 (graded per-constraint verifier) on every query of each set, then on the frozen frame subsets (the contested near-miss queries and the nested Vript sample, where S3b also ran).}
\label{tab:frames}
\end{table*}

\begin{table*}[t]
\centering\small
\begin{tabular}{@{}p{0.11\textwidth}p{0.50\textwidth}p{0.33\textwidth}@{}}
\toprule
Source & Index description (0.5 fps) & Longest nested query built from the source \\
\midrule
$<$30 s\newline carebench\newline 9 s, 8 frames & A yellow CAT excavator digs into a large dirt pit while dump trucks wait nearby in an outdoor construction area surrounded by palm trees and lush vegetation. & ($k=4$) Find a deep, wide pit beside a yellow excavator with a yellow main body, a dense grove behind it, and three dark-colored letters on its mechanical arm. \\
\addlinespace
30--60 s\newline vript short\newline 58 s, 30 frames & A video compilation features a man performing various comedic skits and physical comedy stunts in indoor settings, including wearing a skull mask, balancing objects on his body, and demonstrating a colorful chemical reaction with M\&M candies arranged in a circle on a white plate. & ($k=2$) A person wears a vivid green T-shirt and matching shorts; the room has beige walls and a brown door. \\
\addlinespace
1--2 min\newline vript long\newline 99 s, 50 frames & A man handles a plastic tub filled with moss and white snake eggs in a reptile room lined with rack shelving. Later, three men stand and talk in the same narrow animal housing facility. & ($k=4$) A clip showing oblong pearly-white eggs on wood shavings, a plastic container filled with brown fibrous substrate, someone wearing a turtle graphic T-shirt, and handwritten storage-rack labels. \\
\addlinespace
2--4 min\newline vript long\newline 216 s, 108 frames & A man in a light blue t-shirt sits on a sofa in a living room, demonstrating iOS features on a smartphone screen displayed alongside him. The video features screen recordings of an iPhone interface, showing lock screen widgets, 3D Touch control center shortcuts, and various messaging features including Digital Touch sketches, stickers, and predictive text. & ($k=4$) An iPhone displays emoji stickers, a red scribble, and messaging options for sketch, fireball, kiss, and heartbeat; a purple phone shows Store, Recents, Images, Smileys, and a highlighted virtual hug GIF. \\
\bottomrule
\end{tabular}
\caption{One randomly drawn test source per length bin (seed 1): its index description and the longest nested query built from its caption. The description is all that S0, S2b and S4 see of the video.}
\label{tab:desc_samples}
\end{table*}

\section{Description Length}
\label{app:length}

Section~\ref{sec:results} asks whether longer descriptions would fix what a short description omits. We tested this twice: a development ablation that reran description systems on longer descriptions (most of this appendix), and a test-set run that scores long descriptions against frames on the same ten candidates (next paragraph). The development ablation rewrote the descriptions with three longer prompts and reran S0, S2b and S4 on the nested development set, and S0, S1 and S4 on the near-miss set.

\paragraph{Prompts and length.} The three longer-description prompts differ from the default index-description prompt in the number of sentences requested: 3--5, 5--7 and 14--18 instead of 1--3. The longest (14--18 sentences) also requires every sentence to state a visible fact that no earlier sentence states and forbids summary, transition and atmosphere sentences (for example about pacing or mood), so it differs by more than length. An earlier draft that asked for 10--14 sentences without that rule ended in generic lines and was rewritten. Mean length grows from 44 to 70, 101 and 202 words, and the describer writes 57, 89, 126 and 249 output tokens per video (Table~\ref{tab:length}). Frames, describer, embedder and videos are unchanged. Of the 500 sampled videos, a prompt returned no description for 3 to 4 (4 for the default prompt, which leaves the 496 videos used elsewhere). Each comparison uses the queries whose source has a description under every prompt compared: 362 of the 366 nested queries (97 sources) and 345 of the 353 near-miss queries (98 sources).

\paragraph{Setup.} The development subset is 496 videos stratified by corpus part and length, with the nested set (366 queries from 98 sources) and the near-miss set (353 queries). On the 362 nested queries used in the length tables, S0 scores .654 against .656 on all 366 (the set of the gold-split run in Appendix~\ref{app:results}). As in the other development runs, S2b and S4 rerank S0's top 20 with \texttt{gpt-6-sol} at low effort, with strict labels (source only). Intervals are 95\% cluster-bootstrap intervals over sources (10,000 resamples). The numbers for the default 44-word descriptions come from the earlier development run, not from a run alongside the others.

\paragraph{Test set: long descriptions against frames on the same ten candidates.} We wrote descriptions with the longest (14--18 sentence) prompt (mean 195 words) for the 3,470 videos in the top 10 of S4@50 or of S0 on the 722 test queries, from the same 0.5\,fps frames as the index (Gemini~3.5 Flash-Lite, about 20.3k input tokens per video, about \$23 in all), and ran S4 on them as S4L with the S4 prompt and verifier, on exactly S5's ten candidates in each pool. Only the evidence string changes, so S5 $-$ S4L isolates frames against a longer text and S4L $-$ S4 isolates length. One video's describer output was empty three times; it keeps its 72-word index description (9 pairs, 7 queries). The 195-word mean is the test mean; the development ablation used the same prompt (202 words). Table~\ref{tab:w5} shows that S4L recovers $0.65$ $[0.54, 0.75]$ of the frame gain on S4@50's pool and $0.72$ $[0.54, 0.87]$ on S0's, but frames stay ahead on S4@50's pool, where the 90\% interval of S5 $-$ S4L, $[+0.019, +0.034]$, is not fully contained within $\pm0.02$, so equivalence is not established. On S0's pool the 90\% interval $[+0.006, +0.017]$ lies inside $\pm0.02$, so the two are practically equivalent, although the difference is above zero. Under strict labels the order is the same and S5 $-$ S4L is smaller, with the interval touching zero on S0's pool. On the contested queries ($n=96$ and $69$), the HN video ranks above the source for $20.8$, $6.2$ and $1.0\%$ of queries after S4, S4L and S5 on S4@50's pool and $14.5$, $7.2$ and $0.0\%$ on S0's. At query time S4 reads 2.1k input tokens per query, S4L about 3.9k and S5 80.3k on S4@50's pool; the describer cost is paid once per video, here for the pool videos only.

\begin{table*}[!ht]
\centering
\small
\setlength{\tabcolsep}{4pt}
\begin{tabular}{@{}llccccccc@{}}
\toprule
Pool & Labels & S4 & S4L & S5 & Oracle & S5 $-$ S4L & S4L $-$ S4 & S4L share \\
\midrule
S4@50's top 10 & expanded & .631 & .681 & .707 & .716 & $+.027$\,{\scriptsize[$+.018$, $+.036$]} & $+.050$\,{\scriptsize[$+.034$, $+.067$]} & .65 {\scriptsize[.54, .75]} \\
S4@50's top 10 & strict & .550 & .601 & .616 & .652 & $+.015$\,{\scriptsize[$+.005$, $+.025$]} & $+.051$\,{\scriptsize[$+.034$, $+.069$]} & .77 {\scriptsize[.64, .91]} \\
S0's top 10 & expanded & .528 & .557 & .568 & .575 & $+.011$\,{\scriptsize[$+.005$, $+.018$]} & $+.028$\,{\scriptsize[$+.017$, $+.041$]} & .72 {\scriptsize[.54, .87]} \\
S0's top 10 & strict & .475 & .503 & .510 & .521 & $+.007$\,{\scriptsize[$-.000$, $+.014$]} & $+.029$\,{\scriptsize[$+.016$, $+.042$]} & .81 {\scriptsize[.61, 1.00]} \\
\bottomrule
\end{tabular}
\caption{Long descriptions (S4L, mean 195 words) against short descriptions (S4; S4@10 on S0's pool) and frames (S5) on the same ten candidates, pooled over the 722 test queries: nDCG@10, 95\% cluster-bootstrap intervals over sources (10,000 resamples). Oracle is the best reordering of the ten. Share is the part of S5 $-$ S4 that S4L recovers.}
\label{tab:w5}
\end{table*}

\begin{table}[!ht]
\centering
\small
\setlength{\tabcolsep}{4pt}
\begin{tabular}{lcccc}
\toprule
 & 44w & 70w & 101w & 202w \\
\midrule
Sentences asked & 1--3 & 3--5 & 5--7 & 14--18 \\
Words, mean & 44 & 70 & 101 & 202 \\
Describer output tokens & 57 & 89 & 126 & 249 \\
S4 input tokens per call & 209 & 241 & 278 & 402 \\
S2b input tokens per call & 289 & 319 & 355 & 472 \\
\midrule
S0 nDCG@10 & .654 & .673 & .675 & .701 \\
S2b nDCG@10 & .625 & .673 & .693 & .738 \\
S4 nDCG@10 & .770 & .785 & .825 & .824 \\
\bottomrule
\end{tabular}
\caption{Description-length ablation on the nested development set (362 queries from 97 sources: the 366 nested development queries of the gold-split run whose source has a description under every prompt; strict labels). S0 is therefore .654 here against .656 on all 366. Column heads give the mean description length in words on this set; 44w is the default prompt used elsewhere, and 202w is the longest prompt (195 words on test). Token counts are per rerank call (S0's top 20) and per description.}
\label{tab:length}
\end{table}

\begin{table*}[!ht]
\centering
\small
\setlength{\tabcolsep}{5pt}
\begin{tabular}{lccc}
\toprule
nDCG@10, minus 44w & 70w & 101w & 202w \\
\midrule
S0 & $+.018$\,{\scriptsize[$-.017$, $+.055$]} & $+.021$\,{\scriptsize[$-.020$, $+.064$]} & $+.047$\,{\scriptsize[$+.005$, $+.091$]} \\
S2b & $+.049$\,{\scriptsize[$+.012$, $+.088$]} & $+.068$\,{\scriptsize[$+.026$, $+.114$]} & $+.113$\,{\scriptsize[$+.066$, $+.162$]} \\
S4 & $+.015$\,{\scriptsize[$-.023$, $+.056$]} & $+.055$\,{\scriptsize[$+.010$, $+.101$]} & $+.054$\,{\scriptsize[$+.013$, $+.097$]} \\
\midrule
S4 $-$ S2b at this length & $+.112$\,{\scriptsize[$+.083$, $+.143$]} & $+.132$\,{\scriptsize[$+.101$, $+.165$]} & $+.086$\,{\scriptsize[$+.060$, $+.114$]} \\
S4 $-$ S2b gap narrowed from 44w & $+.033$\,{\scriptsize[$+.004$, $+.065$]} & $+.013$\,{\scriptsize[$-.023$, $+.050$]} & $+.059$\,{\scriptsize[$+.019$, $+.102$]} \\
S4 $-$ S0 at this length & $+.113$\,{\scriptsize[$+.080$, $+.146$]} & $+.149$\,{\scriptsize[$+.113$, $+.187$]} & $+.123$\,{\scriptsize[$+.088$, $+.161$]} \\
\bottomrule
\end{tabular}
\caption{Paired differences for Table~\ref{tab:length} (nested development set, 362 queries, strict labels; 95\% cluster-bootstrap intervals over sources). At 44w the S4 $-$ S2b gap is $+.145$\,{\scriptsize[$+.107$, $+.186$]} and S4 $-$ S0 is $+.116$\,{\scriptsize[$+.079$, $+.155$]}. Between 101w and 202w, S2b changes by $+.045$\,{\scriptsize[$+.002$, $+.085$]} and S4 by $-.001$\,{\scriptsize[$-.043$, $+.040$]}.}
\label{tab:length-gains}
\end{table*}

\paragraph{S4 gains about 0.055 from longer descriptions, and we see no further gain beyond about 100 words.} S4 rises by $+0.015$ $[-0.023, +0.056]$ at 70 words, $+0.055$ $[+0.010, +0.101]$ at 101 and $+0.054$ $[+0.013, +0.097]$ at 202, and the difference between the last two is $-0.001$ $[-0.043, +0.040]$, so a further gain of up to about 0.04 is not excluded (Table~\ref{tab:length-gains}). On the near-miss set the gains are $+0.062$ $[+0.015, +0.117]$ and $+0.068$ $[+0.024, +0.116]$ (Table~\ref{tab:length-j}). S4 stays ahead of S0 at every length: its lead is $+0.116$, $+0.113$, $+0.149$ and $+0.123$ at 44, 70, 101 and 202 words on the nested set, and $+0.117$, $+0.144$ and $+0.154$ on the near-miss set. But S0 itself gains $+0.047$ $[+0.005, +0.091]$ at 202 words on the nested set, and on the near-miss set S0 gains $+0.036$ and $+0.032$ against S4's $+0.062$ and $+0.068$, and S1 gains $+0.042$ $[+0.001, +0.084]$, so at 202 words most of S4's nested-set gain ($+0.054$) may come from a better candidate pool rather than a better rerank. We did not rerank a fixed pool to separate the two.

\paragraph{S2b gains more and keeps gaining, and S4 still leads at every length.} S2b rises by $+0.049$, $+0.068$ and $+0.113$ (Table~\ref{tab:length-gains}), and from 101 to 202 words it still gains $+0.045$ $[+0.002, +0.085]$ where S4 does not. The lead of S4 over S2b shrinks from $+0.145$ $[+0.107, +0.186]$ to $+0.086$ $[+0.060, +0.114]$, a narrowing of $+0.059$ $[+0.019, +0.102]$, but stays above zero. The narrowing is not monotone: it is $+0.033$ at 70 words and $+0.013$ $[-0.023, +0.050]$ at 101 words, so only the shortest and the longest prompt narrow the gap significantly. S2b's gain at 202 words is larger where a query has several constraints (Table~\ref{tab:length-k}): $+0.061$ at $k=1$, not significant, and $+0.125$ to $+0.144$ from $k=2$ to $4$, though the intervals overlap. S2b was not run on the near-miss set.

\begin{table}[!ht]
\centering
\small
\setlength{\tabcolsep}{4pt}
\begin{tabular}{lcccc}
\toprule
 & \multicolumn{2}{c}{S2b} & \multicolumn{2}{c}{S4} \\
\cmidrule(lr){2-3}\cmidrule(l){4-5}
$k$ & 101w & 202w & 101w & 202w \\
\midrule
1 & $+.027$ & $+.061$ & $+.040$ & $+.039$ \\
2 & $+.045$ & $\mathbf{+.125}$ & $+.053$ & $+.048$ \\
3 & $\mathbf{+.108}$ & $\mathbf{+.130}$ & $\mathbf{+.075}$ & $\mathbf{+.076}$ \\
4 & $\mathbf{+.100}$ & $\mathbf{+.144}$ & $+.054$ & $\mathbf{+.055}$ \\
\bottomrule
\end{tabular}
\caption{nDCG@10 gain over 44w by number of constraints $k$ (nested development set). Bold: the 95\% interval excludes zero.}
\label{tab:length-k}
\end{table}

\begin{table}[!ht]
\centering
\small
\setlength{\tabcolsep}{3pt}
\begin{tabular}{lccc}
\toprule
 & 44w & 101w $-$ 44w & 202w $-$ 44w \\
\midrule
S0 & .575 & $+.036$\,{\scriptsize[$-.008$, $+.083$]} & $+.032$\,{\scriptsize[$-.014$, $+.079$]} \\
S1 & .440 & $+.024$\,{\scriptsize[$-.020$, $+.067$]} & $+.042$\,{\scriptsize[$+.001$, $+.084$]} \\
S4 & .693 & $+.062$\,{\scriptsize[$+.015$, $+.117$]} & $+.068$\,{\scriptsize[$+.024$, $+.116$]} \\
\bottomrule
\end{tabular}
\caption{The same comparison on the near-miss development set (345 queries from 98 sources; strict labels). First column: nDCG@10 with 44w; others: paired differences with 95\% intervals. S2b was not run. S4 $-$ S0 is $+.117$\,{\scriptsize[$+.081$, $+.155$]}, $+.144$\,{\scriptsize[$+.106$, $+.182$]} and $+.154$\,{\scriptsize[$+.117$, $+.191$]} at 44w, 101w and 202w.}
\label{tab:length-j}
\end{table}

\paragraph{The longer descriptions cost about 1.2 to 1.9 times the S4 rerank input, a small share of what frames cost.} An S4 call reads 209 tokens per candidate with the default 44-word descriptions, 241, 278 and 402 with the longer prompts, and an S2b call 289, 319, 355 and 472 (Table~\ref{tab:length}); on the near-miss set S4 reads 207, 273 and 395. For comparison, an S3b call on frames reads 5,932 input tokens on the nested development set, where each video has 16 frames; with the 26 frames per candidate of the development run an S3b call reads about 10.8k (Appendix~\ref{app:stats}), and about 8.6k per candidate in the test run. At index time the longest prompt adds about 190 output tokens per video to the roughly 20,000 input tokens its frames already cost (Appendix~\ref{app:stats}), about 1\% more tokens, although output tokens are priced higher.

\paragraph{Longer descriptions also state more things that conflict with the facts a query is built from.} A text-only judge (\texttt{gpt-6-sol}, with the text-only judge prompt of Appendix~\ref{app:prompts}) reads a description and marks each of the source's mined facts as stated, contradicted or not mentioned. These facts were confirmed on the frames by two models (Appendix~\ref{app:mining}), so a contradiction means the description says something different about a detail that is there; statements about anything else go unchecked. Over the nested queries whose source has a description under both prompts (362 queries, 97 sources), the share of facts stated rises from .116 at 44 words to .210 at 101 and .327 at 202, and the share contradicted from .070 to .127 and .187. The differences from the 44-word descriptions are $+.057$ $[+.015, +.104]$ and $+.116$ $[+.076, +.165]$ for contradicted facts and $+.095$ $[+.042, +.152]$ and $+.211$ $[+.157, +.267]$ for stated ones (95\% cluster-bootstrap intervals over sources). At 3--5 sentences (70 words, 366 queries, 98 sources) the shares are .115 to .151 stated and .069 to .097 contradicted, with intervals that include zero. So the contradicted share grows with length, to 2.7 times its 44-word value at 202 words, but so does the stated share: contradicted facts stay at about 0.6 per stated fact at every length (.60, .60 and .57), and stated minus contradicted still rises (.046, .083 and .140). Longer descriptions therefore bring more conflicts but also more matches, and these shares do not show why S4's gain does not grow past about 100 words; we did not test this. A calibration of the 20-video draft of the longest prompt (7 query sources, 25 facts) gave a similar picture but is too small to count. Contradiction is measured only against the mined facts, by a text-only judge whose labels we did not check against human labels; a description that is accurate about a different moment of the video could be marked as contradicting a fact.

\paragraph{Caveats.} The ablation uses strict labels; the expanded labels were judged on the top candidates of the run with the default 44-word descriptions, so they would favour that prompt and we did not recompute them. One describer (Gemini~3.5 Flash-Lite) was used, with 97--98 sources, so the intervals are wide (the lower end of S4's gain is about $+0.01$).

\section{Human Labels}
\label{app:human}

\begin{table}[t]
\centering
\footnotesize
\setlength{\tabcolsep}{3pt}
\resizebox{\columnwidth}{!}{%
\begin{tabular}{llcccc}
\toprule
Labels & Set & S5 & S4@50 & S5 $-$ S4@50 & S5 wins \\
\midrule
Humans & all (235) & $0.657$ & $0.176$ & $+0.481$\,{\scriptsize[$+0.405$, $+0.549$]} & 115/118 \\
 & nested (120) & $0.636$ & $0.169$ & $+0.466$\,{\scriptsize[$+0.363$, $+0.563$]} & 56/57 \\
 & near-miss (115) & $0.678$ & $0.183$ & $+0.496$\,{\scriptsize[$+0.374$, $+0.607$]} & 59/61 \\
\midrule
Expanded (eval) & all (235) & $0.655$ & $0.162$ & $+0.494$\,{\scriptsize[$+0.412$, $+0.570$]} & 118/120 \\
Judges, every video & all (235) & $0.557$ & $0.157$ & $+0.400$\,{\scriptsize[$+0.323$, $+0.477$]} & 97/100 \\
\midrule
Humans, neither top video the source & all (133) & $0.405$ & $0.206$ & $+0.198$\,{\scriptsize[$+0.113$, $+0.287$]} & 28/30 \\
Expanded, same queries & all (133) & $0.398$ & $0.188$ & $+0.211$\,{\scriptsize[$+0.126$, $+0.299$]} & 29/30 \\
\bottomrule
\end{tabular}}
\caption{S5 against S4@50 on all 235 test queries where they rank a different video first (Task A): the share of queries whose top video meets every constraint, the paired difference with a 95\% cluster-bootstrap interval over source videos, and S5's wins among the queries where exactly one top video meets the query. Expanded labels count the source as relevant and the HN video as not; judges, every video applies the two judges' fact answers to every video, the source and HN video included. Queries with a \emph{can't tell} on a deciding fact are left out (2). S5's top video is the source on 95 queries (S4@50's on 7), and 87 of S5's wins are the source; the last two rows leave out every query where either top video is the source. A top-1 contrast on these queries, not nDCG@10.}
\label{tab:human-top1}
\end{table}

\begin{table}[t]
\centering
\footnotesize
\setlength{\tabcolsep}{4pt}
\begin{tabular}{lcccc}
\toprule
Against humans & $n$ & Precision & Recall & $\kappa$ \\
\midrule
\multicolumn{5}{l}{\emph{Per fact on a video (Task A, off the source)}} \\
Gemini & 1105 & $0.848$ & $0.944$ & $0.789$ \\
\texttt{gpt-6-astra} & 1105 & $0.902$ & $0.925$ & $0.835$ \\
Both judges & 1105 & $0.923$ & $0.904$ & $0.838$ \\
Second annotator & 291 & -- & -- & $0.917$ \\
\midrule
\multicolumn{5}{l}{\emph{Per video meeting a whole query (off the source)}} \\
Expanded labels & 789 & $0.825$ & $0.818$ & $0.789$ \\
S5 score $= 10$ & 499 & $0.850$ & $0.321$ & $0.395$ \\
S5 score $\ge 9$ & 499 & $0.782$ & $0.575$ & $0.589$ \\
S5 score $\ge 8$ & 499 & $0.735$ & $0.708$ & $0.648$ \\
S5 score $\ge 7$ & 499 & $0.532$ & $0.783$ & $0.510$ \\
\bottomrule
\end{tabular}
\caption{Model labels against the first annotator on the Task A videos other than the source, that is each disagreement query's two top videos and the HN video, not the judged pool (humans' \emph{met} as truth; \emph{can't tell} left out). Precision and recall are those of the model's yes. Judge \emph{unsure} counts as no. A video meets a query when every constraint of the query is met; off the source the expanded labels are the two judges' answers on every fact. S5 scores each video 0--10 for the whole query, on the videos in that query's S4@50 top 10. Second annotator: agreement with the first on the 25 overlap items, $\kappa$ on met against not met or can't tell.}
\label{tab:human-agree}
\end{table}

\begin{table}[t]
\centering
\footnotesize
\setlength{\tabcolsep}{3pt}
\resizebox{\columnwidth}{!}{%
\begin{tabular}{lccc}
\toprule
 & HN & Nested acceptance & Near-miss acceptance \\
$k$ & valid & Judges / humans (strict) & Judges / humans (strict) \\
\midrule
1 & 1.00 (64) & .89 / .98 (.97) & .93 / 1.00 (.98) \\
2 & .99 (71) & .71 / .98 (.95) & .92 / 1.00 (.97) \\
3 & .97 (67) & .61 / .98 (.93) & .88 / 1.00 (.96) \\
4 & .95 (59) & .51 / .97 (.89) & .86 / .99 (.94) \\
\midrule
All & .98 (261) & .69 / .98 (.94) & .90 / 1.00 (.96) \\
\bottomrule
\end{tabular}}
\caption{Per number of constraints $k$, from humans' labels ($n$ in brackets). HN valid: $d$ met on the source, $d$ not met on the HN video and every shared fact met on the HN video (Task A and C: the HN pair of all 98 near-miss sources, on 261 of their queries; Task A carries only the disagreement queries). Source acceptance: the share of queries whose source both judges accept on every fact, against the share whose every fact humans find present on the source (Task B+D, all 722 test queries); strict counts the 11 facts the annotator noted as borderline as absent. Of the facts at least one judge rejected, humans find 98.6\% present (86.5\% strict; \emph{can't tell} left out).}
\label{tab:human-k}
\end{table}

Section~\ref{sec:setup} sets out the four tasks. The labelling pages show the videos (or, for Task E, the description) and constraints but no system names or model answers. Video roles are not shown, but the source is identifiable on the Task B+D sheet, which shows one video with its queries. One annotator, the same person throughout, labels every task. The sheets were built after an earlier labelling round on many of the same videos, and every answer in them is a human label. Where the annotator had already answered the same fact on the same video, in that round or on another sheet, the cell is pre-filled with that earlier answer, marked as such and open to change, so the same judgment is not made twice: 478 of the 1,472 Task A cells (the 353 source cells, which take the Task B+D answer, and 125 others), all 715 Task B+D cells and all 292 Task C cells. Where the annotator had answered a similar but not identical fact on the video, the earlier answer is shown as a hint and the cell is left empty (719 Task A cells, 15 Task B+D, 157 Task C). The second annotator's sheets are filled the same way from that annotator's own earlier answers (52 of 291 Task A cells, all 69 Task B+D cells). The statistics are computed from these labels by the released analysis code; intervals are 95\% cluster-bootstrap intervals over source videos.

\paragraph{Task A: every top-1 disagreement.}
Task A covers all 235 disagreement queries (120 nested, 115 near-miss); where the two rerankers agree, the comparison cannot change. A source's queries nest, so each source is one item (127 items: 68 nested, 59 near-miss): its longest disagreeing query, the facts of all its disagreeing queries as constraints, and every video any of them needs, that is each query's two top videos plus the source and, for near-miss items, the HN video (2--7 videos, 494 in all), under shuffled letters. The annotator marks each constraint met, not met or can't tell on each video (1,472 labels, 14 can't tell). A video meets a query when every constraint of the query is met. The source column asks the same question as Task B+D on the same video, so it holds the Task B+D answer (no cell was changed). The comparisons with model labels below are off the source and use the other 1,119 labels; on the source, the same 353 answers are compared with the judges through Task B+D, which covers every fact.

\paragraph{Task B+D: every fact on its source.}
For each of the 100 sources, the annotator marks every fact of every one of its queries present, absent or can't tell on the source video (715 facts covering all 722 queries; the queries are shown read-only), without seeing which facts a judge rejected. A query is accepted by humans when every one of its facts is present. Task B is the part of this sheet that judges rejected (74 facts in 145 queries), Task D the part they accepted (641 facts).

\paragraph{Task C: the other near-miss sources.}
Task A's near-miss items carry the HN check for 59 sources, but only on their disagreement queries. Task C covers the other 39, with met, not met or can't tell as the answers. Together they check the HN pair of all 98 near-miss sources on 261 of the 372 near-miss queries. A near-miss query is valid when $d$ is met on the source, not met on the HN video, and every shared fact is met on the HN video.

\paragraph{Task E: what the description states.}
For each of the 100 sources, the annotator reads the index description, never the video, and marks each of the source's 715 facts stated, implied or absent.

\paragraph{Agreement and borderline facts.}
A second annotator relabels 25 Task A items (seeded, stratified by set; 291 labels), 10 Task B+D sources (69 facts) and 10 Task E sources (69 facts), and labels all 39 Task C items (292 cells). On Task C the two agree on 290 of 292 cells (99.3\%; $\kappa=0.97$ $[0.93, 1.00]$ on met against the rest, the same on the three answers since neither used can't tell); the two disagreements, C05's source (``A person is visible handling a rifle'': not met against met) and C37's HN video (``A girl wearing black clothing is present'': met against not met), are items the first annotator had noted as borderline. Of the 146 Task C near-miss queries, 146 are valid for the first annotator and 145 for the second, who differs only on C37's $k=4$ query. On Task A the two agree on 94\% of labels ($\kappa=0.88$ on the three answers, $0.92$ on met against the rest; $0.91$ off the source and $0.93$ on cells that were not pre-filled). Of the 17 disagreements, the first annotator says met where the second says not met 8 times and the reverse 3 times, and says can't tell where the second decides 6 times. On Task B+D they agree on 68 of 69 facts, but only 4 of the 69 were rejected by a judge and almost every answer is present, so this overlap does not test the rejected facts and $\kappa$ is uninformative. On Task E they agree on 77\% ($\kappa=0.59$ on the three answers, $0.73$ $[0.53, 0.90]$ on stated against the rest); 14 of the 16 disagreements involve \emph{implied}, so stated is the more reliable label and implied the soft one. Where the annotators disagree, the first annotator's label is kept. The first annotator noted 11 Task B+D facts on 7 sources as borderline (a faint colour, a partly visible hand, a green object that may be a box or a table) and resolved 9 of them as present and 2 as can't tell. The strict reading counts these 11 as absent.

\paragraph{Results: the top-1 contrast.}
On the 235 disagreement queries (233 decided; Table~\ref{tab:human-top1}), S5 is right on 115 of the 118 queries where exactly one top video is, and its lead over S4@50 is similar on nested and near-miss queries ($+0.47$ and $+0.50$) and under expanded labels ($+0.49$). S5's top video is the source on 95 queries and S4@50's on 7, and 87 of S5's 115 wins are the source. Leaving out every query where either top video is the source, humans give $+0.20$ $[+0.11, +0.29]$ (28 of 30 decisive) and expanded labels $+0.21$. The source labels behind this come from the Task B+D sheet, so the source-recovery part of the contrast and the source acceptance below are not independent checks. The judges' answers alone give a smaller difference ($+0.40$), because they reject facts that are present on the source.

\paragraph{Results: model labels against humans.}
Off the source, both judges together have per-fact precision 0.92, recall 0.90 and $\kappa=0.84$; \texttt{gpt-6-astra} agrees with humans more than Gemini (Table~\ref{tab:human-agree}). Per video and query off the source, where the expanded labels are the judges' answers, they miss 22 of the 121 videos humans find relevant and add 21 that humans reject (precision 0.83, recall 0.82; 13 of the 22 misses at $k=1$). By $k$, precision is 0.86, 0.81, 0.73 and 0.71 and recall 0.84, 0.77, 0.92 and 0.62 at $k=1$ to 4 (Table~\ref{tab:human-agree-k}); every interval at $k\ge2$ contains the $k=1$ value, and nested queries alone have only 6, 1 and 0 relevant pairs at $k=2$ to 4, so these labels cannot show whether agreement falls with $k$.

\begin{table}[t]
\centering\footnotesize\setlength{\tabcolsep}{4pt}
\begin{tabular}{lcccc}
\toprule
$k$ & $n$ (relevant) & Precision & Recall & $\kappa$ \\
\midrule
1 & 216 (79) & $0.86$ [$0.77, 0.94$] & $0.84$ [$0.75, 0.91$] & $0.76$ \\
2 & 213 (22) & $0.81$ [$0.63, 0.95$] & $0.77$ [$0.57, 0.95$] & $0.77$ \\
3 & 182 (12) & $0.73$ [$0.43, 0.94$] & $0.92$ [$0.67, 1.00$] & $0.80$ \\
4 & 178 (8)  & $0.71$ [$0.25, 1.00$] & $0.62$ [$0.33, 1.00$] & $0.65$ \\
\bottomrule
\end{tabular}
\caption{Off-source agreement of expanded labels with humans (Table~\ref{tab:human-agree}) by the query's $k$, nested and near-miss pooled. $n$ counts (query, video) pairs; in brackets, the pairs humans find relevant. 95\% cluster-bootstrap intervals over source videos.}
\label{tab:human-agree-k}
\end{table}

 These videos are the disagreement queries' top videos and HN videos, not a sample of the judged pool. On the source, expanded labels count the source relevant by rule; humans reject it on 4 of those rows. Among S5's scores off the source, a threshold of 8 agrees best with humans ($\kappa=0.65$).

\paragraph{Results: facts on the source and near-miss validity.}
Humans find 709 of the 715 facts present on their source, 3 absent and 3 can't tell, so every fact is present for 706 of the 715 queries without a can't-tell fact (98.7\%; 97.8\% of all 722). Two of the absent facts were accepted by both judges, one was rejected by one judge; of the 72 rejected facts humans could decide, 71 are present (64 of 74 under the strict reading). The judges accept 79.9\% of the queries (Table~\ref{tab:human-k}); their gap to humans is largest on nested queries at $k=4$ (\S\ref{sec:res-labels}), where one rejected fact rejects every query that contains it. Of the 261 checked near-miss queries, 255 are valid for humans (97.7\% $[95.7, 99.3]$): $d$ is on the source and not on the HN video in all of them, and the six failures are a shared fact missing from the HN video, all at $k\ge2$ and all among the Task A queries. The validity check does not ask that the shared facts be on the source; requiring that too gives 249 of 261 (95.4\%). Two Task C source cells disagree with the same fact on the Task B+D sheet (C05, a person handling a rifle, of which the annotator noted only a hand is visible: not met in Task C, present in Task B+D; C16: not met against can't tell). The judges' labels give 92.3\% valid; their HN-side answers are met by construction, so this figure measures only $d$ on the source.

\paragraph{Results: what the descriptions state.}
The index descriptions state 32\% $[29, 36]$ of the facts, imply 15\% and leave 52\% $[48, 56]$ absent. The share differs by set: 73\% of the facts of nested queries are absent, against 33\% of the coarser facts of near-miss queries. Attributes are absent most often (77\%; 83\% in nested queries) and scenes least often (32\%, with 31\% implied); fact type is confounded with set, since most object facts are near-miss facts. Per query, every fact is stated for 14\% of queries: 18\% of nested queries at $k=1$ and at most 2\% at $k=2$ to 4 (1\% at $k=4$), and 40\% of near-miss queries at $k=1$ falling to 5\% at $k=4$; with implied counted as stated, every fact is covered for 27\% and 1\% of nested queries at $k=1$ and $k=4$. A description reranker thus rarely reads every condition of a nested query at $k\ge2$; the frame gain, however, is similar at every $k$, so these labels describe the descriptions and do not by themselves explain the gain. Against the text-only fact judge of App.~\ref{app:mining} on the 350 kept nested facts both cover, humans' stated and the judge's stated agree with $\kappa=0.71$; humans mark 51 of these facts stated and the judge 40, and the judge, which has no \emph{implied} answer, marks 37 of the 44 facts humans find implied as unclear.

\end{document}